\documentclass[10pt,twocolumn,letterpaper]{article}
\usepackage{cvpr} 
\usepackage{xcolor}
\definecolor{cvprblue}{rgb}{0.21,0.49,0.74}
\usepackage[pagebackref,breaklinks,colorlinks,citecolor=cvprblue]{hyperref}

\usepackage{overpic}

\definecolor{dblue}{rgb}{0.0,0.0,0.5}
\definecolor{dgreen}{rgb}{0.0,0.5,0.0}
\definecolor{dred}{rgb}{0.6,0.0,0.0}
\definecolor{dorange}{rgb}{0.6,0.25,0.0}
\definecolor{dyellow}{rgb}{0.5,0.5,0.0}

\newcommand{\ignorethis}[1]{}

\begin{document}

\title{DiLightNet: Fine-grained Lighting Control \\ 
for Diffusion-based Image Generation}
\author
{
\parbox{\textwidth}{\centering
Chong Zeng$^{1,2}$ \quad Yue Dong$^{2}$ \quad Pieter Peers$^{3}$ \quad Youkang Kong$^{4,2}$ \quad Hongzhi Wu$^{1}$ \quad Xin Tong$^{2}$
}\\
\parbox{\textwidth}{\centering $^{1}$State Key Lab of CAD and CG, Zhejiang University \quad $^{2}$Microsoft Research Asia \\ $^{3}$College of William \& Mary \quad $^{4}$Tsinghua University}
%}
}

\twocolumn[{%
\renewcommand\twocolumn[1][]{#1}%
\maketitle
%!TEX root = ../../DiffusionRelightHint.tex

\newcommand{\teaserFigWidth}{0.253\textwidth}
%\begin{teaserfigure}
\centering
\renewcommand{\arraystretch}{0.25}
\addtolength{\tabcolsep}{-6.5pt}
 \begin{tabular}{ ccccc }
 \includegraphics[width=\teaserFigWidth]{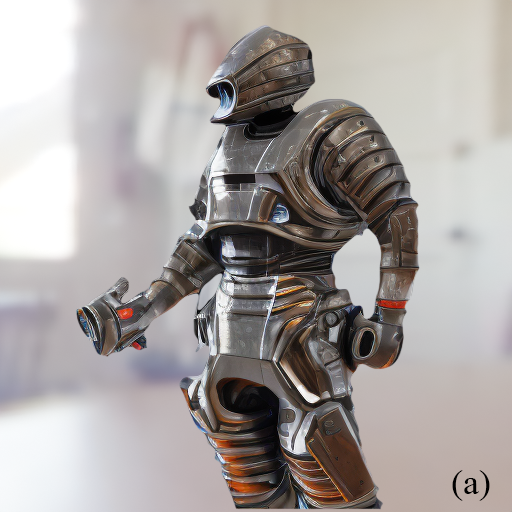}
 &
 \includegraphics[width=\teaserFigWidth]{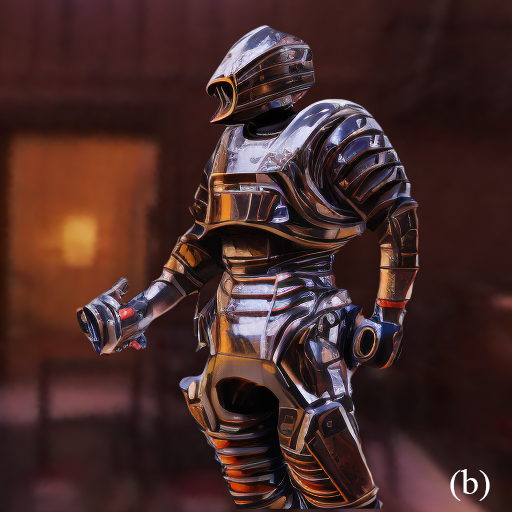}
 &
 \includegraphics[width=\teaserFigWidth]{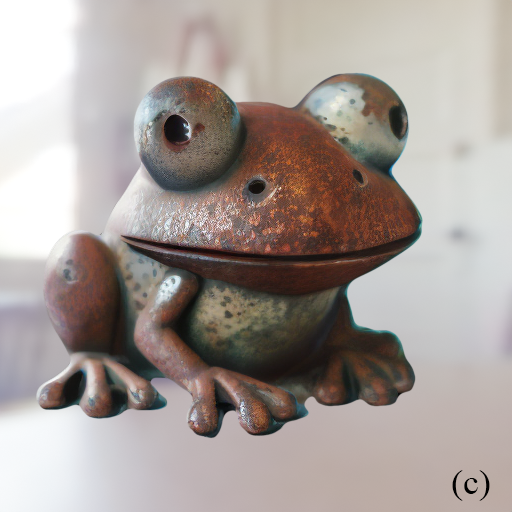}
 &
 \includegraphics[width=\teaserFigWidth]{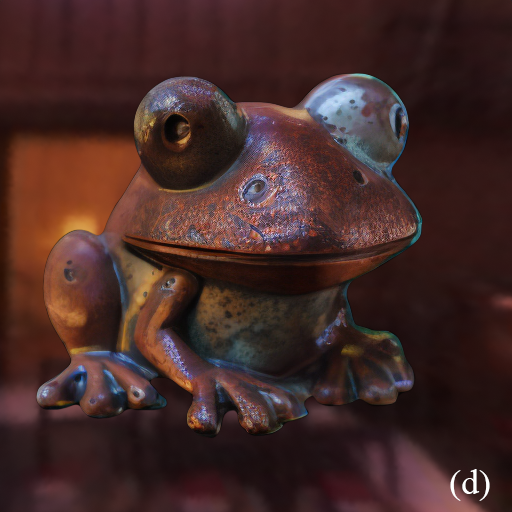} \\
  \multicolumn{2} {c} {
  \scriptsize {
    \emph{``futuristic soldier with advanced armor weaponry and helmet''}
    }}
  &\multicolumn{3} {c} {
  \scriptsize {
    \emph{``rusty steel toy frog with spatially varying materials with the body diffuse but shinny eyes''}
    }}
 \end{tabular} 
 \captionof{figure}{Examples of generated images specified via a text-prompt
   (listed below each example) and with fine-grained lighting
   control. Each prompt is plausibly visualized under two different
   user-provided lighting environments.}
  \label{fig:teaser}
%\end{teaserfigure}

\vspace{2em}
}]

%!TEX root = ../DiffusionRelightHint.tex

\begin{abstract}

  This paper presents a novel method for exerting fine-grained
  lighting control during text-driven diffusion-based image
  generation. While existing diffusion models already have the ability
  to generate images under any lighting condition, without additional
  guidance these models tend to correlate image content and
  lighting. Moreover, text prompts lack the necessary expressional
  power to describe detailed lighting setups.  To provide the content
  creator with fine-grained control over the lighting during image
  generation, we augment the text-prompt with detailed lighting
  information in the form of radiance hints, i.e., visualizations of
  the scene geometry with a homogeneous canonical material under the
  target lighting.  However, the scene geometry needed to produce the
  radiance hints is unknown.  Our key observation is that we only need
  to guide the diffusion process, hence exact radiance hints are not
  necessary; we only need to point the diffusion model in the right
  direction.  Based on this observation, we introduce a three stage
  method for controlling the lighting during image generation.  In the
  first stage, we leverage a standard pretrained diffusion model to
  generate a provisional image under uncontrolled lighting.  Next, in
  the second stage, we resynthesize and refine the foreground object
  in the generated image by passing the target lighting to a refined
  diffusion model, named DiLightNet, using radiance hints computed on
  a coarse shape of the foreground object inferred from the
  provisional image.  To retain the texture details, we multiply the
  radiance hints with a neural encoding of the provisional synthesized
  image before passing it to DiLightNet.  Finally, in the third stage,
  we resynthesize the background to be consistent with the lighting on
  the foreground object.  We demonstrate and validate our lighting
  controlled diffusion model on a variety of text prompts and lighting
  conditions.
\end{abstract}

%!TEX root = ../DiffusionRelightHint.tex

\section{Introduction}
\label{sec:intro}

Text-driven generative machine learning methods, such as diffusion
models~\cite{Nichol:2022:GTP,Ramesh:2022:HTC,Rombach:2022:HRI,Saharia:2022:PTI},
can generate fantastically detailed images from a simple text prompt.
However, diffusion models also have built in biases. For example,
Liu~\etal~\shortcite{Liu:2023:Z1T} demonstrate that diffusion models
tend to prefer certain viewpoints when generating images.  As shown
in~\autoref{fig:lightingbias}, another previously unreported bias is
the lighting in the generated images.  Moreover, the image content and
lighting are highly correlated. While diffusion models have the
capability to sample different lighting conditions, there currently
does not exist a method to precisely control the lighting and the
image content independently in the generated images.

In this paper we aim to exert fine-grained control on the effects of
lighting during diffusion-based image generation
(\autoref{fig:teaser}).  While text prompts have been used to provide
relative control of non-rigid deformations of
objects~\cite{Cao:2023:MTF,Kawar:2023:MTR}, the identity and gender of
subjects~\cite{Kim:2022:DTG}, and the material
properties~\cite{Sharma:2023:APC} of objects, it is more difficult to
impose precise control over the lighting via a text prompt; language
generally offers only qualitative (e.g., warm, cold, cozy, etc.) and
coarse positional (e.g., left, right, rim-lighting, etc.) descriptions
of lighting.  Furthermore, current text embeddings also have
difficulty in encoding fine-grained information~\cite{Paiss:2023:TCT}.
However, due to the entanglement of the lighting and text embeddings,
simply conditioning the text-to-image model on the lighting (e.g., by
passing the light direction) will not allow for independent control of
lighting and image content.  Moreover, using a lighting representation
such as a light direction vector or an environment map limits the
types of lighting that can control the image generation.

In this paper we employ an alternative method of passing lighting
conditions, namely radiance hints; a rendering of the target scene
with a canonical homogeneous material lit by the target
lighting. However, this typically requires precise knowledge of the
underlying geometry which is unknown in the case of text-driven image
generation.  A key observation is that even though the diffusion
model's sampling of the distribution of images is biased in terms of
lighting, the learned distribution does contain the effects of
different lighting conditions.  Hence, in order to control the
lighting during image generation, we need to guide the diffusion
sampling process. Armed with this key observation, we revisit radiance
hints and note that for guiding the sampling process, we do not need
exact radiance hints, only a coarse approximation; we rely on the
generative powers of the diffusion model to fill in the details.

We present a novel three stage method for providing fine-grained
lighting control for diffusion-based image generation from text
prompts.  Since the background in an image is part of the lighting
condition imposed on the foreground object, we focus primarily on
controlling the lighting on the foreground object, allowing the
background to change accordingly.  In a first stage, we generate a
provisional image of the given text prompt under uncontrolled (biased)
lighting using a standard pretrained diffusion model.  In the second
stage, we compute a proxy shape from the provisional image using an
off-the-shelf depth estimation network~\cite{Bhat:2023:ZZS} and
foreground mask generator~\cite{Qin:2020:UGD}, from which we generate
a set of radiance hints. Next, we resynthesize the image that matches
both the text-prompt and the radiance hints using a refined diffusion
model named \emph{DiLightNet} (\textbf{Di}ffusion \textbf{Light}ing
Control\textbf{Net}).  To retain the rich texture information, we
transform the generated provisional image using a learned encoder and
multiply it with the radiance hints before passing it to DiLightNet.
In the third stage, we inpaint a new background consistent with the
target lighting. As our model is derived from large scale pretrained
diffusion models, we can generate multiple replicates of the
synthesized image that samples ambiguous interpretations of the
materials.

We demonstrate our lighting controlled diffusion model on a variety of
text-prompt-generated images and under different types of lighting,
ranging from point lights to environment lighting. In addition, we
perform an extensive ablation study to demonstrate the efficacy of
each of the components that comprise DiLightNet.

%!TEX root = ../../DiffusionRelightHint.tex

\newcommand{\biastestFigWidth}{0.061\textwidth}

\begin{figure}
\renewcommand{\arraystretch}{-0.1}
\addtolength{\tabcolsep}{-6.0pt}
 \begin{tabular}{ cccccccc }
  \includegraphics[width=\biastestFigWidth]{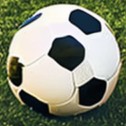}
 &\includegraphics[width=\biastestFigWidth]{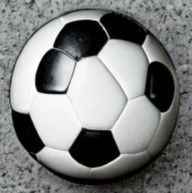}
 &\includegraphics[width=\biastestFigWidth]{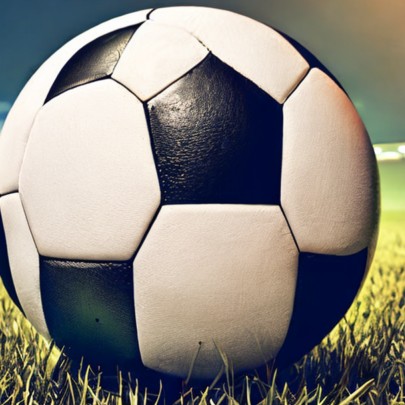}
 &\includegraphics[width=\biastestFigWidth]{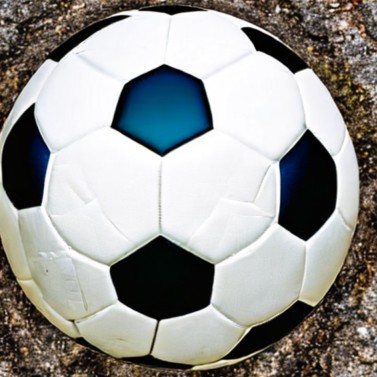}
 &\includegraphics[width=\biastestFigWidth]{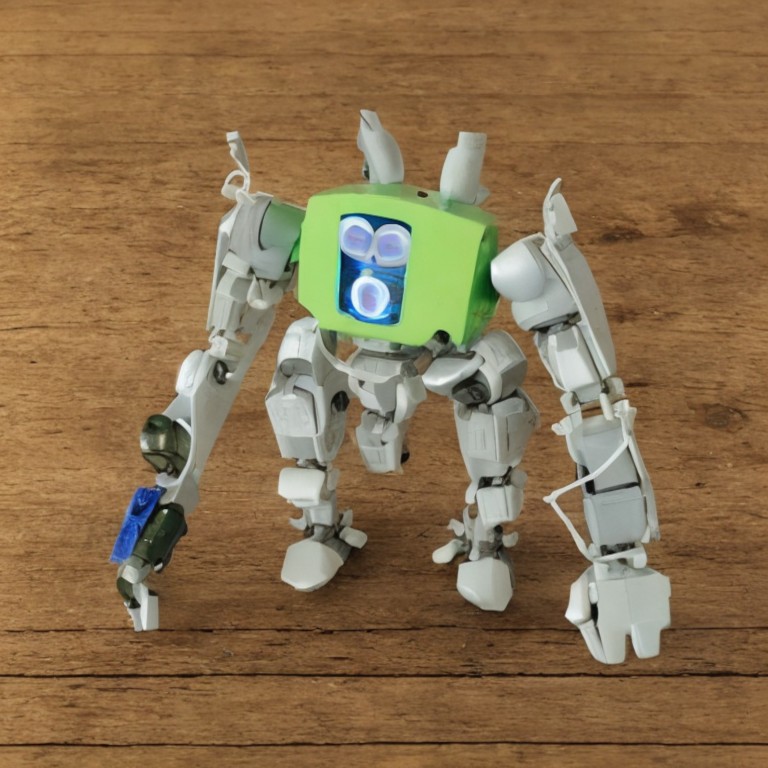}
 &\includegraphics[width=\biastestFigWidth]{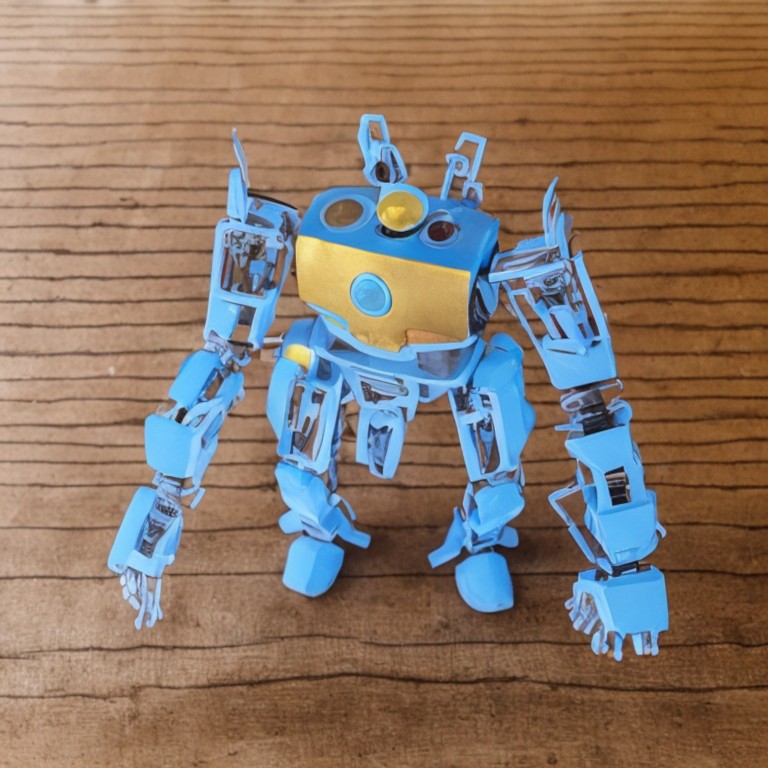}
 &\includegraphics[width=\biastestFigWidth]{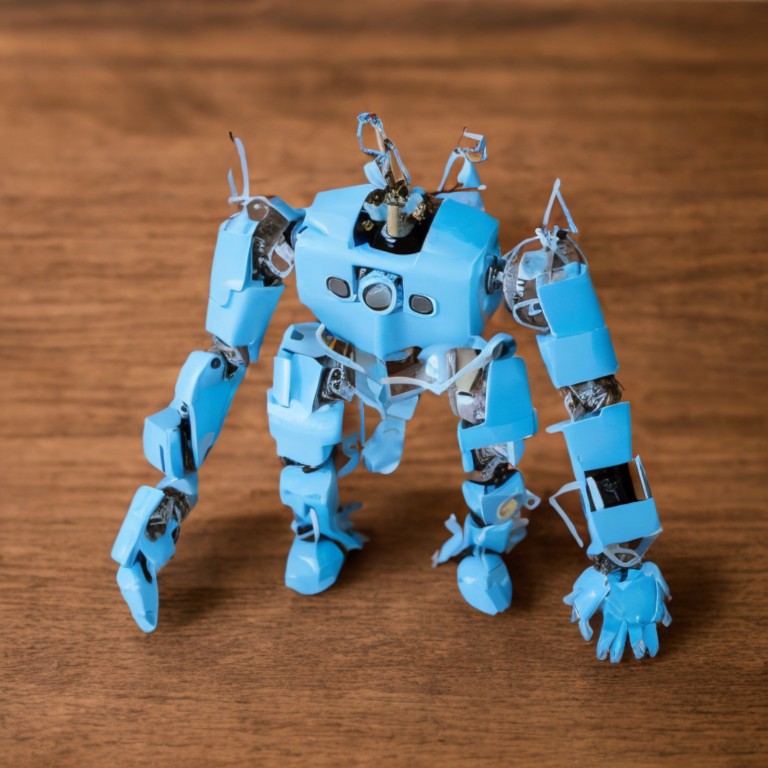}
 &\includegraphics[width=\biastestFigWidth]{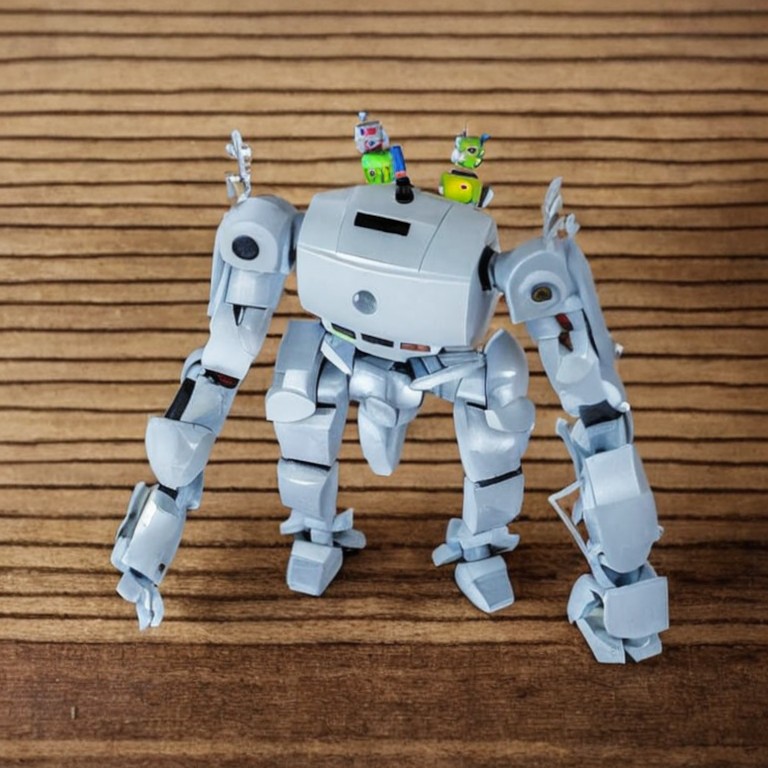}\\
 \includegraphics[width=\biastestFigWidth]{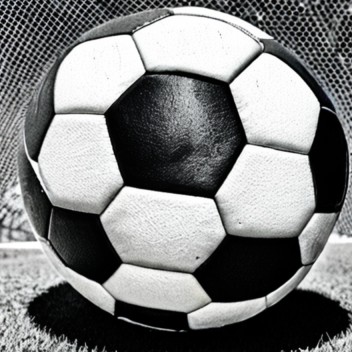}
 &\includegraphics[width=\biastestFigWidth]{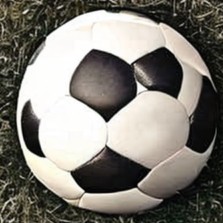}
 &\includegraphics[width=\biastestFigWidth]{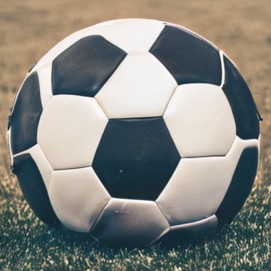}
 &\includegraphics[width=\biastestFigWidth]{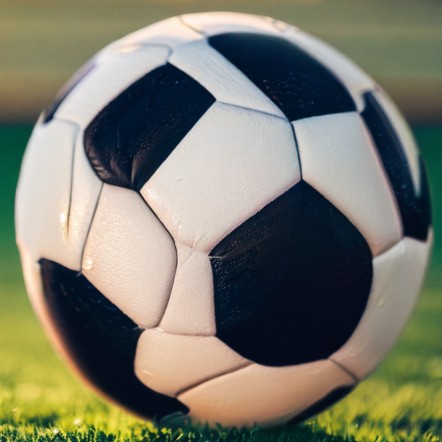}
 &\includegraphics[width=\biastestFigWidth]{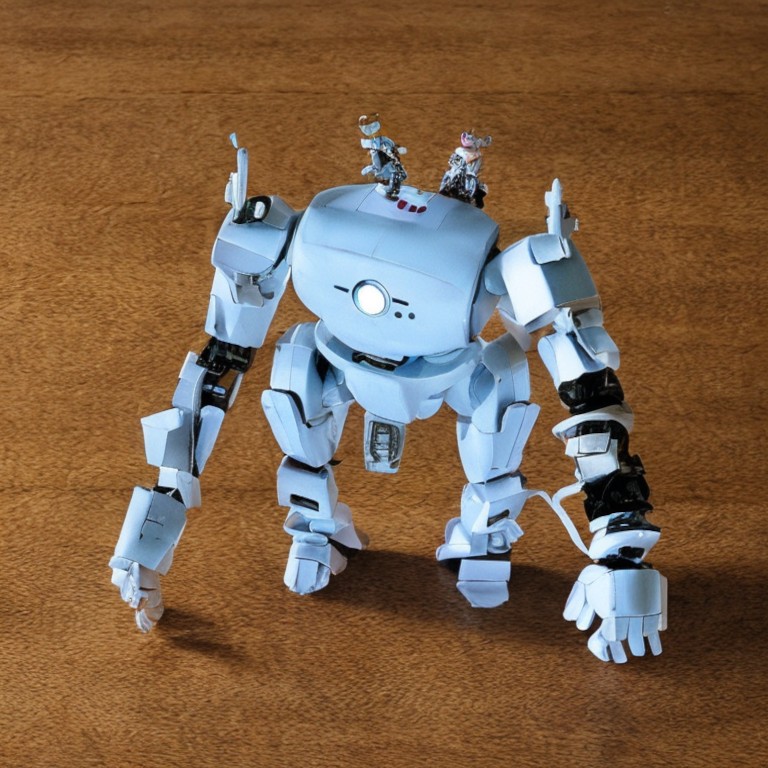}
 &\includegraphics[width=\biastestFigWidth]{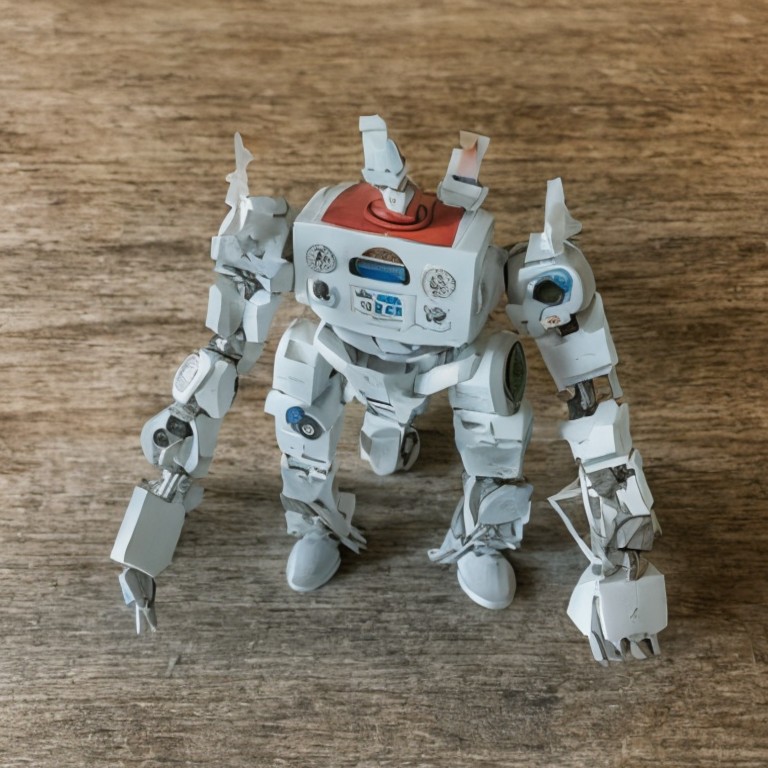}
 &\includegraphics[width=\biastestFigWidth]{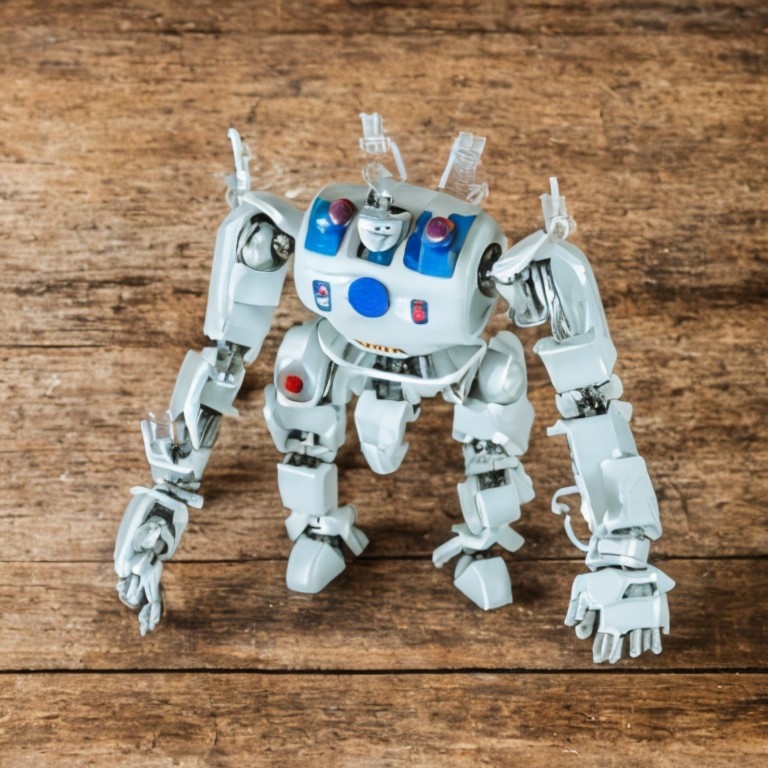}
 &\includegraphics[width=\biastestFigWidth]{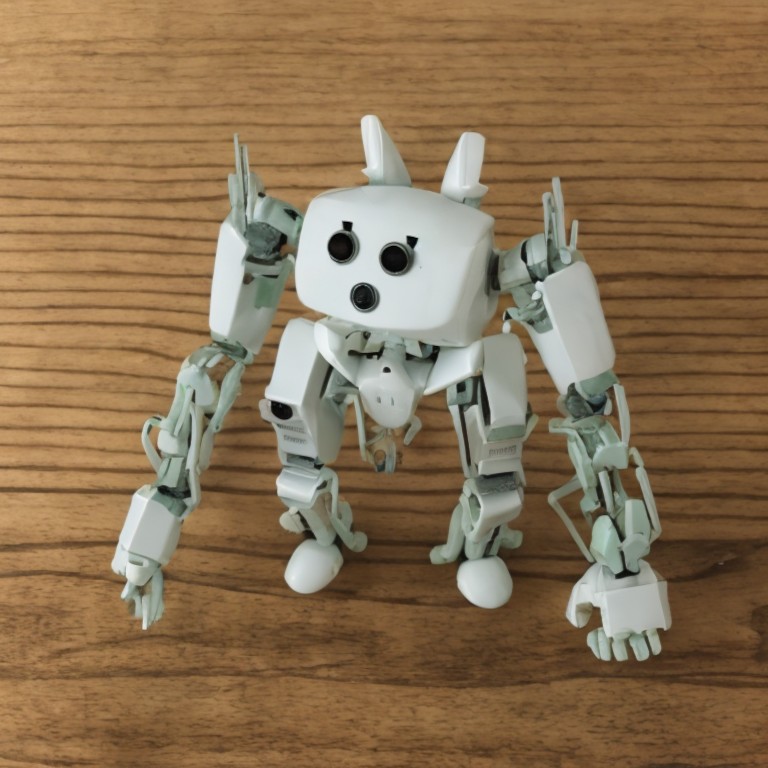}\\
 \includegraphics[width=\biastestFigWidth]{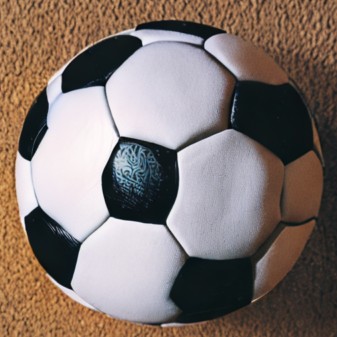}
 &\includegraphics[width=\biastestFigWidth]{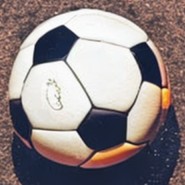}
 &\includegraphics[width=\biastestFigWidth]{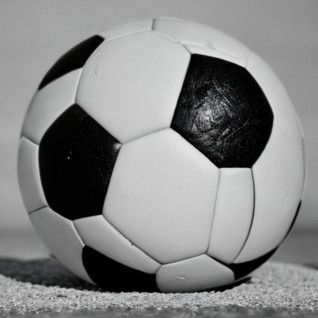}
 &\includegraphics[width=\biastestFigWidth]{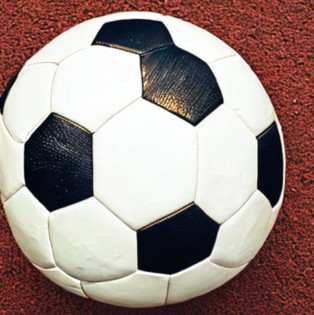}
 &\includegraphics[width=\biastestFigWidth]{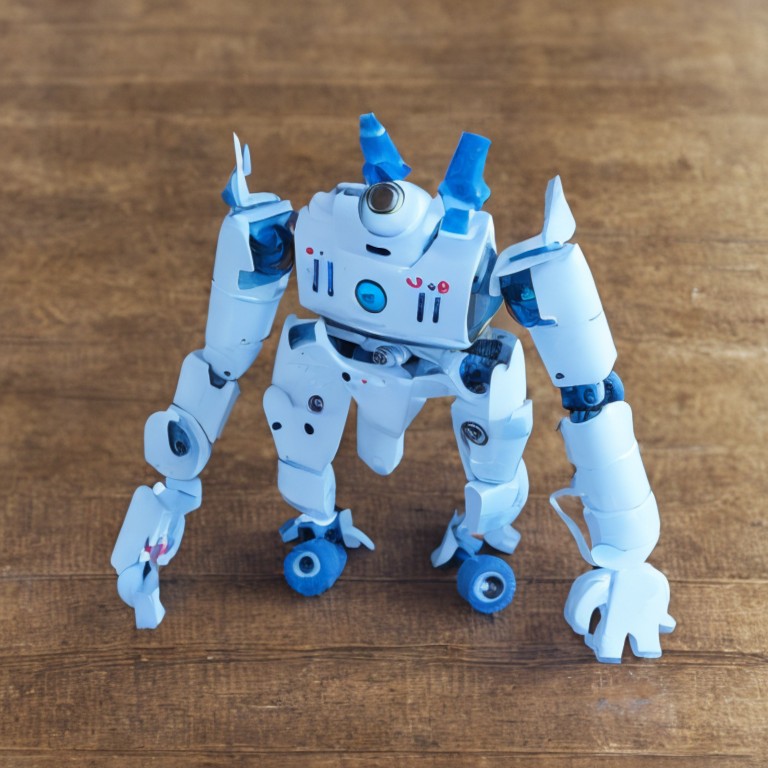}
 &\includegraphics[width=\biastestFigWidth]{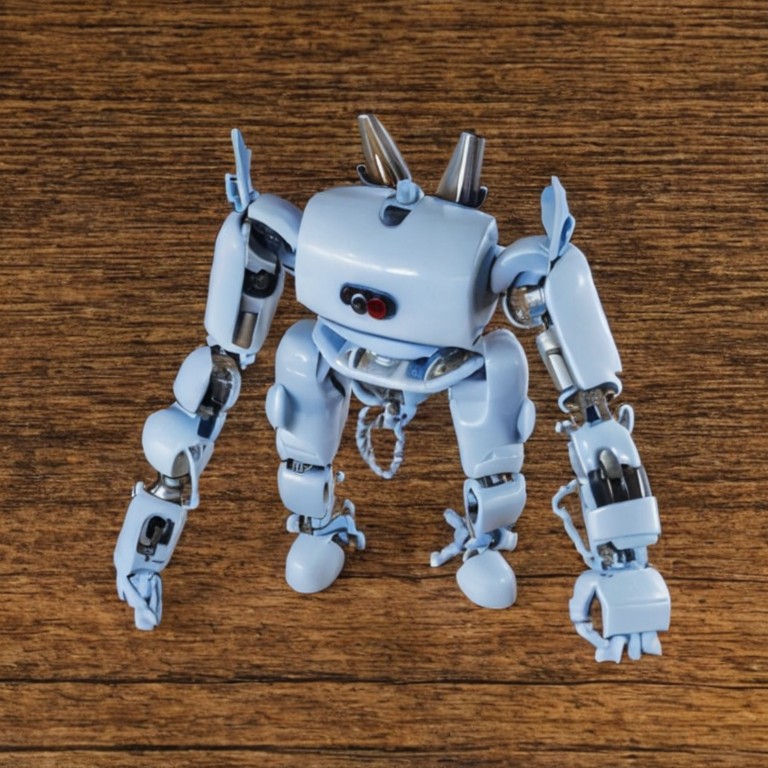}
 &\includegraphics[width=\biastestFigWidth]{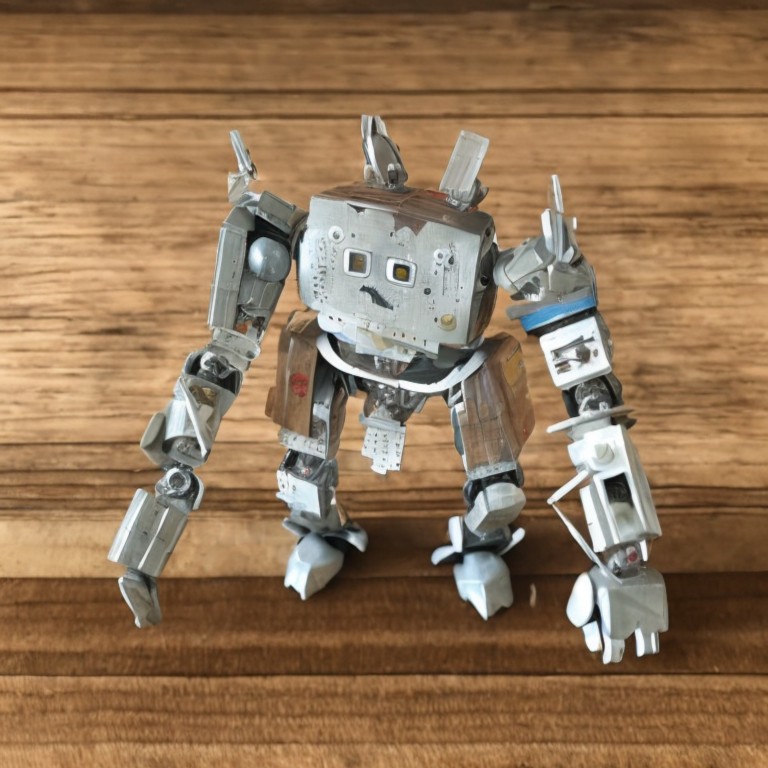}
 &\includegraphics[width=\biastestFigWidth]{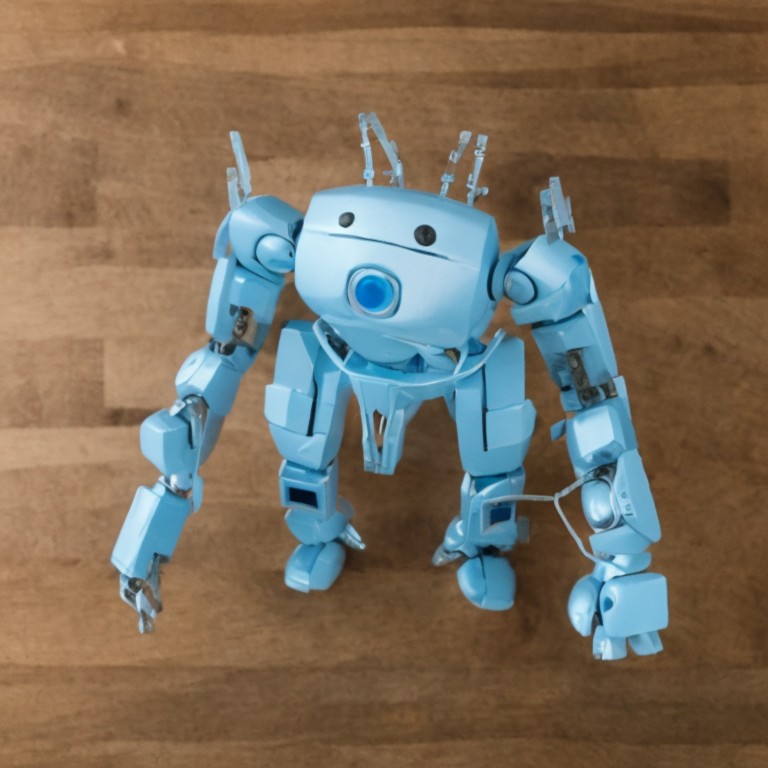}\\
 \includegraphics[width=\biastestFigWidth]{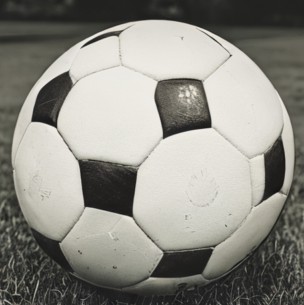}
 &\includegraphics[width=\biastestFigWidth]{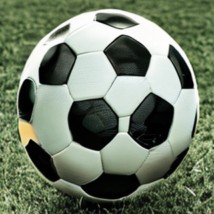}
 &\includegraphics[width=\biastestFigWidth]{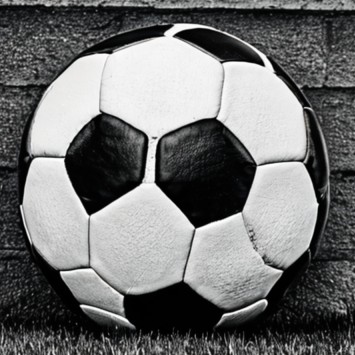}
 &\includegraphics[width=\biastestFigWidth]{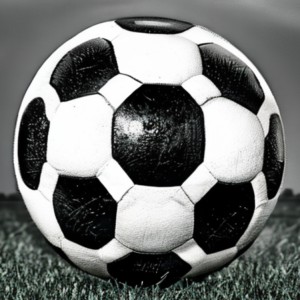}
 &\includegraphics[width=\biastestFigWidth]{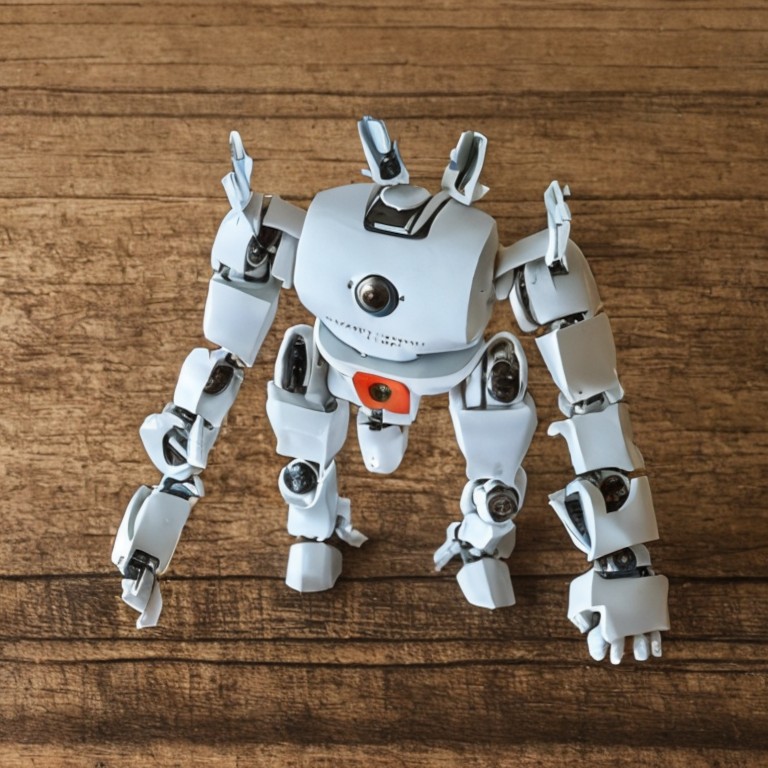}
 &\includegraphics[width=\biastestFigWidth]{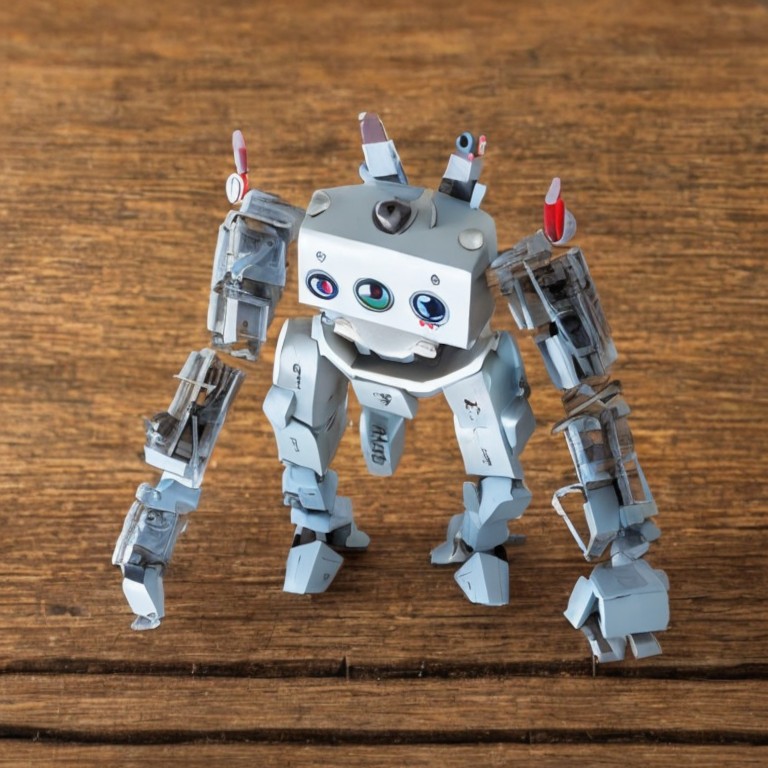}
 &\includegraphics[width=\biastestFigWidth]{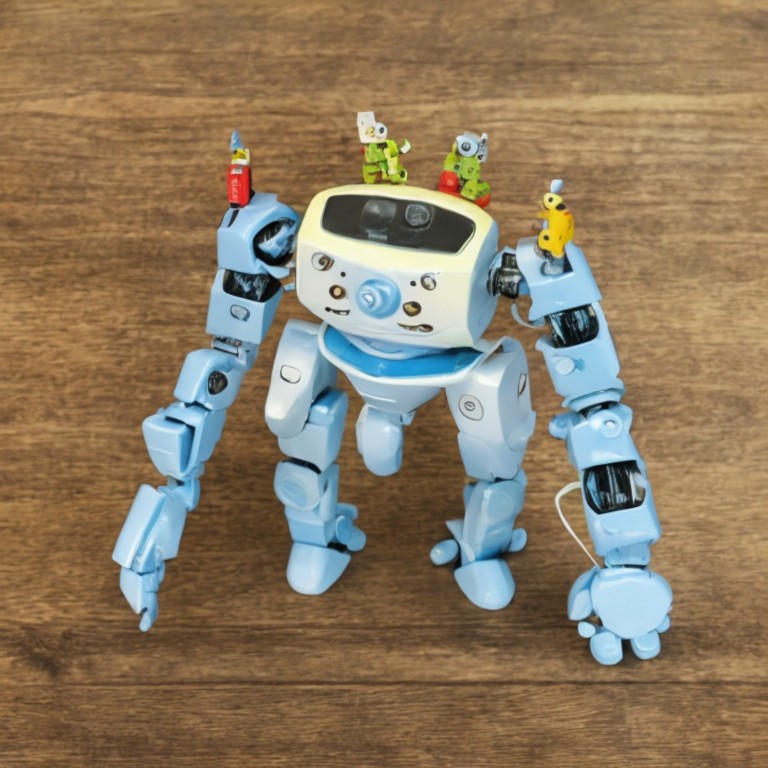}
 &\includegraphics[width=\biastestFigWidth]{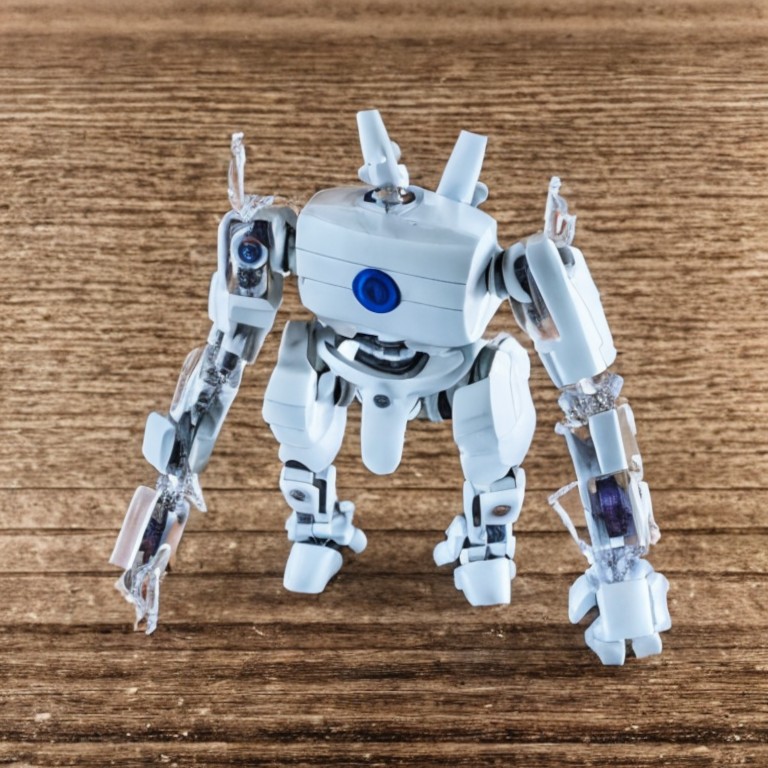}\\
 \end{tabular} 
 \caption{Examples of lighting bias in diffusion-based image
   generation. Left: a batch of $16$ images (text prompt: \emph{``a
     photo of a soccer ball''}). The majority of the images are lit by
   a flash light; only two exhibit off-center lighting (3rd row, 1st
   column and 3rd column).  Right: a batch of generated images of a
   robot dominated by light coming from either the front-left or
   front-right (text prompt: \emph{``a photo of a toy robot standing
     on a wooden table''}; images are generated with a depth
   conditioned model to ensure a consistent shape).}
  \label{fig:lightingbias}
\end{figure}

%!TEX root = ../DiffusionRelightHint.tex

\section{Related Work}
\label{sec:related}

\paragraph{Diffusion Models for Image Generation}
Diffusion models have been shown to excel at the task of generating
high quality images by sampling from a learned distribution (e.g., of
photographs)~\cite{Song:2021:SBG,Karras:2022:EDS}, especially when
conditioned on
text-prompts~\cite{Nichol:2022:GTP,Ramesh:2022:HTC,Rombach:2022:HRI,Saharia:2022:PTI}.
Follow up work has endeavored to enrich text-driven diffusion models
to exert higher level semantic control over the image generation
process~\cite{Avrahami:2022:BDT,Brooks:2023:ILF,Ge:2023:ETI,Hertz:2022:PTP,Liu:2020:OEO,Mokady:2023:NTI,Tumanyan:2023:PPD,Voynov:2023:PET},
including non-rigid semantic edits~\cite{Cao:2023:MTF,Kawar:2023:MTR},
modifying the identity and gender of subjects~\cite{Kim:2022:DTG},
capturing the data distribution of underrepresented
attributes~\cite{Cong:2023:ACT}, and material
properties~\cite{Sharma:2023:APC}.  However, with the exception of
Alchemist~\cite{Sharma:2023:APC}, these methods only offer mid and
high level semantic control.  Similar to Alchemist, our method aims to
empower the user to control low level shading properties.
Complementary to Alchemist which offers relative control over material
properties such as translucency and gloss, our method provides
fine-grained control over the incident lighting in the generated
image.

Alternative guidance mechanisms have been introduced to provide
spatial control during the synthesis process based on (sketch, depth,
or stroke) images~\cite{Voynov:2023:SGT,Ye:2023:IAT,Meng:2022:SGI},
identity~\cite{Ma:2023:SDO,Xiao:2023:FTF,Ruiz:2023:HHF},
photo-collections~\cite{Ruiz:2023:DFT}, and by directly manipulating
mid-level
information~\cite{Ho:2021:CFD,Zhang:2023:ACC,Mou:2023:TAL}. However,
none of these methods provide control over the incident lighting.  We
follow a similar process and inject radiance hints modulated by a
neural encoded version of the image into the diffusion model via a
ControlNet~\cite{Zhang:2023:ACC}.

2D diffusion models have also been leveraged to change viewpoint or
generate 3D
models~\cite{Liu:2023:Z1T,Zhang:2023:DFP,Watson:2022:NVS,Xiang:2023:3IG}. However,
these methods do not offer control over incident lighting, nor
guarantee consistent lighting between viewpoints.
Paint3D~\cite{Zeng:2023:PPA} directly generates diffuse albedo
textures in the UV domain of a given mesh.
Fantasia3D~\cite{Chen:2023:FDG} and MatLaber~\cite{Xu:2023:MMT}
generate a richer set of reflectance properties in the form of shape
and spatially-varying BRDFs by leveraging text-to-image 2D diffusion
models and score distillation. Diffusion-based SVBRDF
estimation~\cite{Sartor:2023:MGD,Vecchio:2023:CAC} and diffusion-based
intrinsic decomposition~\cite{Kocsis:2023:IID} also produce rich
reflectance properties, albeit from a photograph instead of a
text-prompt.  However, all these methods require a rendering algorithm
to visualize the appearance, including indirect lighting and
shadows. In contrast, our method directly controls the lighting during
the sampling process, leveraging the space of plausible image
appearance embedded by the diffusion model.

\paragraph{Single Image Relighting}
While distinct, our method is related to relighting from a single
image, which is a highly underconstrained problem.  To provide
additional constraints, existing single image methods focus
exclusively on either outdoor
scenes~\cite{Wu:2017:IRI,Ture:2021:FNS,Yu:2020:SSO,Liu:2020:LFR,Griffiths:2022:OOS},
faces~\cite{Peers:2007:PFP,Wang:2008:FRS,Shu:2017:NFE,Sun:2019:SIP,Nestmeyer:2020:LPG,Pandey:2021:TRL,Han:2023:L3M,Ranjan:2023:FN3},
or human bodies~\cite{Kanamori:2018:RHO,Lagunas:2021:SIF,Ji:2022:GAS}.
In contrast, our method aims to offer fine-grained lighting control of
general objects.  Furthermore, existing methods expect a captured
photograph of an existing scene as input, whereas, importantly, our
method operates on, possibly implausible, generated images.  The vast
majority of prior single image relighting methods explicitly
disentangle the image in various components, that are subsequently
recombined after changing the lighting. In contrast, similar to
Sun~\etal~\shortcite{Sun:2019:SIP}, we forego explicit decomposition
of the input scene in disentangled components. However, unlike
Sun~\etal, we do not use a specially trained encoder-decoder model,
but rely on a general generative diffusion model to produce realistic
relit images.  Furthermore, the vast majority of prior single image
relighting methods represents incident lighting using a Spherical
Harmonics encoding~\cite{Ramamoorthi2002signal}.  Notable exceptions
are methods that represent the incident lighting by a shading image.
Griffiths~\etal~\shortcite{Griffiths:2022:OOS} pass a cosine weighted
shadow map (along with normals and the main light direction) to a
relighting network for outdoor scenes.  Similarly,
Kanamori~\etal~\shortcite{Kanamori:2018:RHO} and
Ji~\etal~\shortcite{Ji:2022:GAS} pass shading and ambient occlusion
maps to a neural rendering network.  To better model specular
reflections, Pandey~\etal~\shortcite{Pandey:2021:TRL} and
Lagunas~\etal~\shortcite{Lagunas:2021:SIF} pass, in addition to a
diffuse shading image, also one or more specular shading images for
neural relighting of human faces and full bodies respectively.  We
follow a similar strategy and pass the target lighting as a diffuse
and (four) specular radiance hints as conditions to a diffusion
model.

\paragraph{Relighting using Diffusion Models}
Ding~\etal\shortcite{Ding:2023:DLP} alter lighting, pose, and facial
expression by learning a CGI-to-real mapping from surface normals,
albedo, and a diffuse shaded 3D morphable model fitted to a single
photograph~\cite{Feng:2021:LAD}. To preserve the identity of the
subject in the input photograph, the diffusion model is refined on a
small collection ($\sim\!\!20$) of photographs of the subject.
Ponglertnapakorn~\etal\shortcite{Ponglertnapakorn:2023:DDF} leverage
off-the-shelf
estimators~\cite{Feng:2021:LAD,Deng:2019:AAA,Yu:2018:BBS} for the
lighting, a 3D morphable model, the subject's identity, camera
parameters, a foreground mask, and cast-shadows to train a conditional
diffusion network that takes a diffuse rendered model under the novel
lighting (blended on the estimated background), in addition to the
identity, camera parameters, and target shadows to generate a relit
image of the subject.  While we follow a similar overall strategy, our
method differs on three critical points.  First, our method operates
on general scenes which exhibit a broader range of shape and material
variations than faces.  Second, we provide multiple radiance hints
(diffuse and specular) to control the lighting during the diffusion
process. Finally, DiLightNet operates purely on an image generated via
a text-prompt and our method does not require a real-world captured
input photograph.

Lasagna~\cite{bashkirova2023lasagna} also shares the goal of
controlling the lighting in diffusion-based image generation. However,
instead of radiance hints, Lasagna uses language tokens to control the
lighting and thus lacks the fine-grained lighting control of
DiLightNet. Furthermore, it only supports a predefined set of $12$
directional lights while DiLightNet handles both point and
environmental lighting.

%!TEX root = ../DiffusionRelightHint.tex

%!TEX root = ../../DiffusionRelightHint.tex

\begin{figure*}
  \includegraphics[width=1.0\textwidth]{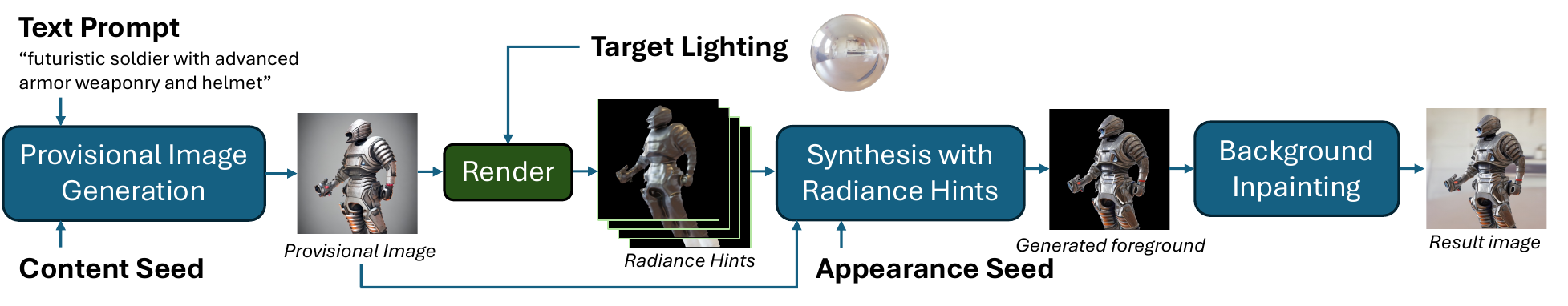}
  \caption{Overview of our pipeline for lighting-controlled
    prompt-driven image synthesis: (1) We start by generating a
    \emph{provisional image} using a pretrained diffusion model under
    uncontrolled lighting given a text prompt and a content-seed. (2)
    Next, we pass an appearance-seed, the provisional image, and a set
    of radiance hints (computed from the target lighting and a coarse
    estimate of the depth) to DiLightNet that will resynthesize the
    image such that becomes consistent with the target lighting while
    retaining the content of the provisional image. (3) Finally, we
    inpaint the background to be consistent with foreground object and
    the target lighting.}
  \label{fig:overview}
\end{figure*}

\section{Overview}
Our method takes as input a text prompt (describing the image
content), the target lighting, a content-seed that controls variations
in shape and texture, and an appearance-seed that controls variations
in light-material interactions.  The resulting output is a generated
image corresponding to the text prompt that is consistent with the
target lighting.  We assume that the image contains an isolated
foreground object, and that the background content is implicitly
described by the target lighting.  We make no assumption on the target
lighting, and support arbitrary lighting conditions.  Finally, while
we do not impose any constraint on the realism of the synthesized
content (e.g., fantastic beasts), we assume an image style that depicts
physically-based light-matter interactions (e.g., we do not support
artistic styles such as cell-shading or surrealistic images).

Our pipeline for lighting-controlled prompt-driven image synthesis
consists of three separate stages (\autoref{fig:overview}):
\begin{enumerate}
\item \emph{Provisional Image Generation:} In the first stage, we
  generate a provisional image with uncontrolled lighting given the
  text-prompt and the content-seed using a pretrained diffusion
  model~\cite{StableDiffusion}.  The goal of this stage is to
  determine the shape and texture of the foreground object.
  Optionally, we add \emph{``white background''} to the text-prompt to
  facilitate foreground detection.
\item \emph{Synthesis with Radiance Hints:} In the second stage
  (\autoref{sec:hints}), we first generate radiance hints given the
  provisional image and target lighting.  Next, the radiance hints are
  multiplied with a neural encoded version of the provisional image,
  and passed to DiLightNet together with the text-prompt and
  appearance-seed.  The result of this second stage is the foreground
  object with consistent lighting.
\item \emph{Background Inpainting:} In the third stage
  (\autoref{sec:background}), we inpaint the background consistent
  with the target lighting.
\end{enumerate} 

\section{Synthesis with Radiance Hints}
\label{sec:hints}

Our goal is to synthesize an image with the same foreground object as
in the provisional image, but with its appearance consistent with the
given target lighting.  We will finetune the same diffusion model used
to generate the provisional image to take in account the target
lighting via a ControlNet~\cite{Zhang:2023:ACC}.  A ControlNet assumes
a control signal per pixel, and thus we cannot directly guide the
diffusion model using a direct representation of the lighting such as
an environment map or a spherical harmonics encoding.  Instead, we
encode the \emph{effect} of the target lighting on each pixel's
outgoing radiance using radiance hints.

\subsection{Radiance Hint Generation}
\label{sec:radiancehint}
A radiance hint is a visualization of the target shape under the
target illumination, where the material of the object is replaced by a
homogeneous proxy material (e.g., uniform diffuse). However, we do not
have access to the shape of the foreground object.  To circumvent this
challenge, we observe that ControlNet typically does not require very
precise information and it has been shown to work well on sparse
signals such as sketches.  Hence, we argue that an approximate
radiance hint computed from a coarse estimate of the shape suffices.

To estimate the shape of the foreground object, we first segment the
foreground object from the provisional image using an off-the-shelf
salient object detection network.  Practically, we use
U2Net~\cite{Qin:2020:UGD} as it offers a good trade-off between speed
and accuracy; we revert to SAM~\cite{Kirillov:2023:SAM} for the rare
cases where U2Net fails to provide a clean foreground segmentation.
Next, we apply another off-the-shelf depth estimation network
(ZoeDepth~\cite{Bhat:2023:ZZS}) on the segmented foreground object.
The estimated depth map is subsequently triangulated in a mesh and
rendered under the target lighting with the proxy materials.  However,
single-image depth estimation is a challenging problem, and the
resulting triangulated depth maps are far from perfect.  Empirically
we find that ControlNet is less sensitive to low-frequency errors in
the resulting shading, while high-frequency errors in the shading can
lead to artifacts.  We therefore apply a Laplace smoothing filter
over the mesh to reduce the impact of high-frequency discontinuities.

Inspired by the positional encoding in
NeRFs~\cite{Mildenhall:2020:NRS}, we also encode the impact of
different frequencies in the target lighting on the appearance of the
foreground shape in separate radiance hints.  Leveraging the fact that
a BRDF acts as a band-pass filter on the incident lighting, we
generate $4$ radiance hints, each rendered with a different material
modeled with the Disney BRDF model~\cite{Burley:2012:PBS} (one pure
diffuse material and three specular materials with roughness set to
$0.34$, $0.13$, and $0.05$ respectively). We render the radiance
hints, inclusive of shadows and indirect lighting, with Blender's
Cycles path tracer.

%!TEX root = ../../DiffusionRelightHint.tex

\begin{figure}
  \centering
  \includegraphics[width=0.35\textwidth]{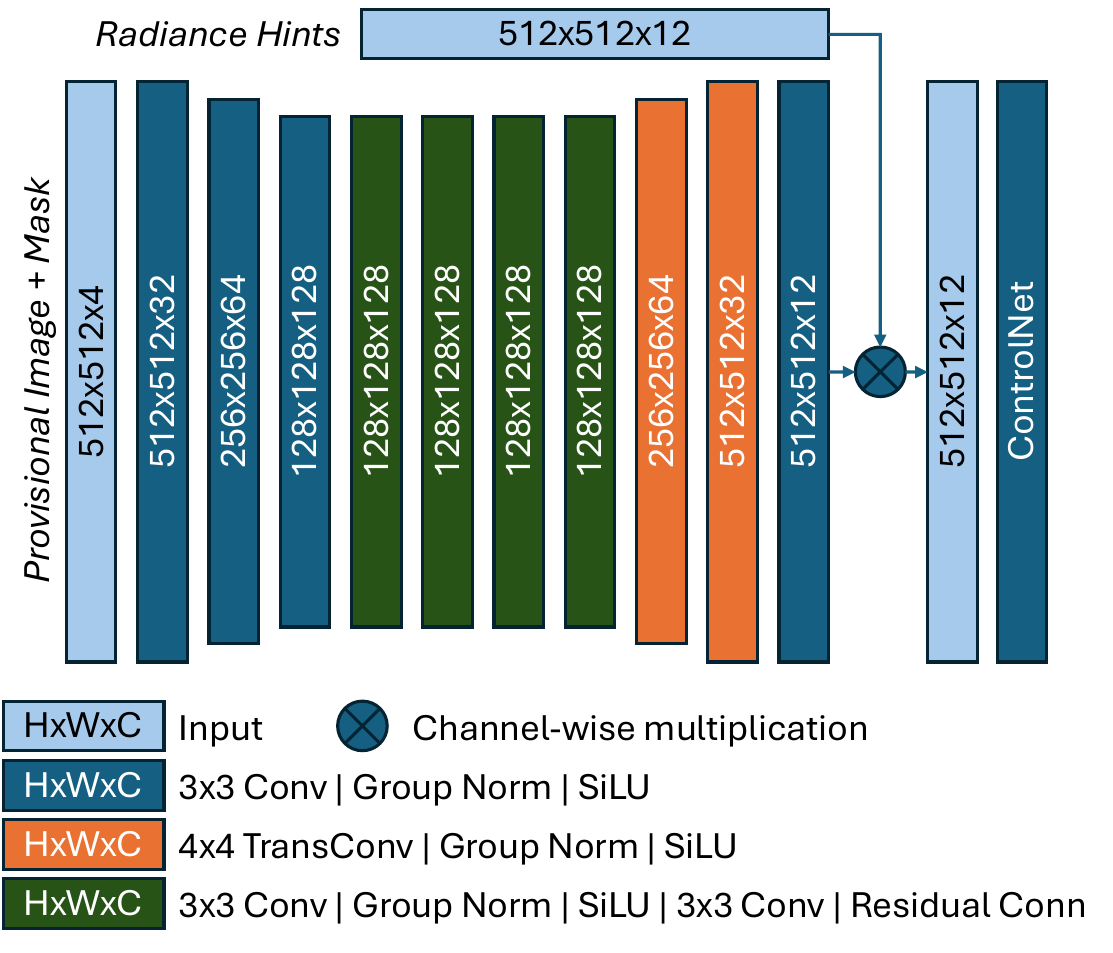}
  \caption{Provisional image encoder architecture. The output of the
    encoder is channel-wise multiplied with the radiance hints before
    passing the resulting $12$-channel feature map to a ControlNet.}
  \label{fig:network}
\end{figure}

\subsection{Lighting Conditioned ControlNet}
As noted before, we finetune a diffusion model to incorporate the
radiance hint images using ControlNet, as well as the original text
prompt used to generate the provisional image, and the
appearance-seed.  However, as we finetune the model, there is no
guarantee that it will generate a foreground object with the same
shape and texture as in the provisional image.  Therefore, we want to
include the provisional image into the diffusion process.  However,
the texture and shape information in the provisional image is
entangled with the unknown lighting from the first stage.  We
disentangle the relevant texture and shape information by first
encoding the provisional image (with the alpha channel set to the
segmentation mask).  Our encoder follows
Gao~\etal's~\shortcite{gao2020deferred} deferred neural relighting
architecture, but with a reduced number of channels to limit memory
usage. In addition, we include a channel-wise multiplication between
the $12$-channel encoded feature map of the provisional image and the
$4 \times 3$-channel radiance hints, which is subsequently passed to
ControlNet. The encoder architecture is summarized
in~\autoref{fig:network}.

\subsection{Training}
To train DiLightNet, we opt for a synthetic 3D training set that allows
us to precisely control the lighting, geometry, and the material
distributions.  It is critical that the synthetic training set
contains a wide variety of shapes, materials, and lighting.

\paragraph{Shape and Material Diversity}
We select synthetic objects from the LVIS category in the Objaverse
dataset~\cite{objaverse} that also have either a roughness map, a
normal map, or both, yielding an initial subset of $13K$ objects.
In addition, we select $4K$ objects from the Objaverse dataset (from
the LVIS category) that only contain a diffuse texture map and assign
a homogeneous specular BRDF with a roughness log-uniformly selected in
$[0.02, 0.5]$ and specular tint set to $1.0$.  To ensure that the
refined diffusion model has seen objects with homogeneous materials,
we select an additional $4K$ objects (from the LVIS category) and
randomly assign a homogeneous diffuse albedo and specular roughness
sampled as before.

Empirically, we found that the diversity of detailed spatially varying
materials in the Objaverse dataset is limited. Therefore, we further
augment the dataset with the shapes with the most ``likes'' (a
statistic provided by the Objaverse dataset) from each LVIS category.
For each of these selected shapes we automatically generate UV
coordinates using Blender (we eliminate the shapes ($17$) for which
this step failed), and create $4$ synthetic objects per shape by
assigning a randomly selected spatially varying material from the
INRIA-Highres SVBRDF dataset~\cite{Deschaintre:2020:GFT}, yielding a
total of $4K$ additional objects with enhanced materials.

In total, our training set contains $25K$ synthetic objects with a wide
variety of shapes and materials.  We scale and translate each object
such that its bounding sphere is centered at the origin with a radius
of 0.5m.

\paragraph{Lighting Diversity}
We consider five different lighting categories:
\begin{enumerate}
\item \emph{Point Light Source} random uniformly sampled on the upper
  hemisphere (with $0 \leq \theta \leq 60^\circ$) surrounding the
  object with radius sampled in $[4m, 5m]$, and with the power
  uniformly chosen in $[500W, 1500W]$.  To avoid completely black
  images when the point light is positioned behind the object, we also
  add a $1W$ uniform white environment light.
\item \emph{Multiple Point Light Sources:} three light sources sampled
  in the same manner as the single light source case, including the
  uniform environment lighting.
\item \emph{Environment Lighting} sampled from a collection of $679$
  environment maps from Polyhaven.com.
\item \emph{Monochrome Environment Lighting} are the luminance only
  versions of the environment lighting category. Including this
  category combats potential inherent biases in the overall color
  distribution in the environment lighting.
\item \emph{Area Light Source} simulates studio setups with large
  light boxes.  We achieve this by randomly placing an area light
  source on the hemisphere surrounding the object (similar to point
  light sources) aimed at the object, with a size randomly chosen in
  the range $[5m, 10m]$ and total power sampled in $[500W,
  1500W]$. Similar to the point lighting, we add a uniform white
  environment light of $1W$.
\end{enumerate}

\paragraph{Rendering}
We render each of the $25K$ synthetic objects from four viewpoints
uniformly sampled on the hemisphere with radius uniformly sampled from
$[0.8m, 1.1m]$ and $10^\circ \leq \theta \leq 90^\circ$, aimed at the
object with a field of view sampled from $[25^\circ, 30^\circ]$, and
lit with $12$ different lighting conditions, selected with a relative
ratio of $3\!:\!1\!:\!3\!:\!2\!:\!3$ for point source lighting, multiple point
sources, environment maps, monochrome environment maps, and area light
sources respectively. For each rendered viewpoint, we also require
corresponding radiance hints. However, at \emph{evaluation} time, the
radiance hints will be constructed from estimated depth maps; using
the ground truth geometry and normals during \emph{training} would
therefore introduce a domain gap.  We observe that depth-derived
radiance hints include two types of approximations.  First, due to the
smoothed normals, the resulting shading will also be smoothed and
shading effects due to intricate geometrical details are lost; i.e.,
it locally affects the radiance hints.  Second, due to the ambiguities
in estimating depth from a single image, missing geometry and global
deformations cause incorrect shadows; i.e., a non-local effect.  We
argue that diffusion models can plausibly correct the former, whereas
the latter is more ambiguous and difficult to correct.  Therefore, we
would like the training radiance hints to only introduce
approximations on the local shading.  This is achieved by using the
ground truth geometry with modified shading normals.  We consider two
different approximations for the shading normals, and randomly select
at training time which one to use: (1) we use the geometric normals
and ignore any shading normals from the object's material model, or
(2) we use the corresponding normals from the smoothed triangulated
depth (to reduce computational costs, we estimate the depth for each
synthetic object for each viewpoint under uniform white lighting
instead for each of the $9$ sampled lighting conditions).

\paragraph{Training Dataset}
At training time we dynamically compose the input-output pairs. We
first select a synthetic object and view uniformly. Next, we select
the lighting for the input and output image. To select the lighting
condition for the input training image, we note that images generated
with diffusion models tend to be carefully white balanced.  Therefore,
we exclude the input images rendered under (colored) environment
lighting.  For the output image, we randomly select any of the $12$
precomputed renders (including those rendered with colored environment
lighting). We select the radiance hints corresponding to the output
with a 1:9 ratio for the radiance hints with smoothed depth-estimated
normals versus geometric normals.  To further improve robustness with
respect to colored lighting, we apply an additional color augmentation
to the output images by randomly shuffling their RGB color channels;
we use the same color channel permutation for the output image and its
corresponding radiance hints.

\section{Background Inpainting}
\label{sec:background}

\paragraph{Environment-based Inpainting}
When the target lighting is specified by an environment map, we can
directly render the background image using the same camera
configuration as for the radiance hints.  We composite the foreground
on the background using the previously computed segmentation mask
filtered with a $3 \times 3$ average filter to smooth the mask edges.

\paragraph{Diffusion-based Inpainting}
For all other lighting conditions, we use a pretrained diffusion-based
inpainting model~\cite{Rombach:2022:HRI} (i.e., the
\emph{stable-diffusion-2-inpainting}
model~\cite{StableDiffusionInpainting}). We input the synthesized
foreground image along with the (inverse) segmentation mask, as well
as the original text prompt, to complete the foreground image with a
consistent background.

%!TEX root = ../../DiffusionRelightHint.tex

\newcommand{\mainResFullFigWidth}{0.167\textwidth}

\begin{figure*}
\renewcommand{\arraystretch}{0.35}
\addtolength{\tabcolsep}{-6.0pt}
 \begin{tabular}{ cccccc }
  \includegraphics[width=\mainResFullFigWidth]{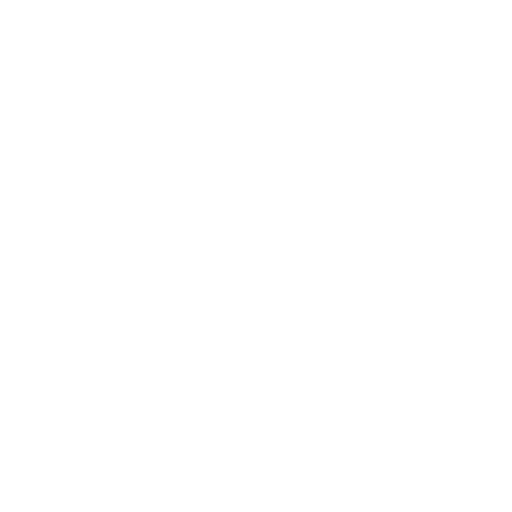}
  &\includegraphics[width=\mainResFullFigWidth]{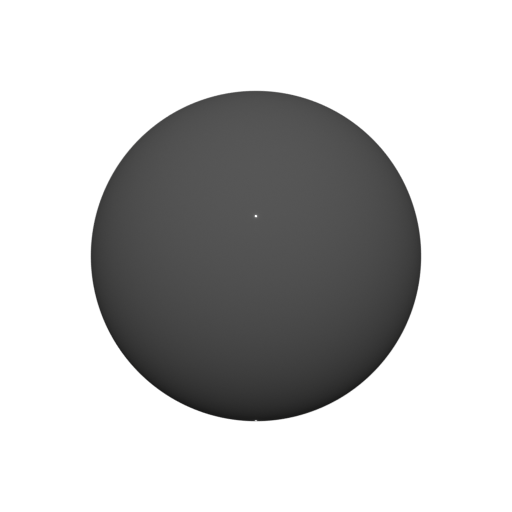}
  &\includegraphics[width=\mainResFullFigWidth]{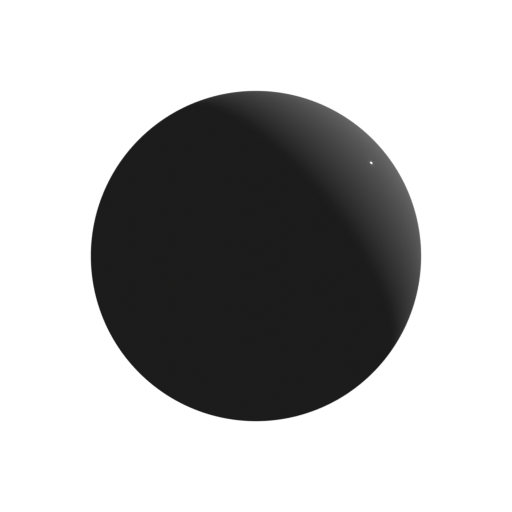}
  &\includegraphics[width=\mainResFullFigWidth]{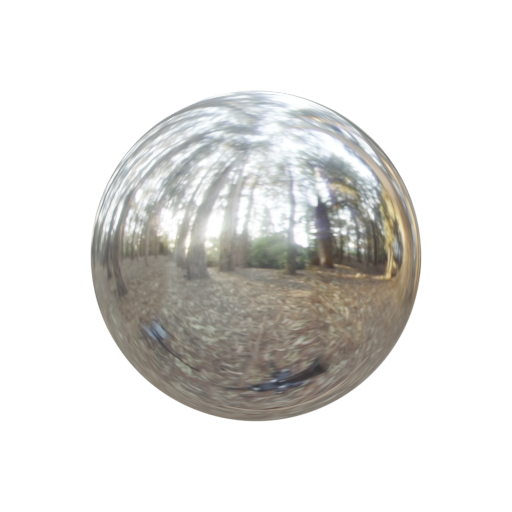}
  &\includegraphics[width=\mainResFullFigWidth]{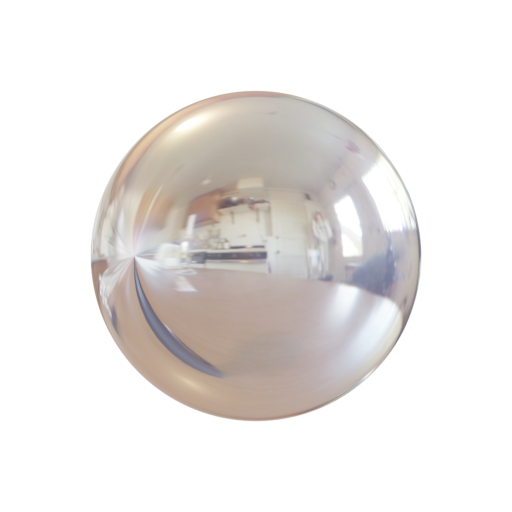}
  &\includegraphics[width=\mainResFullFigWidth]{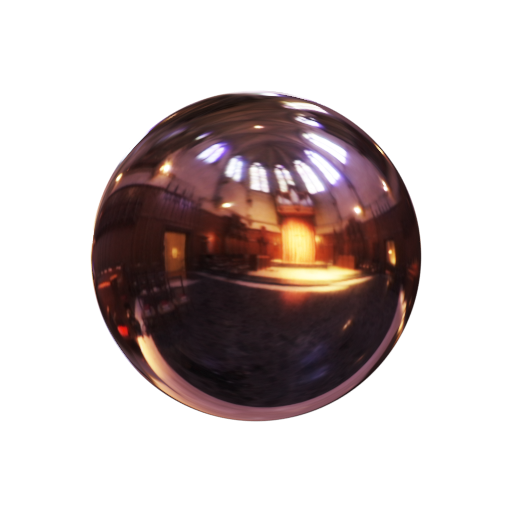}\\

  \includegraphics[width=\mainResFullFigWidth]{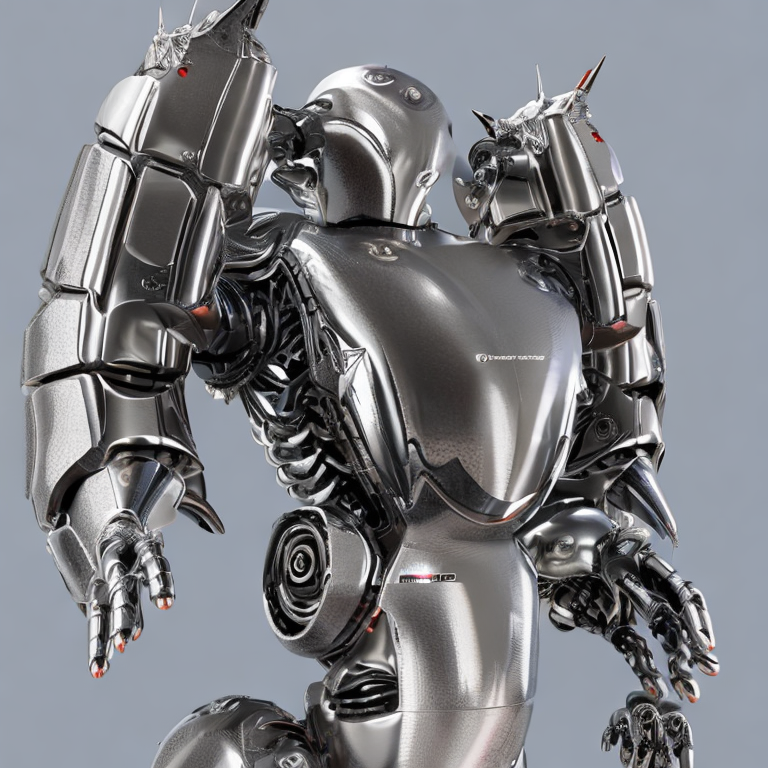}
 &\includegraphics[width=\mainResFullFigWidth]{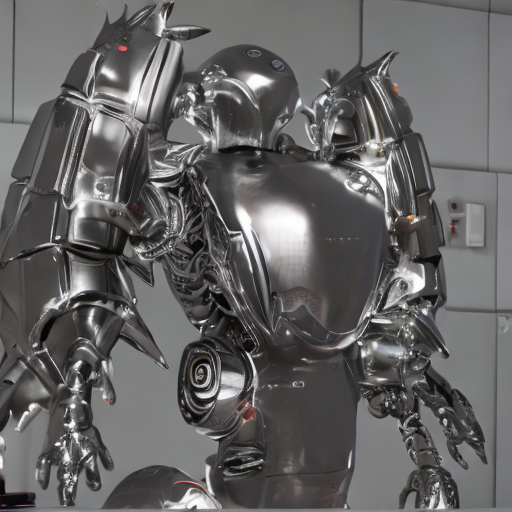}
 &\includegraphics[width=\mainResFullFigWidth]{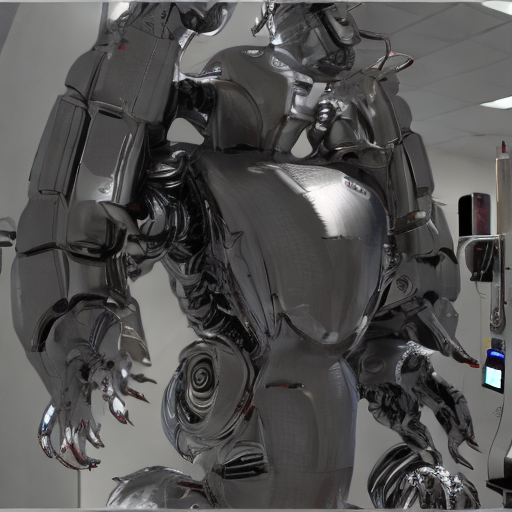}
 &\includegraphics[width=\mainResFullFigWidth]{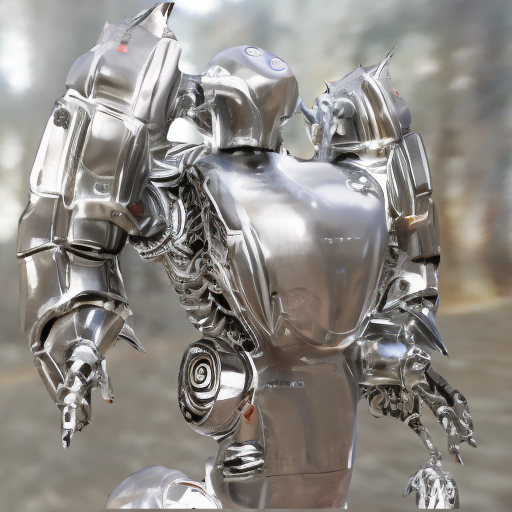}
 &\includegraphics[width=\mainResFullFigWidth]{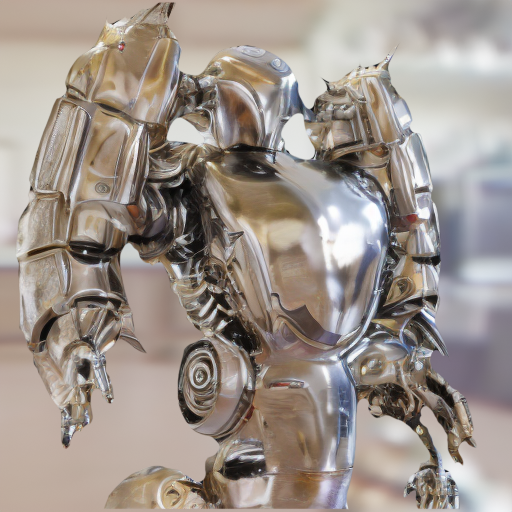}
 &\includegraphics[width=\mainResFullFigWidth]{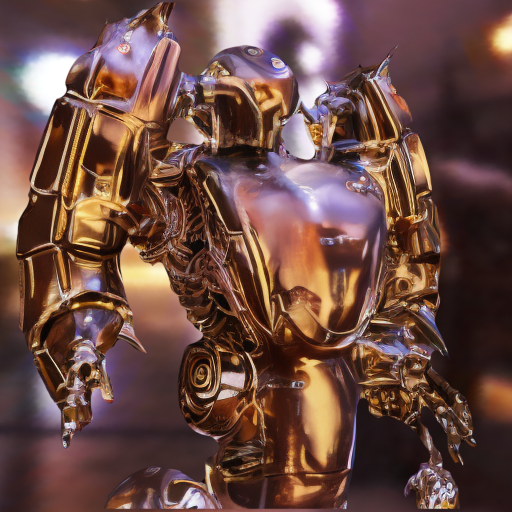}\\
  \multicolumn{6} {c} {
    Prompt: \emph{``machine dragon robot in platinum''}.
    }\\
 \includegraphics[width=\mainResFullFigWidth]{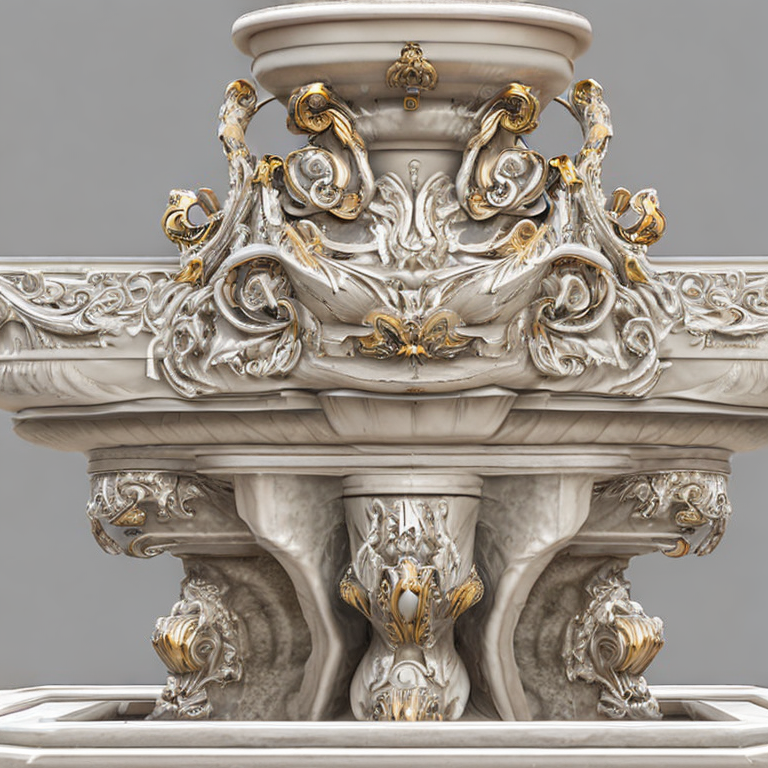}
 &\includegraphics[width=\mainResFullFigWidth]{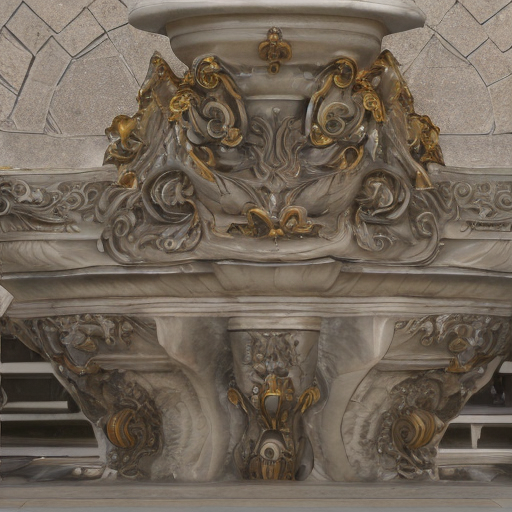}
 &\includegraphics[width=\mainResFullFigWidth]{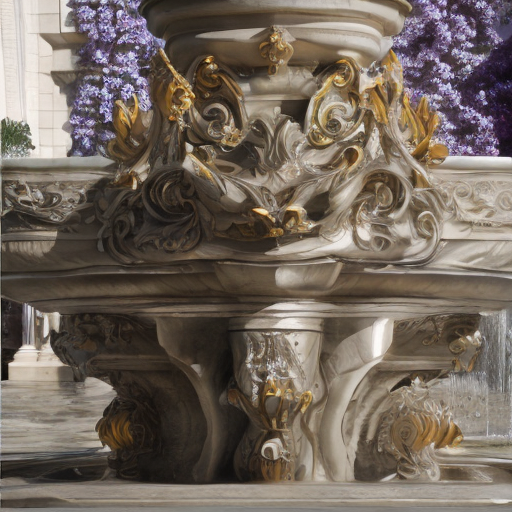}
 &\includegraphics[width=\mainResFullFigWidth]{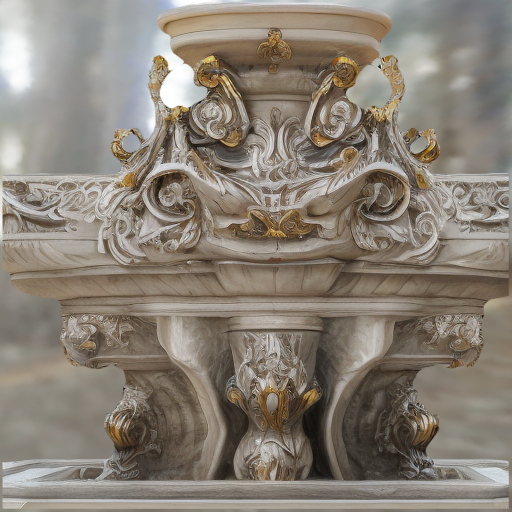}
 &\includegraphics[width=\mainResFullFigWidth]{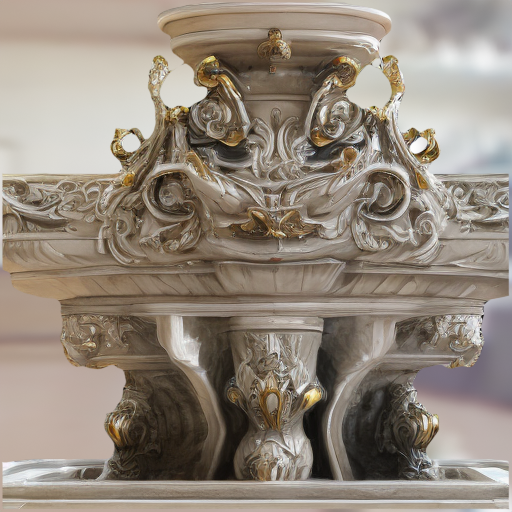}
 &\includegraphics[width=\mainResFullFigWidth]{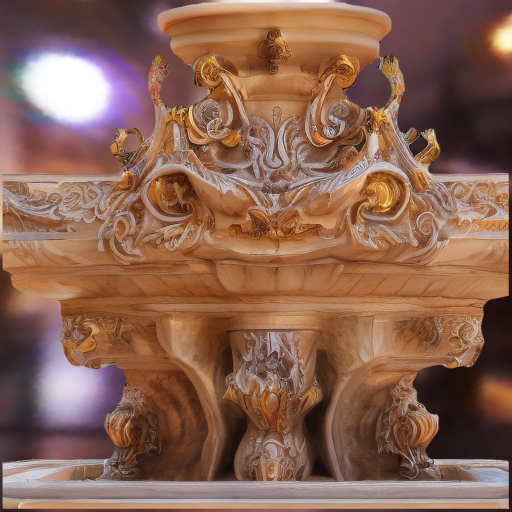}\\
  \multicolumn{6} {c} {
    Prompt: \emph{``gorgeous ornate fountain made of marble''}.
    }\\
  \includegraphics[width=\mainResFullFigWidth]{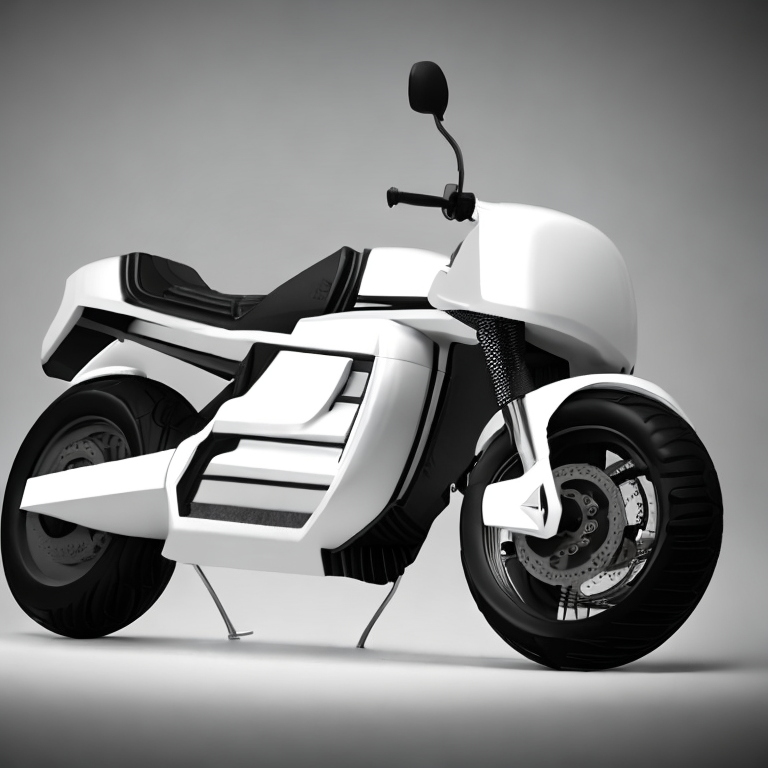}
 &\includegraphics[width=\mainResFullFigWidth]{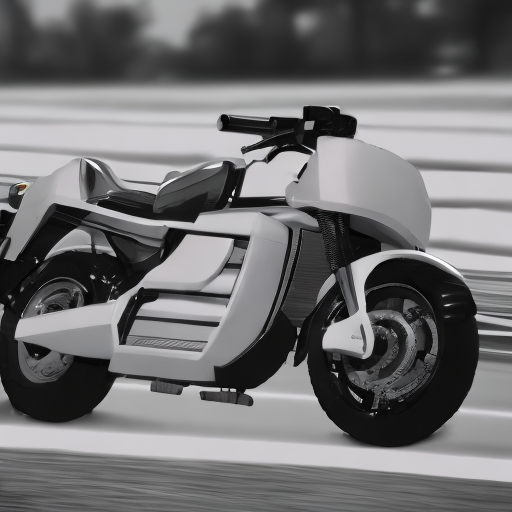}
 &\includegraphics[width=\mainResFullFigWidth]{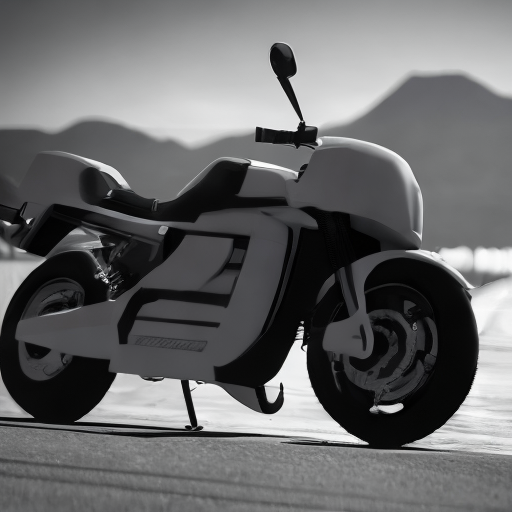}
 &\includegraphics[width=\mainResFullFigWidth]{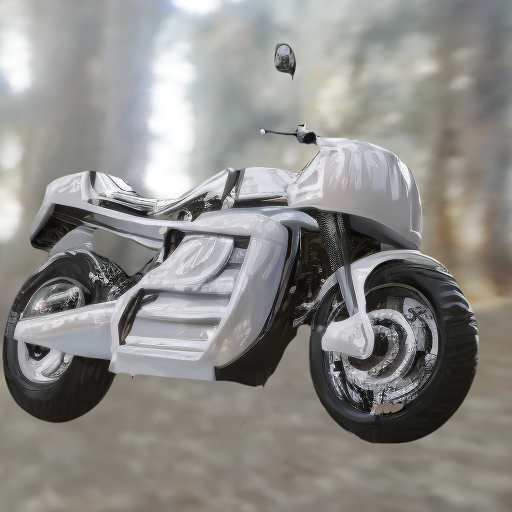}
 &\includegraphics[width=\mainResFullFigWidth]{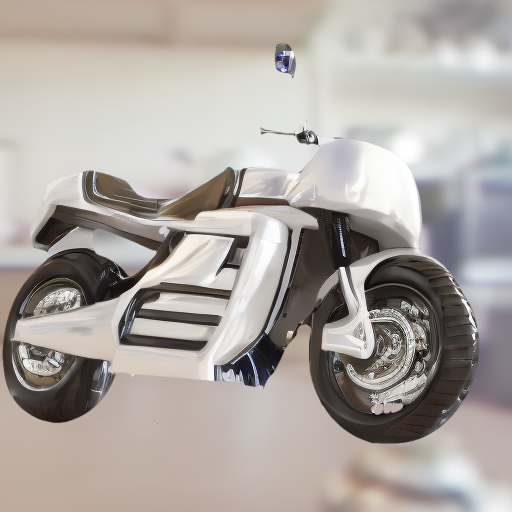}
 &\includegraphics[width=\mainResFullFigWidth]{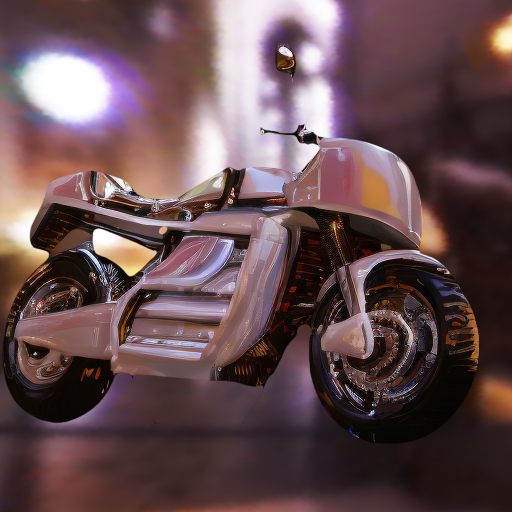}\\
    \multicolumn{6} {c} {
    Prompt: \emph{``Storm trooper style motorcycle''}.
    }\\
   \includegraphics[width=\mainResFullFigWidth]{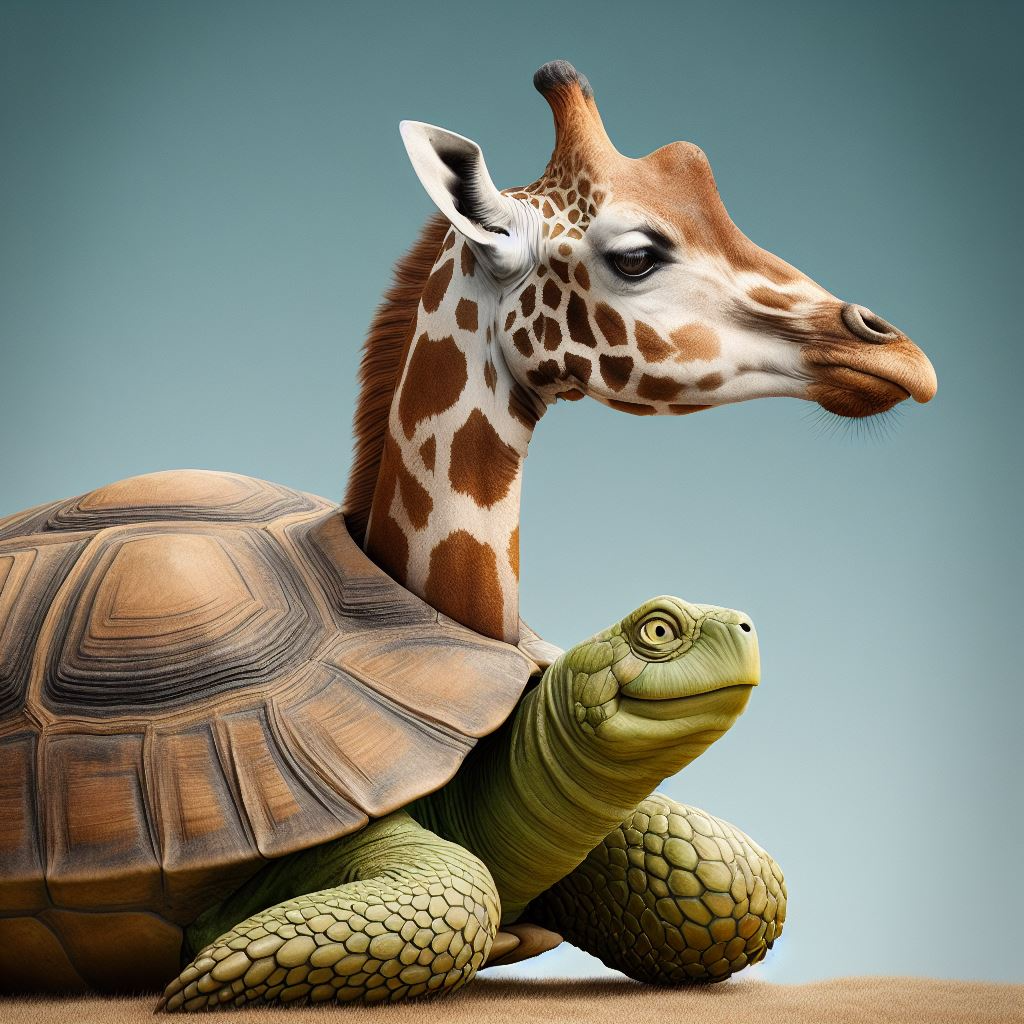}
 &\includegraphics[width=\mainResFullFigWidth]{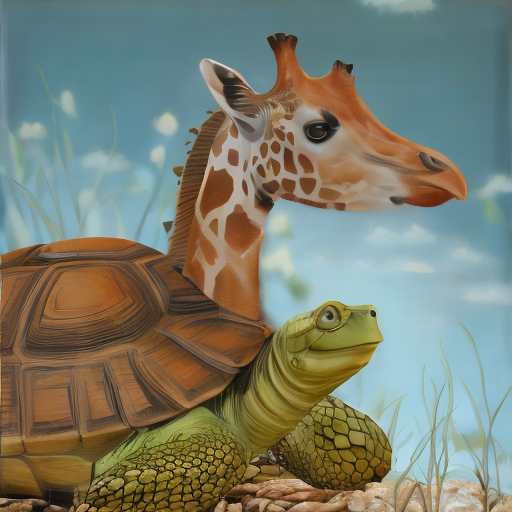}
 &\includegraphics[width=\mainResFullFigWidth]{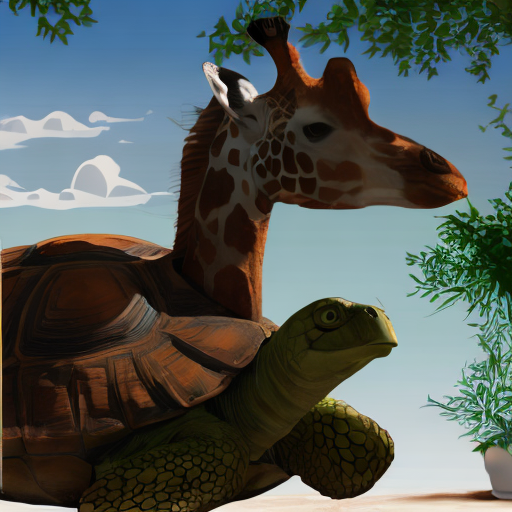}
 &\includegraphics[width=\mainResFullFigWidth]{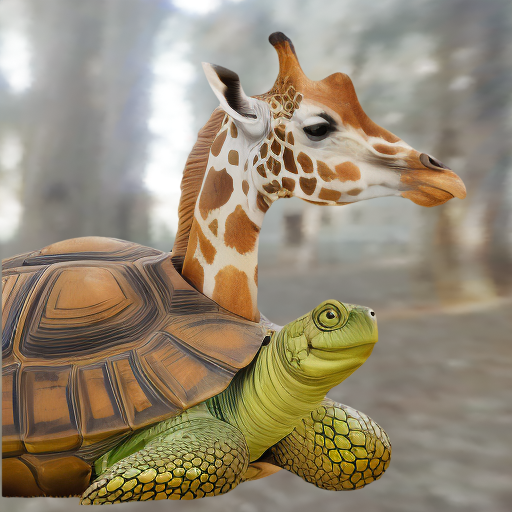}
 &\includegraphics[width=\mainResFullFigWidth]{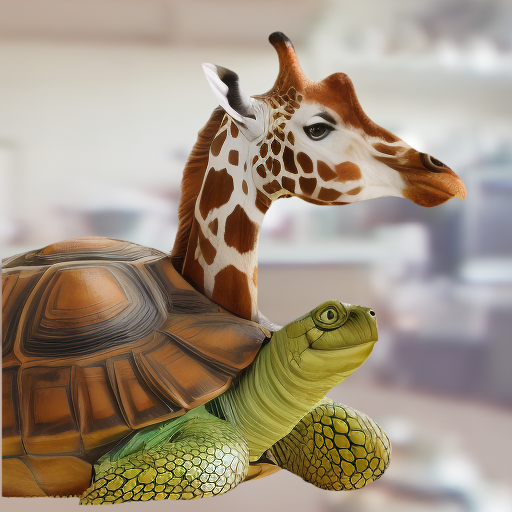}
 &\includegraphics[width=\mainResFullFigWidth]{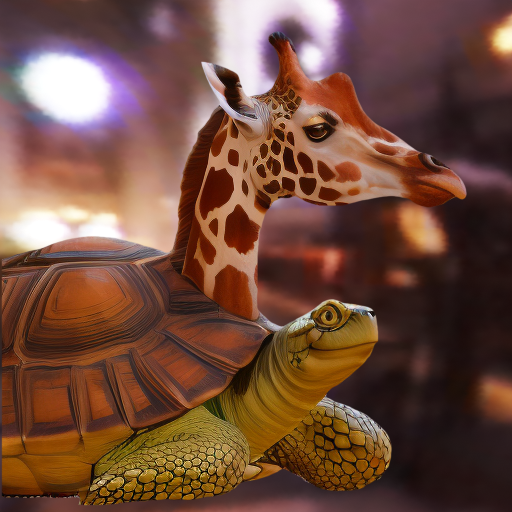}\\
   \multicolumn{6} {c} {
    Prompt: \emph{``A giraffe imitating a turtle, photorealistic''}.
    }\\

    \includegraphics[width=\mainResFullFigWidth]{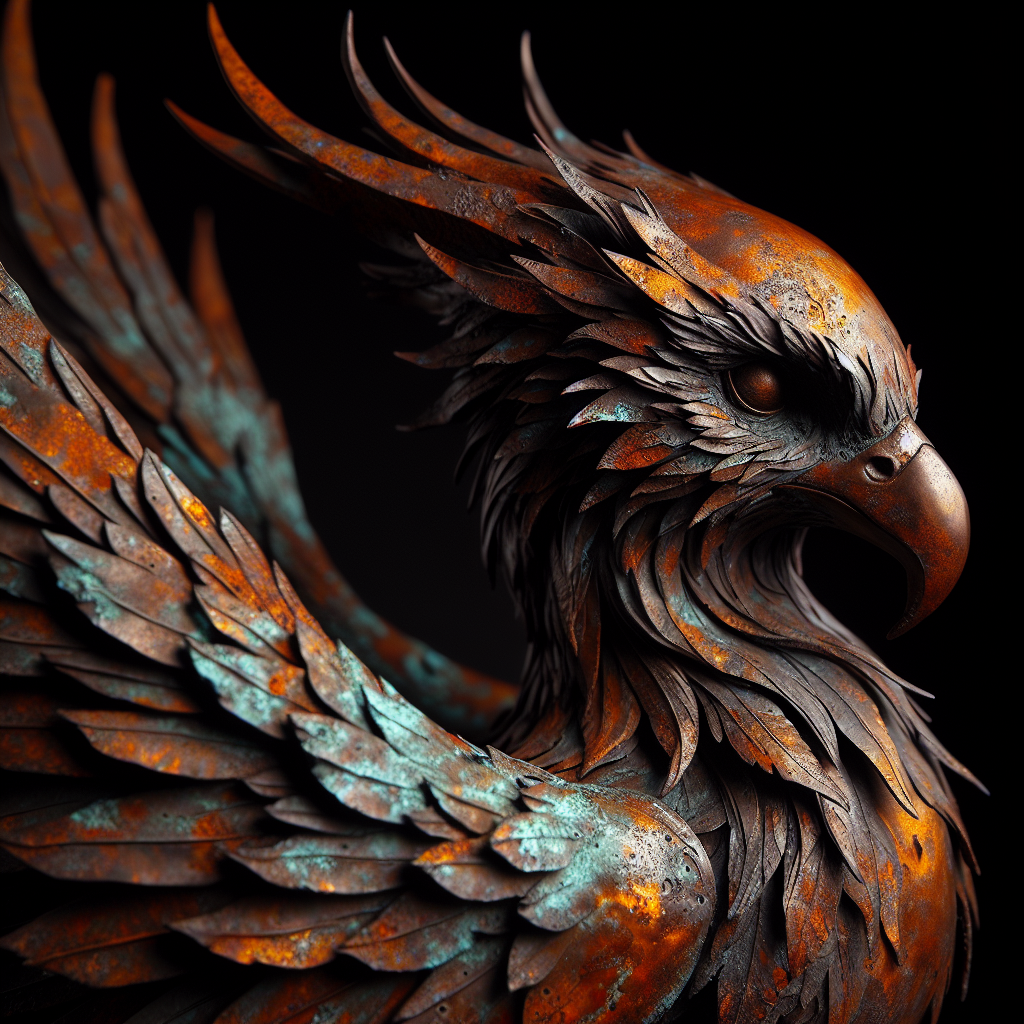}
 &\includegraphics[width=\mainResFullFigWidth]{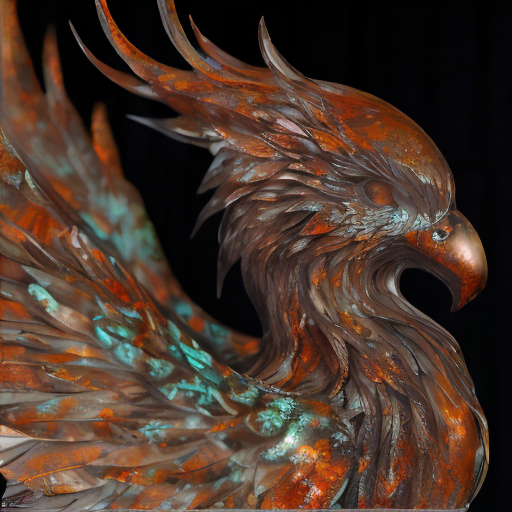}
 &\includegraphics[width=\mainResFullFigWidth]{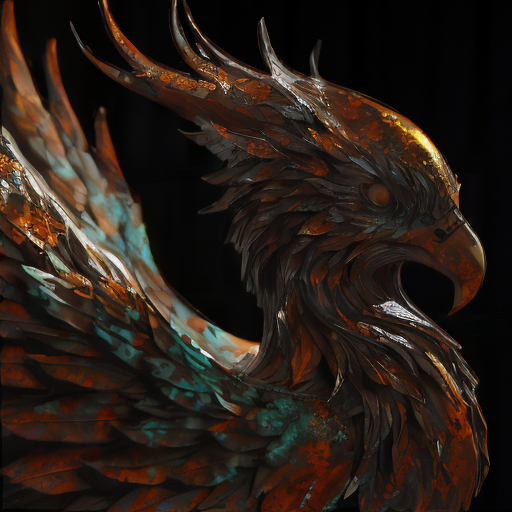}
 &\includegraphics[width=\mainResFullFigWidth]{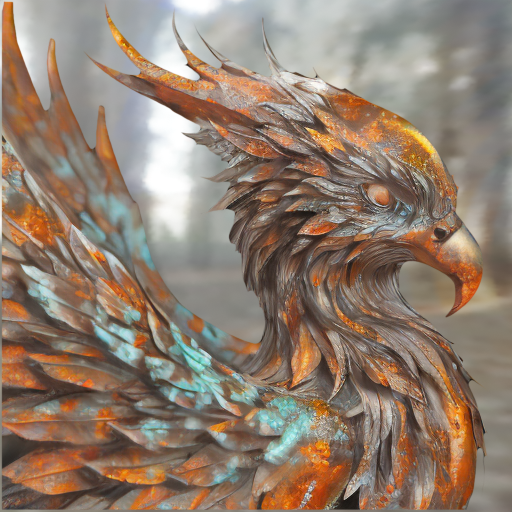}
 &\includegraphics[width=\mainResFullFigWidth]{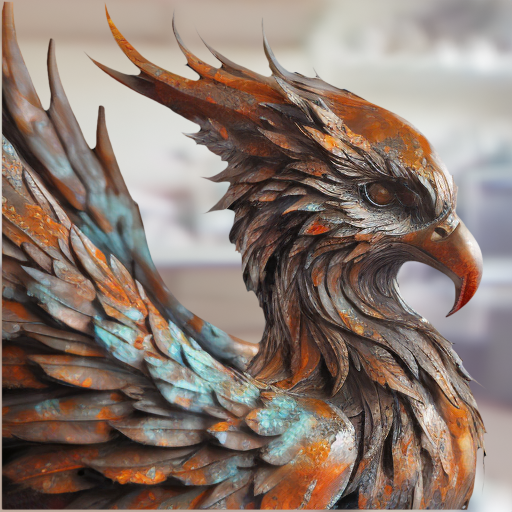}
 &\includegraphics[width=\mainResFullFigWidth]{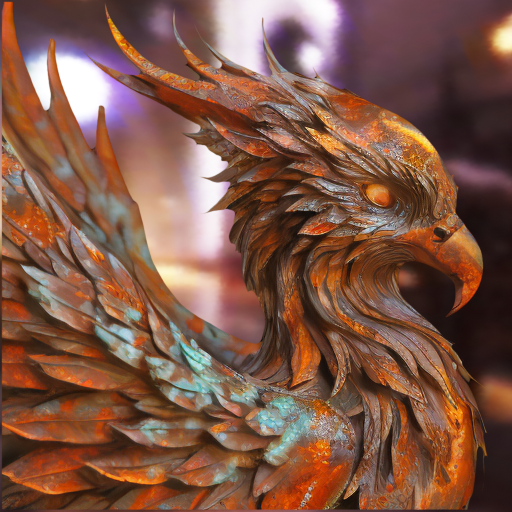}\\
   \multicolumn{6} {c} {
    Prompt: \emph{``Rusty sculpture of a phoenix with its head more polished yet the wings are more rusty''}.
    }
 \end{tabular} 
 \caption{Text-to-image generated results with lighting control. The
   first column shows the provisional image as a reference, whereas
   the last five columns are generated under different user-specified
   lighting conditions (point lighting (columns 2-3) and environment
   lighting (columns 4-6)).  The provisional images for the last two
   examples are generated with \emph{DALL-E3} instead of \emph{stable
     diffusion v2.1} to better handle the more complex prompt.}
  \label{fig:text2img_main}
\end{figure*}

%!TEX root = ../DiffusionRelightHint.tex

\section{Results}
\label{sec:results}

We implemented DiLightNet in PyTorch~\cite{Paszke:2019:Pytorch} and
use \emph{stable diffusion v2.1}~\cite{StableDiffusion} as the base
pretrained diffusion model to refine.  We jointly train the
provisional image encoder as well as the ControlNet using
AdamW~\cite{Loshchilov:2018:DWD} with a $10^{-5}$ learning rate (all
other hyper-parameter are kept at the default values) for $150K$
iterations using a batch size of $64$. Training took approximately $30$
hours using $8 \times$ NVidia V100 GPUs. The training data is rendered
using Blender's Cycles path tracer~\cite{Blender} at $512 \times 512$
resolution with $4096$ samples per pixel.

\paragraph{Consistent Lighting Control}
\autoref{fig:text2img_main} shows five generated scenes (the
provisional image is shown in the first column for reference) under
$5$ different lighting conditions (point light (2nd and 3rd column),
and 3 different environment maps from~\cite{Debevec1998}: Eucalyptus
Grove (4th column), Kitchen (5th column), and Grace Cathedral (last
column)) for five different prompts. Each prompt was chosen to
demonstrate our method's ability to handle different material and
geometric properties such high specular materials (1st row), rich
geometrical details (2nd row), objects with multiple homogeneous
materials (3rd row), non-realistic geometry (4th row), and
spatially-varying materials (last row).  The provisional image in the
last two rows are generated with \emph{DALL-E3} instead of
\emph{stable diffusion v2.1} to better model the more complex prompt.
We observe that DiLightNet produces plausible results and that the
appearance is consistent under the same target lighting for different
prompts.  Furthermore, the lighting changes are plausible over each
prompt.  Please refer to the supplemental material for additional
results. 

%!TEX root = ../../DiffusionRelightHint.tex

\newcommand{\seedFigWidth}{0.2\textwidth}
\begin{figure*}
\centering
\renewcommand{\arraystretch}{0.25}
\addtolength{\tabcolsep}{-6.0pt}
 \begin{tabular}{ ccccc }
 \includegraphics[width=\seedFigWidth]{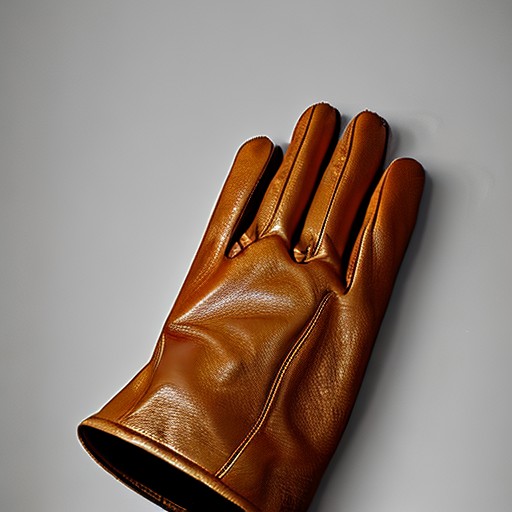}
 &
 \includegraphics[width=\seedFigWidth]{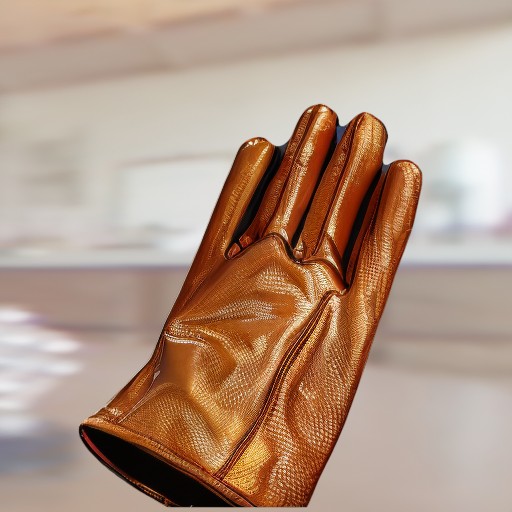}
 &
 \includegraphics[width=\seedFigWidth]{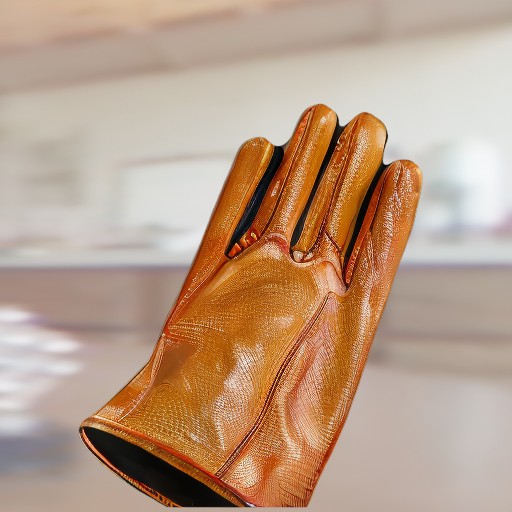}
 &
 \includegraphics[width=\seedFigWidth]{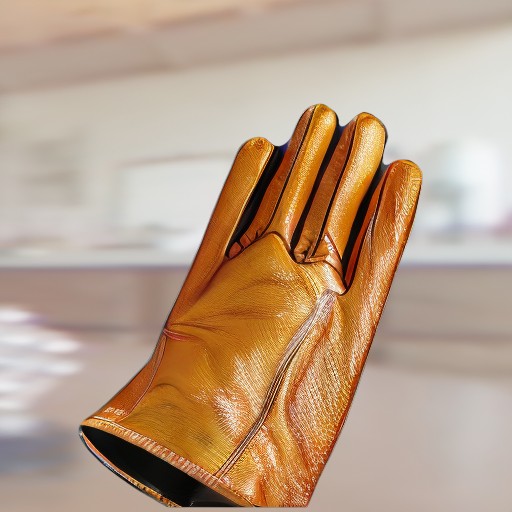}
 &
 \includegraphics[width=\seedFigWidth]{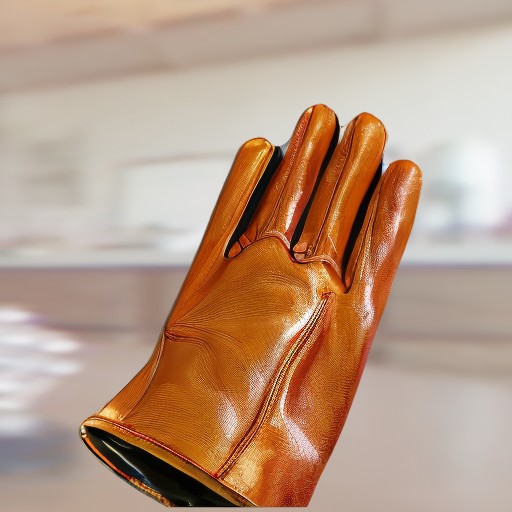}\\
 %Provisional image & Seed A & Seed B & Seed C & Seed D\\
 \end{tabular} 
 \caption{Impact of changing the appearance-seed. If not sufficiently
   constrained by the text prompt, the generated provisional image
   (left) might not provide sufficient information for DiLightNet to
   determine the exact materials of the object.  Altering the
   appearance-seed directs DiLightNet to sample a different
   interpretation of light-matter interaction in the provisional
   image. In this example, altering the appearance-seed induces
   changes in the interpretation of the glossiness and smoothness of
   the leather gloves.}
  \label{fig:seeds}
\end{figure*}

%!TEX root = ../../DiffusionRelightHint.tex

\newcommand{\materialpromptFigWidth}{0.2\textwidth}
\begin{figure*}
\renewcommand{\arraystretch}{0.25}
\addtolength{\tabcolsep}{-6.0pt}
 \begin{tabular}{ ccccc }
  \includegraphics[width=\materialpromptFigWidth]{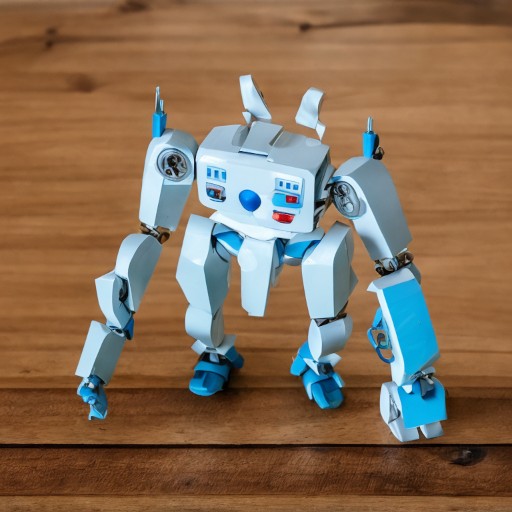}
 &\includegraphics[width=\materialpromptFigWidth]{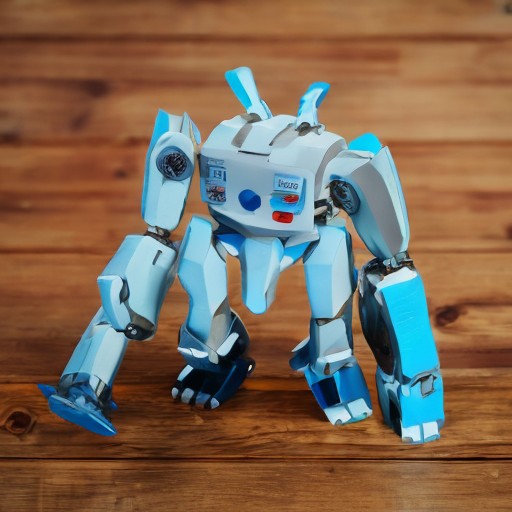}
 &\includegraphics[width=\materialpromptFigWidth]{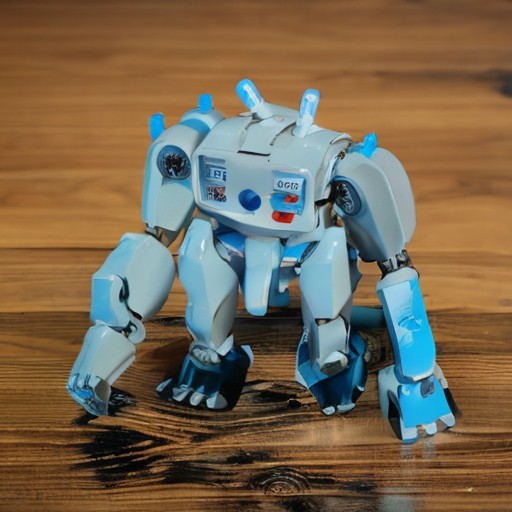}
 &\includegraphics[width=\materialpromptFigWidth]{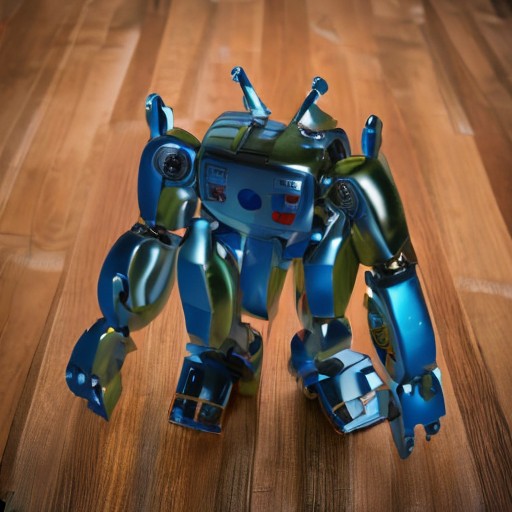}
 &\includegraphics[width=\materialpromptFigWidth]{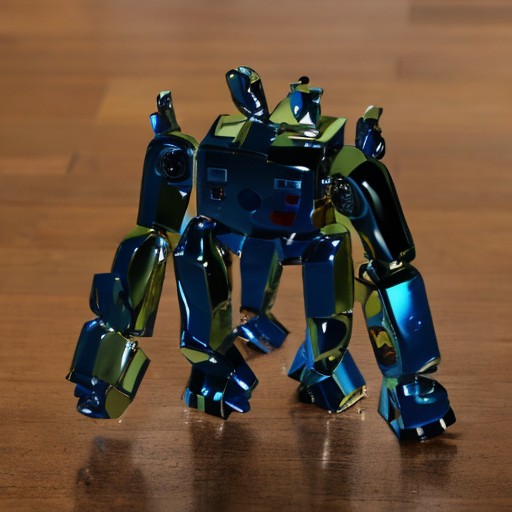}\\
  \footnotesize{Provisional image} & \footnotesize{\emph{"paper made"}} & \footnotesize{\emph{"plastic"}} & \footnotesize{\emph{"specular shinny metallic"}} & \footnotesize{\emph{"mirror polished metallic"}}
 \end{tabular} 
 \caption{Impact of prompt specialization in DiLightNet. Instead of
   altering the appearance-seed, the user can also specialize the
   prompt with additional material information in the 2nd stage. In
   this example the initial prompt (\emph{``toy robot''}) is augmented
   with additional material descriptions while keeping the (point
   lighting) fixed.}
  \label{fig:material_prompt}
\end{figure*}

\paragraph{Additional User Control}
One advantage of our three step solution is that the user can alter
the appearance-seed in the second stage to modify the interpretation
of the materials in the provisional image.  \autoref{fig:seeds}
showcases how different appearance-seeds affect the generated
results. Altering the appearance-seed yields alternative explanations
of the appearance in the provisional image. Conversely, using the same
appearance-seed produces a consistent appearance under different
controlled lighting conditions (as demonstrated
in~\autoref{fig:text2img_main}).

In addition to the appearance-seed, we can further specialize the text
prompt between the first and second stage to provide additional
guidance on the material properties.  \autoref{fig:material_prompt}
shows four specializations of an initial prompt (\emph{``toy robot''})
by adding: \emph{``paper made''}, \emph{``plastic''}, \emph{``specular
  shinny metallic''}, and \emph{``mirror polished metallic''}.  From
these results we can see that all variants are consistent under the
same lighting, but with a more constrained material appearance (i.e.,
diffuse without a highlight, a mixture of diffuse and specular,
and two metallic surfaces with a different roughness).

\paragraph{User Study}
We perform two user studies to measure the perceptual lighting
accuracy and the consistency of the resulting appearance under varying
lighting; i.e., how well changes induced by the target lighting are
disentangled from the appearance-seed.

In the first study, participants rate the lighting similarity of the
foreground objects in image pairs (four-level rating range where 0
means least similar and 3 means most similar) selected from three
groups of image pairings (10 pairs in each group):
\begin{enumerate}
\item a synthetic object rendered under the target lighting is paired
  with any of the generated images shown in this paper and the
  supplemental material under identical lighting;
\item a pair of synthetic objects rendered under identical target
  lighting (this serves as the positive baseline); and
\item a synthetic image paired with a generated image without lighting
  control (the negative baseline). To avoid that the background
  affects the judgment, we replace the background with the target
  environment lighting.
\end{enumerate}
The average total rating over 20 non-expert participants with images
shown in randomized order for each of the three classes is:
$19.61/19.85/12.25$, showing that DiLightNet scores similar to the
positive reference.

In a second study, participants rate the appearance consistency of the
foreground objects in image pairs generated with rotated environment
lighting. We opt for rotating the lighting to retain the overall color
balance and frequency of lighting. The three groups of pairings under
rotated lighting are:
\begin{enumerate}
  \item image pairs generated with the same prompt
    and seeds;
  \item image pairs rendered with the same synthetic object
    (positive baseline); and
  \item pairs generated without lighting control with the same
    text prompt but different content-seeds (negative baseline).
\end{enumerate}
The average total rating was $25.75/25.05/11.35$, confirming appearance
consistency on par with the positive baseline.

\section{Ablation Study}
\label{sec:ablation}

We perform a series of qualitative and quantitative ablation studies
to better understand the impact of the different components that
comprise our method.  For quantitative evaluation, we create a
synthetic test set by selecting objects from the Objaverse dataset
that have the 'Staff Picked' label and \emph{no} LVIS label, ensuring
that there is no overlap between the training and test set.  To ensure
high quality synthetic objects, we manually remove scenes that are not
limited to a single object and/or objects with low quality scanned
textures with baked in lighting effects, yielding a test set of $50$
high quality synthetic objects.  We render each test scene for $3$
viewpoints and $6$ lighting conditions.  We quantify errors with
the PSNR, SSIM, and LPIPS~\cite{Zhang:2018:LPIPS} metrics. Because the
appearance-seed is a user controlled parameter, we assume that the
user would select the appearance-seed that produces the most plausible
result.  To simulate this process, we report the errors for each
scene/view/lighting combination that produces the lowest LPIPS errors
on renders generated with $4$ different appearance-seeds.

\paragraph{Provisional Image Encoding}
DiLightNet multiplies the (encoded) provisional image with the
radiance hints.  We found that both the encoding, as well as the
multiplication is critical for obtaining good results.
\autoref{fig:ablation_visual} shows a comparison of DiLightNet versus
two alternate architectures:
\begin{enumerate}
\item \emph{Direct ControlNet} passes the provisional image directly
  as an additional channel (in addition to the radiance hints) instead
  of multiplying, yielding 16 channels input for ControlNet
  (3-channels for the provisional image, plus ($4 \times 3$)-channels
  for the radiance hints, and $1$ channel for the mask); and
\item \emph{Non-encoded Multiplication} of the provisional image
  (without encoding) with the radiance hints.
\end{enumerate}
Neither of the variants generates satisfactory results.  This
qualitative result is further quantitatively confirmed
in~\autoref{tab:ablation} (rows 1-3).

%!TEX root = ../../DiffusionRelightHint.tex 
\begin{table}[]
  \caption{Quantitative comparison of different variants of passing
    radiance hints to the DiLightNet (rows 1-3), the number of
    radiance hints (rows 4-6), impact of including the segmentation
    mask (row 7-8) and different training data augmentation schemes
    (rows 9-12).}
   \label{tab:ablation}
   \begin{tabular}{l|ccc}
   \hline
   Variant              & PSNR   & SSIM   & LPIPS  \\ \hline
\textbf{Our Network}  & \textbf{22.97} & \textbf{0.8249} & \textbf{0.1165} \\
   Direct ControlNet & 22.82 & 0.8216 & 0.1212 \\
   Non-Encoded Multiplication & 22.40 & 0.8174 & 0.1232 \\
   \hline
   3 Radiance Hints & 22.92 & 0.8197 & 0.1188 \\
   \textbf{4 Radiance Hints} & \textbf{22.97} & \textbf{0.8249} & \textbf{0.1165} \\
   5 Radiance Hints & 22.79 & 0.8200 & 0.1176 \\ \hline
   \textbf{w/ Mask} & \textbf{22.97} & \textbf{0.8249} & \textbf{0.1165} \\
   w/o Mask & 22.23 & 0.8148 & 0.1184 \\ \hline
   \textbf{Full Augmentation}& \textbf{22.97} & \textbf{0.8249} & \textbf{0.1165} \\
   w/o Material Augmentation    & 22.90 & 0.8235 & 0.1178 \\ 
   w/o Smoothed Normal     & 21.88 & 0.7974 & 0.1314 \\ 
   w/o Color Augmentation     & 22.54 & 0.8161 & 0.1223 \\ \hline
   \end{tabular}
\end{table}

%!TEX root = ../../DiffusionRelightHint.tex

\newcommand{\ablationvisualFigWidth}{0.20\textwidth}
\begin{figure*}[h]
\centering
  \renewcommand{\arraystretch}{0.0}
  \addtolength{\tabcolsep}{-6.0pt}
 \begin{tabular}{ ccccc }
  \includegraphics[width=\ablationvisualFigWidth]{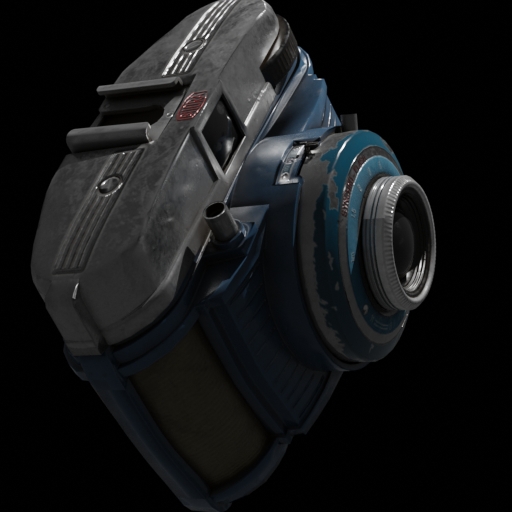}
 &\includegraphics[width=\ablationvisualFigWidth]{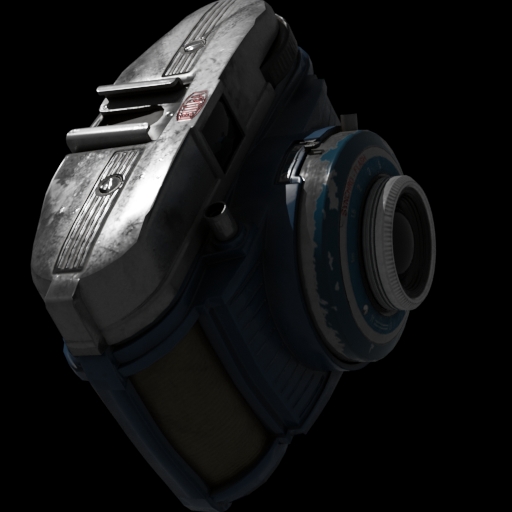}
 &\includegraphics[width=\ablationvisualFigWidth]{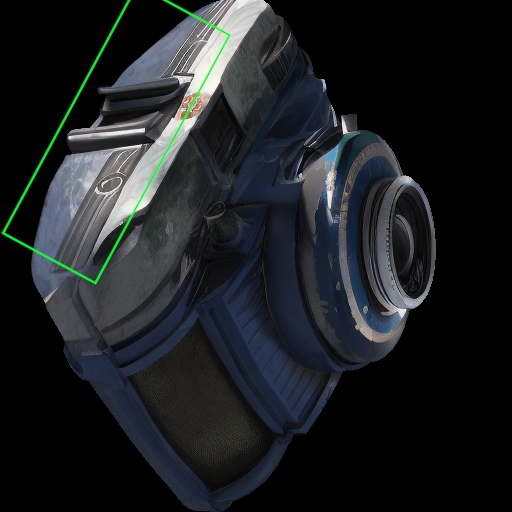}
 &\includegraphics[width=\ablationvisualFigWidth]{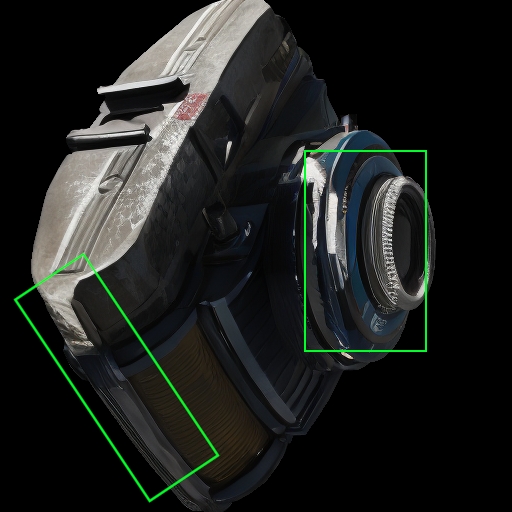}
 &\includegraphics[width=\ablationvisualFigWidth]{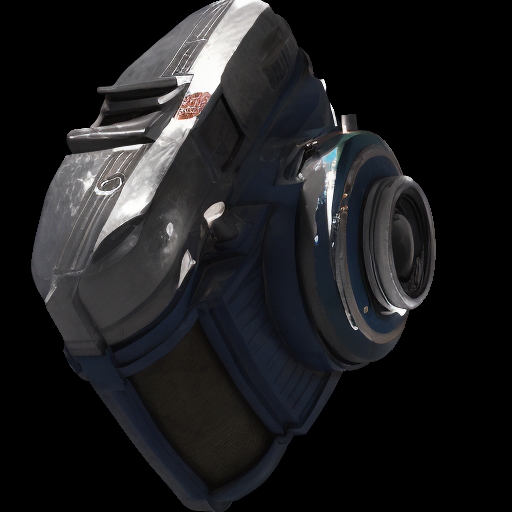}\\
 Provisional & Reference & Direct & Non-Enc & Ours\\
 \end{tabular} 
 \caption{Ablation comparison of different architecture variants that:
   (1) \emph{direct}ly pass the radiance hints and provisional image
   (without multiplication) to ControlNet, and (2) multiply the
   radiance hints with the \emph{non-encoded} (Non-Enc) provisional image.
   DiLightNet's encoded multiplication generates visually more
   plausible results.}
  \label{fig:ablation_visual}
\end{figure*}

\paragraph{Impact of Number of Radiance Hints}
\autoref{tab:ablation} (rows 4-6) compares the impact of changing the
number of (specular) radiance hints; all variants include a diffuse
radiance hint. The $3$ radiance hints variant includes $2$ specular
radiance hints with roughness $0.13$, and $0.34$.  The $4$ radiance
hints variant includes one additional specular radiance hint with
roughness $0.05$.  Finally, the $5$ radiance hints variant includes an
additional (sharp specular) hint with roughness $0.02$.  From the
quantitative results in~\autoref{tab:ablation} we can see that $4$
radiance hints perform best.  Upon closer inspection of the results,
we observe that there is little difference for scenes that exhibit a
simple shape with simple materials. However, for scenes with a more
complex shape we find that the $3$ radiance hints are insufficient to
accurately model the light-matter interactions. For scenes with
complex materials, we found that providing too many radiance hints can
also be detrimental due to the limited quality of the (smoothed)
depth-estimated normals.

\paragraph{Foreground Masking}
DiLightNet takes the foreground mask as additional input. To better
understand the impact of including the mask, we also train a variant
without taking the mask as an additional channel.  Instead we fill the
background with black pixels in the provisional image.  During
training we also remove the background in the reference images.  As a
consequence, DiLightNet will learn to generate a black background.
For the ablation, we only compute the errors over the foreground
pixels.  As shown in \autoref{tab:ablation} (rows 7-8), the variant
trained without a mask produces larger errors especially on cases with
either complex shape or materials.

%\input{src/figures/fig_ablation_visual}

% !TEX root = ../../DiffusionRelightHint.tex

\newcommand{\imagerelightFigWidth}{0.16\textwidth}
\begin{figure}
  \renewcommand{\arraystretch}{0.0}
  \addtolength{\tabcolsep}{-6.0pt}
  \begin{tabular}{ ccc }
    \includegraphics[width=\imagerelightFigWidth]{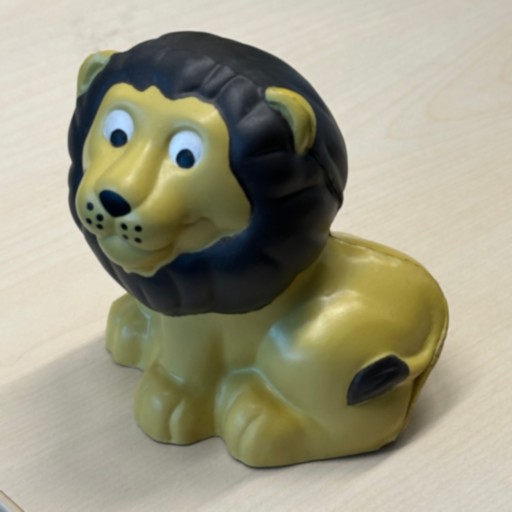}
    & \includegraphics[width=\imagerelightFigWidth]{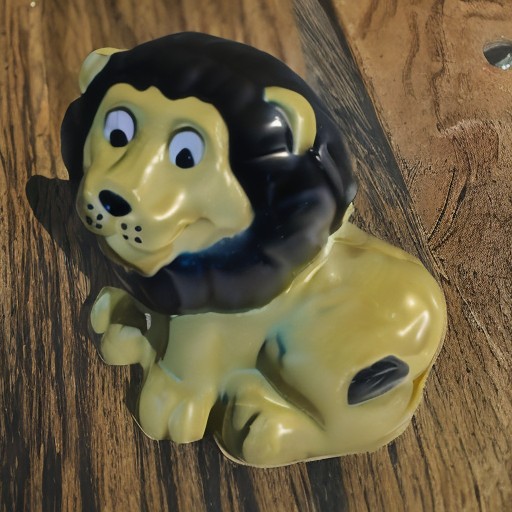}
    & \includegraphics[width=\imagerelightFigWidth]{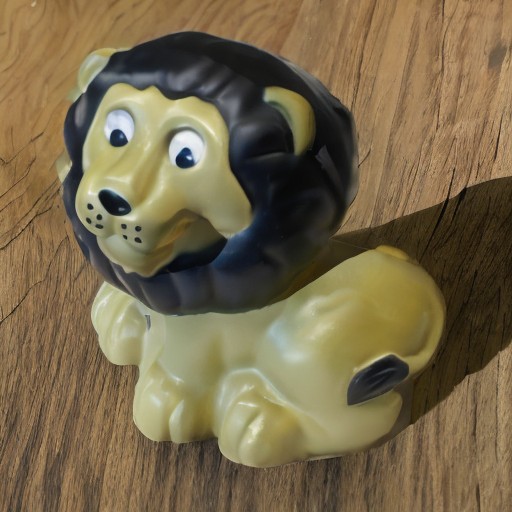}
  \end{tabular}
  \caption{A demonstration of single image relighting obtained by
    bypassing the first stage and directly injecting a captured
    photograph as the provisional image (left).  The resulting
    generated images (middle and right) represent a plausible
    relighting of the given photograph.}
  \label{fig:image_relight}
\end{figure}

%!TEX root = ../../DiffusionRelightHint.tex

\newcommand{\depthcondFigWidth}{0.16\textwidth}
\begin{figure}
\centering
\renewcommand{\arraystretch}{0.0}
\addtolength{\tabcolsep}{-6.0pt}
\begin{tabular}{ ccc }
 \includegraphics[width=\depthcondFigWidth]{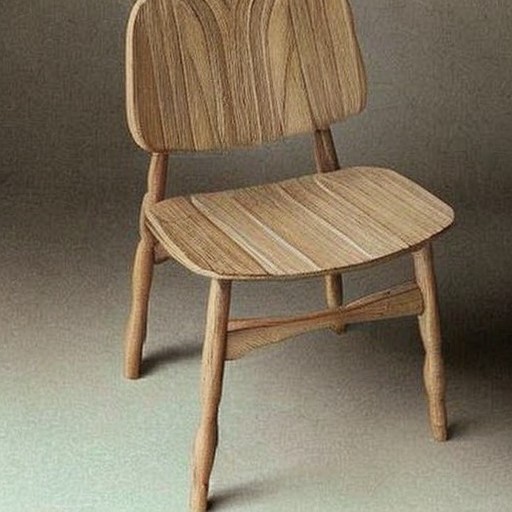}
 &\includegraphics[width=\depthcondFigWidth]{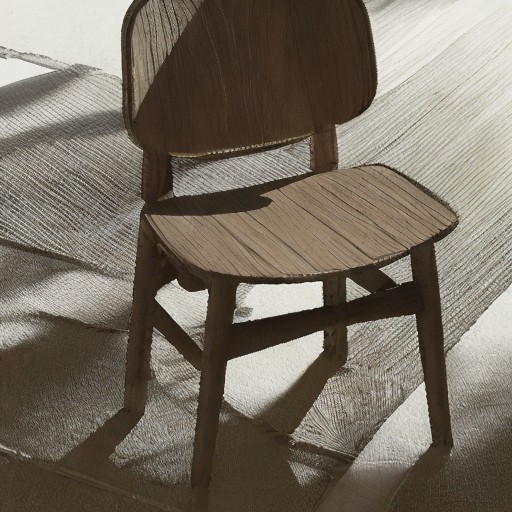}
 &\includegraphics[width=\depthcondFigWidth]{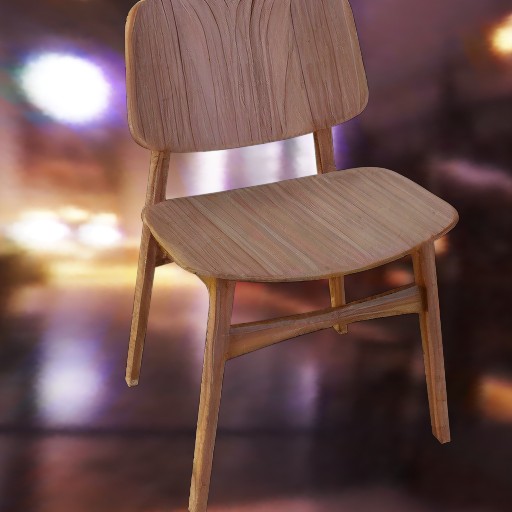}\\
  \includegraphics[width=\depthcondFigWidth]{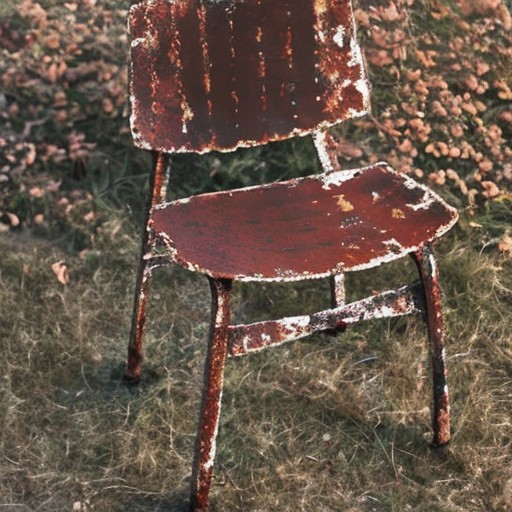}
 &\includegraphics[width=\depthcondFigWidth]{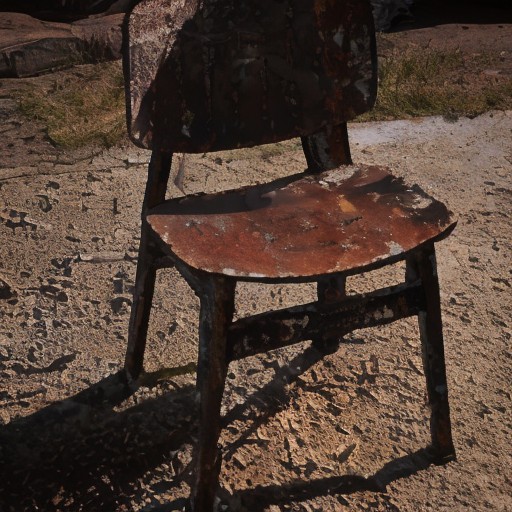}
 &\includegraphics[width=\depthcondFigWidth]{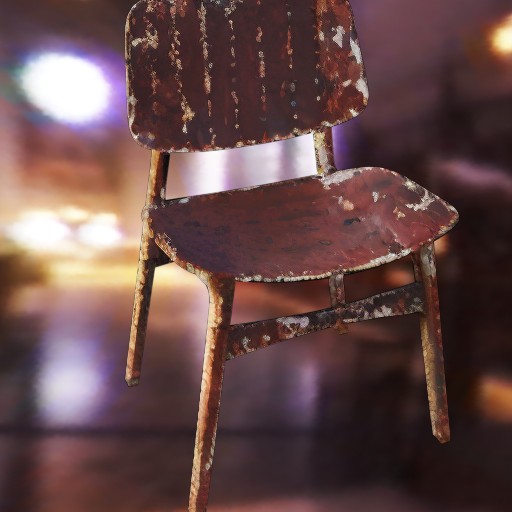}\\
\includegraphics[width=\depthcondFigWidth]{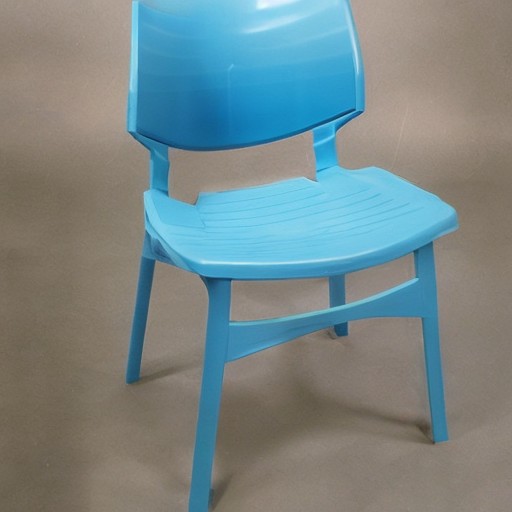}
&\includegraphics[width=\depthcondFigWidth]{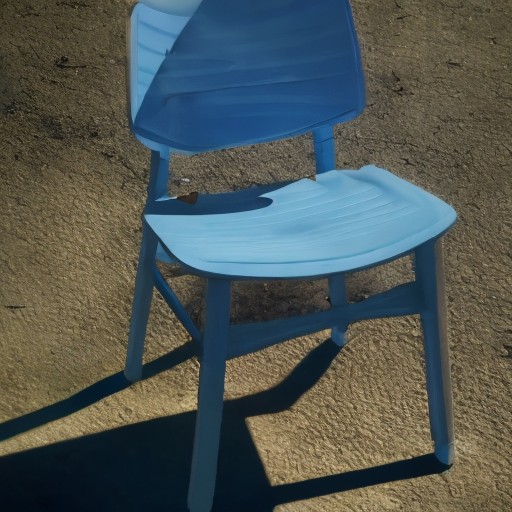}
&\includegraphics[width=\depthcondFigWidth]{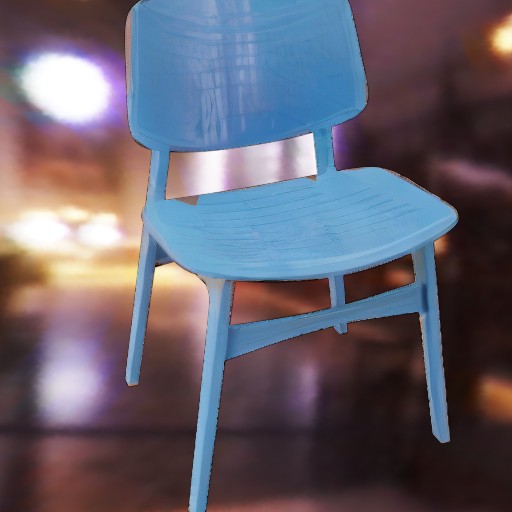}
 \end{tabular}
 \caption{Lighting control results for a depth-controlled
   text-to-image diffusion model improves the quality of the results
   by providing a depth map as additional input.}
  \label{fig:depth_cond_gen}
\end{figure}

\paragraph{Training Augmentation}
We eliminate each of the three augmentations from the training set to
better gauge their impact (\autoref{tab:ablation}, rows 9-12):
\begin{itemize}
\item \emph{Without Normal Augmentation:} This variant is trained
  using radiance hints rendered with the ground truth shading normals,
  instead of the smoothed depth-estimated normals or the geometric
  normals;
\item \emph{Without Color Augmentation:} This variant is trained on
  the full training set without swapping the RGB color channels; and
\item \emph{Without Material Augmentation:} This model is trained
  with the basic $13K$ dataset without material augmentations.
\end{itemize}
From~\autoref{tab:ablation}, we observe that all three augmentations
improve the robustness of DiLightNet.  Of all augmentations, the
normal augmentation has the largest impact as it helps to bridge the
domain gap between perfect shading normals (in the training) and the
smoothed estimated depth normals.  The color augmentation also
improves the quality for all test scenes, albeit to lesser degree.
The benefits of the material augmentation are most noticeable for
objects with smooth shapes (i.e., low geometrical complexity) as
errors in the normal estimation can help to mask inaccuracies in
representing complex materials.

\section{Discussion}
\paragraph{Relation to Single Image Relighting}
By skipping the first stage and directly inputing a captured
photograph as the provisional image into DiLightNet, we can perform
approximate single image relighting
(\autoref{fig:image_relight}). However, due to the lack of a text
prompt, the relighting results might not be ideal.  Furthermore,
unlike existing single image relighting methods that are trained for a
more narrow class of scenes, DiLightNet is trained to handle any type
of synthesized image for which there might not exists a 'real'
reference under novel lighting (e.g., the 'giraffe-turtle'
in~\autoref{fig:text2img_main}), DiLightNet only aims to produce
\emph{plausible} images.  Nevertheless, the relighting results generated
by DiLightNet are plausible for scenes from which a reasonably
accurate depth and mask can be extracted.  Further refining DiLightNet
to be more robust for relighting photographs is a promising avenue for
future research.

\paragraph{Limitations}
Our method is not without limitations.  Due to the limitations of
specifying the image content with text prompts, the user only has
limited control over the materials in the scene.  Consequently, the
material-light interactions might not follow the intention of the
prompt-engineer.  DiLightNet enables some indirect control, beyond
text prompts, through the appearance-seed.  Integrating material aware
diffusion models, such as Alchemist~\cite{Sharma:2023:APC}, could
potentially lead to better control over the material-light
interactions.  Furthermore, our method relies on a number of
off-the-shelf solutions for estimating a rough depth map and
segmentation mask of the foreground object. While our method is robust
to some errors in the depth map, some types of errors (e.g., the
bass-relief ambiguity) can result in non-satisfactory results.  An
interesting alternative pipeline takes a reference depth map as input
(e.g., using a depth conditioned diffusion model such as
\emph{``stable-diffusion-2-depth''}), thereby bypassing the need to
estimate the depth and mask. As demonstrated
in~\autoref{fig:depth_cond_gen}, augmenting the input with a reference
depth map, further increases the quality of the results.  Finally,
animating/altering the lighting using a fixed content-seed can result
in some minor structural shape changes because the images are
generated independently (see supplemental video).  Incorporating
cross-frame consistency to improve temporal stability is an
interesting avenue for future research.

\section{Conclusion}
\label{sec:conclusion}

In this paper we introduced a novel method for controlling the
lighting in diffusion-based text-to-image generation.  Our method
consists of three stages: (1) provisional image synthesis under
uncontrolled lighting using existing text-to-image methods, (2)
resynthesis of the foreground object using our novel DiLightNet
conditioned by the radiance hints of the foreground object, and
finally (3) inpainting of the background consistent with the target
lighting.  Key to our method is DiLightNet, a variant of ControlNet
that takes an encoded version of the provisional image (to retain the
shape and texture information) multiplied with the radiance hints.
Our method is able to generate images that match both the text prompt
and the target lighting.  For future work we would like to apply
DiLightNet to estimate reflectance properties from a single photograph
and for text-to-3D generation with rich material properties.

\section*{Acknowledgments}
  Pieter Peers was supported in part by NSF grant IIS-1909028. Chong
  Zeng and Hongzhi Wu were partially supported by NSF China (62332015 \& 62227806), the Fundamental Research Funds for the Central Universities (226-2023-00145), and Information Technology Center and State Key Lab of CAD\&CG, Zhejiang University.

{
    \small
    \bibliographystyle{ACM-Reference-Format}
    \bibliography{src/reference}
}
% \newpage

\section*{Appendix}
\subsection*{Comparison to Concurrent Work}
Concurrent to our work,
Bashkirova~\etal~\shortcite{bashkirova2023lasagna} introduced a
lighting control method for image generation named ``Lasagna''.
Although Lasagna shares a similar goal as DiLightNet, it uses language
tokens instead of radiance hints to control the lighting and thus
lacks the fine-grained lighting control of DiLightNet.  Furthermore,
Lasagna only supports a predefined set of $12$ directional lights. Due
to ambiguities in the lighting specification used in the publicly
available pretrained Lasagna model, we can only compare both methods
for a synthetic dataset under Lasagna's ID-0 (top) and ID-6 (front)
lighting.  Specifically, we perform lighting control on our synthetic
test dataset, with the lighting either set as a point light source at
the top or in front of the object.  We then follow the same
configuration as our ablation study to measure the quantitative errors
using PSNR, SSIM and LIPIPS~\cite{Zhang:2018:LPIPS}.  As shown in
\autoref{tab:lasagna} our method consistently outperforms Lasagna
across all metrics. A qualitative comparison is shown
in~\autoref{fig:lasagna_comparison}.

\subsection*{Additional Ablation Study}
\paragraph{Mask Ablation:} \autoref{fig:mask_ablation} shows the
  visual impact of passing the mask to DiLightNet.  We observe that
  without a mask, there are more occurrences of incorrect specular
  highlights as the network is unable to differentiate between dark
  foreground pixels and background.

\paragraph{Number of Radiance Hints:} \autoref{fig:ablation_hint_visual}
shows the visual effect of using a different number of radiance
hints.  Using 3 radiance hints often results in missed or blurred
highlights.  Using too many radiance hints also tends to adversely
affect the results due to the limited accuracy of the (smoothed)
depth-estimated normals used for rendering the radiance hints causing
sharp specular highlights to be incorrectly placed.

\subsection*{Additional Results}

\begin{figure}[t]
  \centering
  \includegraphics[width=0.45\textwidth]{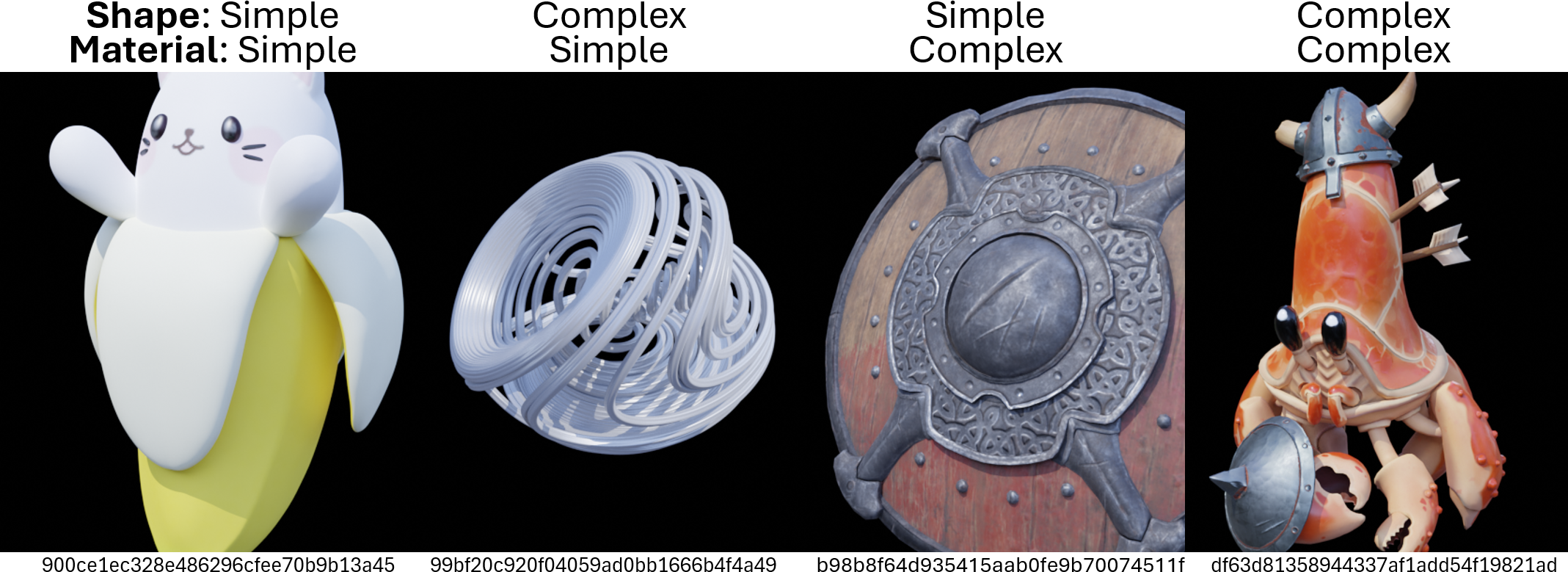}
  \caption{Representative examples, with Objaverse ID for
    completeness, from the synthetic test with different complexities in
    shape and/or material.}
    
  \label{fig:syndata}
\end{figure}

\paragraph{Examples of the synthetic test set.}
\autoref{fig:syndata} shows representative examples from the test set. 
Our test dataset covers a wide range of shapes with different complexities
of shapes and materials.

\paragraph{Example of Radiance Hints:} \autoref{fig:hints} shows
  the radiance hints used by DiLightNet to control the incident
  lighting for a \emph{``leather glove''}.
  
\paragraph{Additional Results:} \autoref{fig:text2img_supp05},
  \ref{fig:text2img_supp01}, \ref{fig:text2img_supp04}, \ref{fig:text2img_supp02}, \ref{fig:text2img_supp03}, \ref{fig:text2img_DE01}, and
  \ref{fig:text2img_DE02} show additional text to image generation 
  results, including the
  impact of changing the content-seed using the same text prompt.  For
  all examples, we show the results for $3$ different lighting
  conditions.

\begin{figure*}
\centering
\renewcommand{\arraystretch}{0.8}
\newcommand{\syntheticResFigWidth}{0.25\textwidth}
\addtolength{\tabcolsep}{-5.0pt}
    \begin{tabular}{ cccc }
    \includegraphics[width=\syntheticResFigWidth]{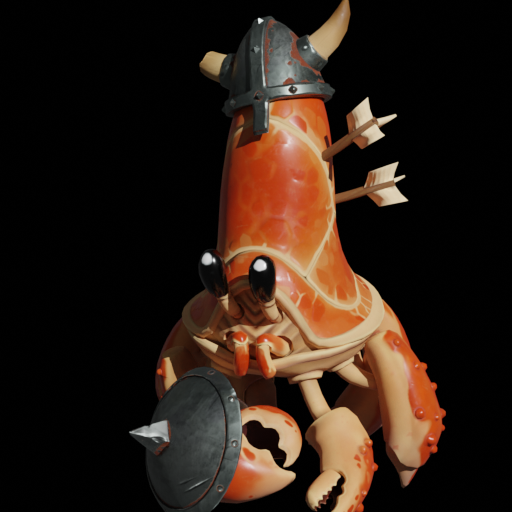}
    &\includegraphics[width=\syntheticResFigWidth]{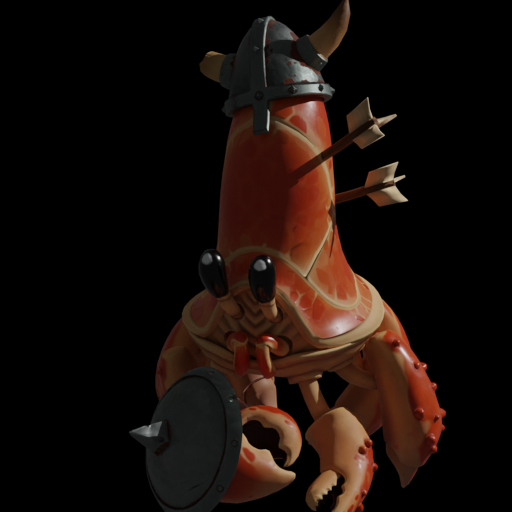}
    &\includegraphics[width=\syntheticResFigWidth]{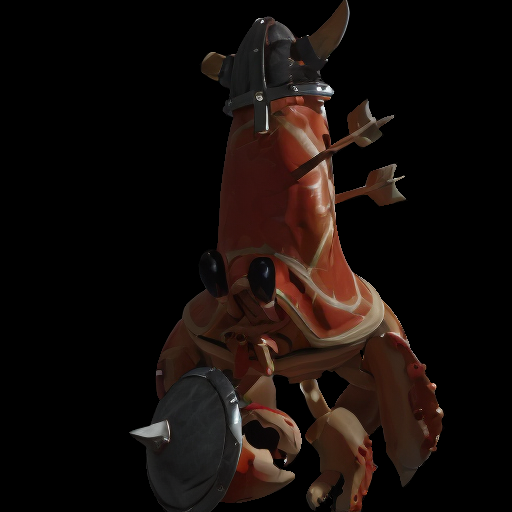}\\
    \includegraphics[width=\syntheticResFigWidth]{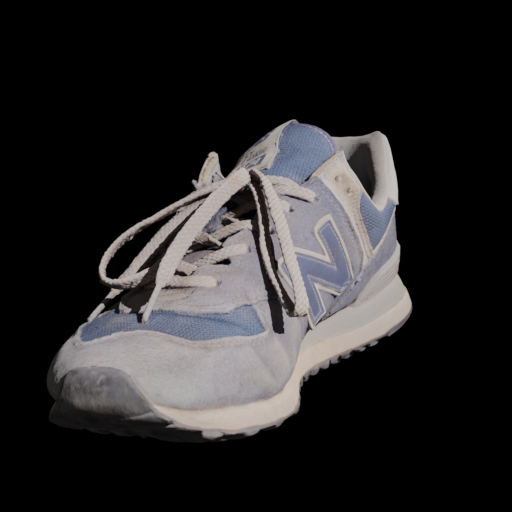}
    &\includegraphics[width=\syntheticResFigWidth]{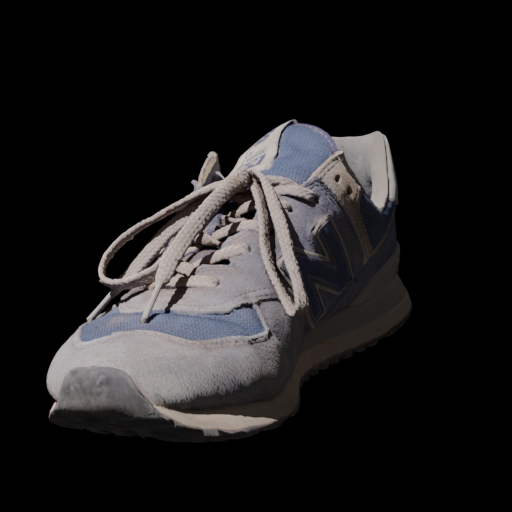}
    &\includegraphics[width=\syntheticResFigWidth]{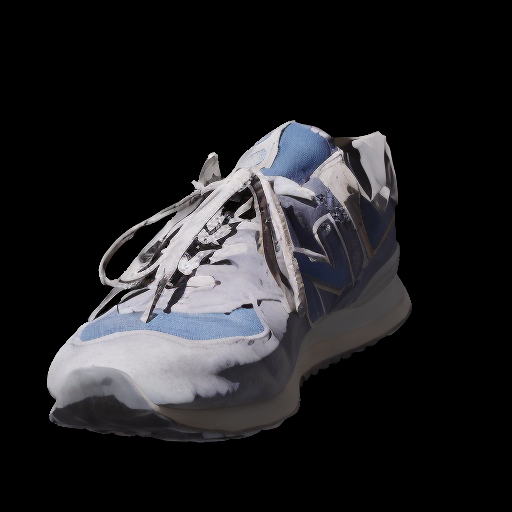}\\
    \includegraphics[width=\syntheticResFigWidth]{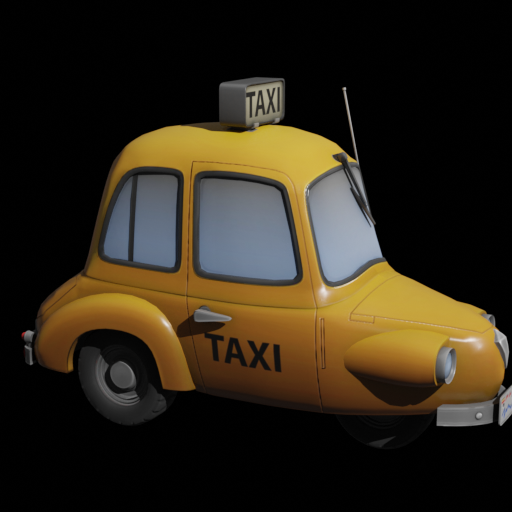}
    &\includegraphics[width=\syntheticResFigWidth]{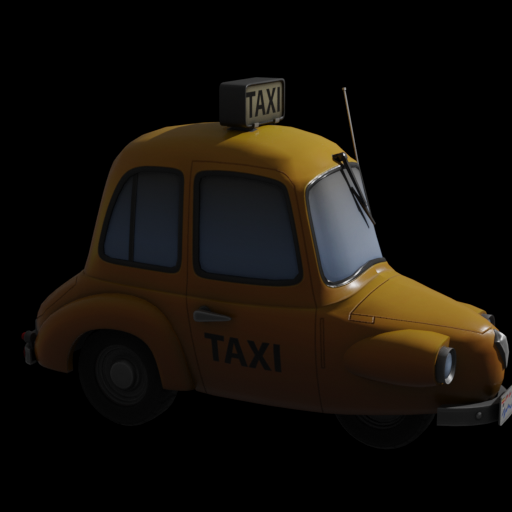}
    &\includegraphics[width=\syntheticResFigWidth]{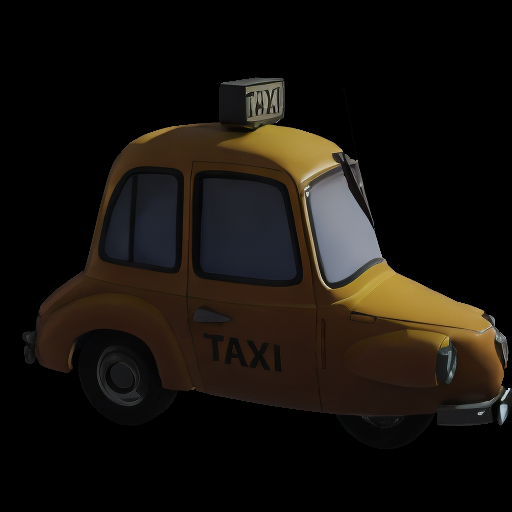}\\
    \includegraphics[width=\syntheticResFigWidth]{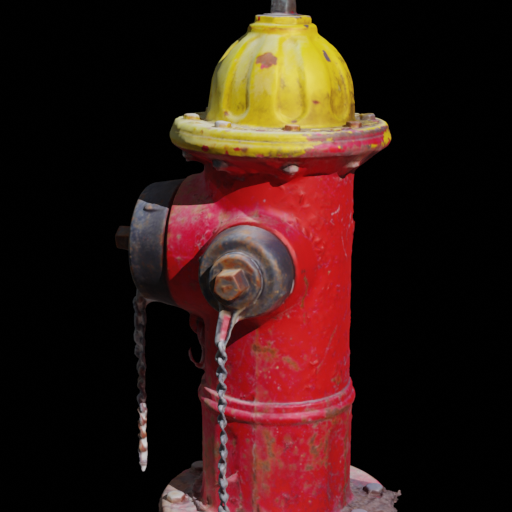}
    &\includegraphics[width=\syntheticResFigWidth]{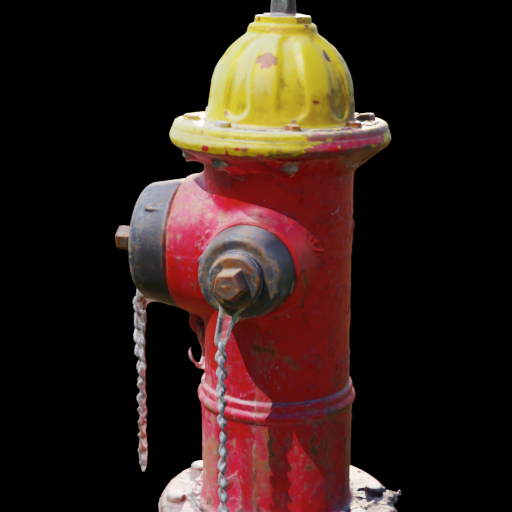}
    &\includegraphics[width=\syntheticResFigWidth]{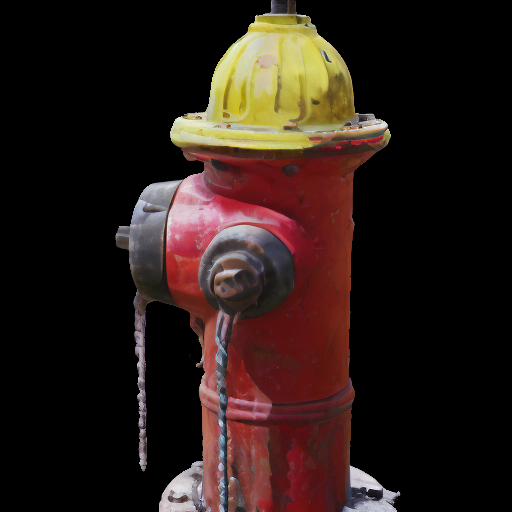}\\
    \end{tabular} 
    \caption{Additional results with synthetic data. The first column
      shows the provisional image as a reference, whereas the second
      column is the reference image rendered under the target
      lighting. The last column is the result generated by DiLightNet
      under the target lighting.}
    \label{fig:synthetic_results}
\end{figure*}

  \paragraph{Synthetic Results:} \autoref{fig:synthetic_results} shows
  additional results with synthetic data. The first column shows the
  provisional image as a reference, and the second column shows the
  reference image rendered under the target lighting. The last column
  shows the result generated under the target lighting (we select the
  best (lowest LPIPS) result from $4$ candidate seeds).  Note that our
  method produces plausible results that qualitatively match the
  reference with some minor differences in the shadows and specular
  highlights. These differences are mostly due to the approximate
  shape of the estimated depth.

\newcommand{\comparisonFigWidth}{0.25\textwidth}
\begin{figure}[t]
\renewcommand{\arraystretch}{0.25}
\addtolength{\tabcolsep}{-5.5pt}
\begin{tabular}{ cc }
  \includegraphics[width=\comparisonFigWidth]{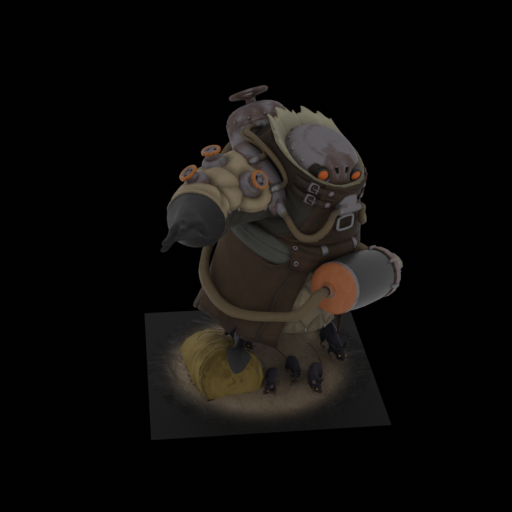}
 &\includegraphics[width=\comparisonFigWidth]{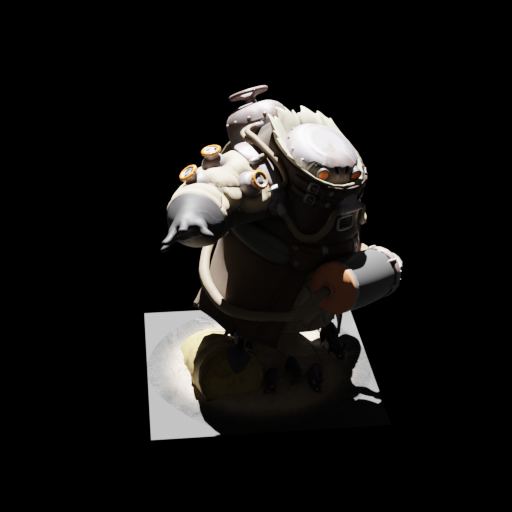}\\
   Provisional & Reference \\
 \includegraphics[width=\comparisonFigWidth]{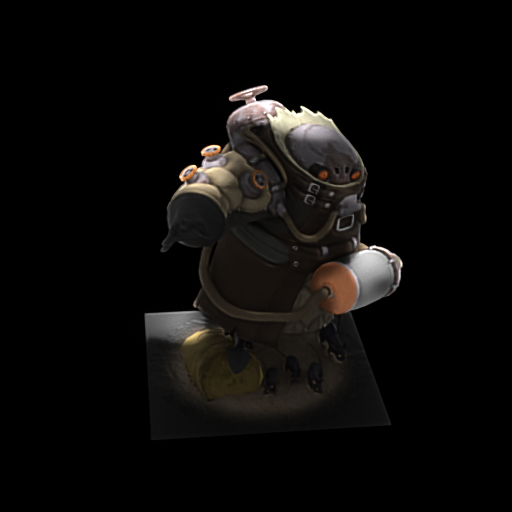}
 &\includegraphics[width=\comparisonFigWidth]{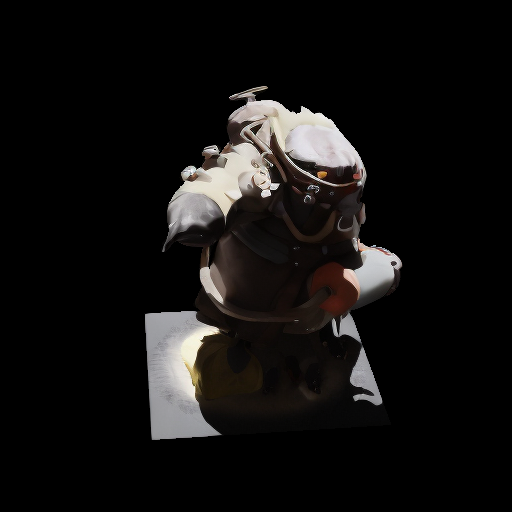}\\
 Lasagna & DiLightNet (ours) \\
 \cite{bashkirova2023lasagna} & \\
 \end{tabular}
 \caption{Visual comparison of DiLIghtNet with Lasagna
   \cite{bashkirova2023lasagna}.  The DiLightNet result more closely
   matches the overall shading and shadow casted by the point light
   source than the Lasagna result which exhibits incorrect shadows and
   shading effects (e.g., on the barrel).}
   \label{fig:lasagna_comparison}
\end{figure}

\begin{table}
\centering
  \caption{Qualitative comparison to
    Lasagna~\cite{bashkirova2023lasagna}.}
    \begin{tabular}{l|ccc}
       & PSNR & SSIM & LPIPS\\
      \hline
      Ours &  21.09 & 0.8443 &0.1152 \\
      Lasagna & 17.41 & 0.8352 & 0.1359
    \end{tabular}
  \label{tab:lasagna}
  \end{table}

%!TEX root = ../../DiffusionRelightHint.tex

%!TEX root = ../../DiffusionRelightHint.tex

\newcommand{\ablationvisualSuppFigWidth}{0.2\textwidth}
\begin{figure*}[h]
\centering 
\renewcommand{\arraystretch}{0.25}
\addtolength{\tabcolsep}{-6.0pt}
\begin{tabular}{ cccc }
  \includegraphics[width=\ablationvisualSuppFigWidth]{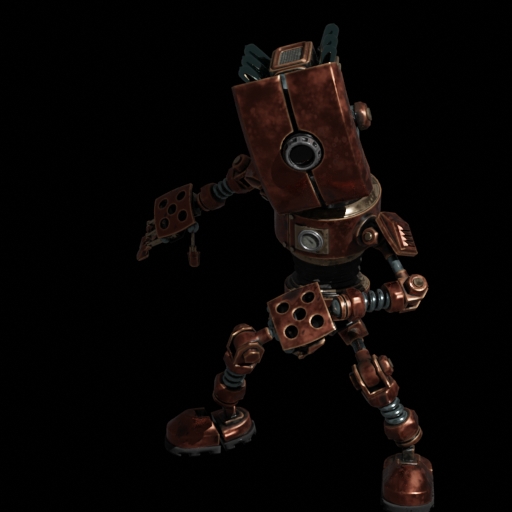}
 &\includegraphics[width=\ablationvisualSuppFigWidth]{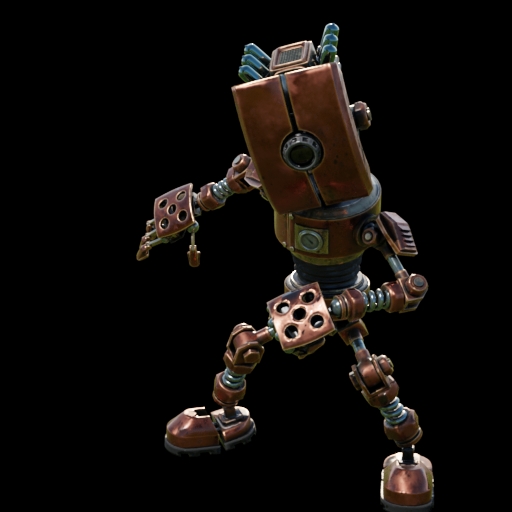}
 &\includegraphics[width=\ablationvisualSuppFigWidth]{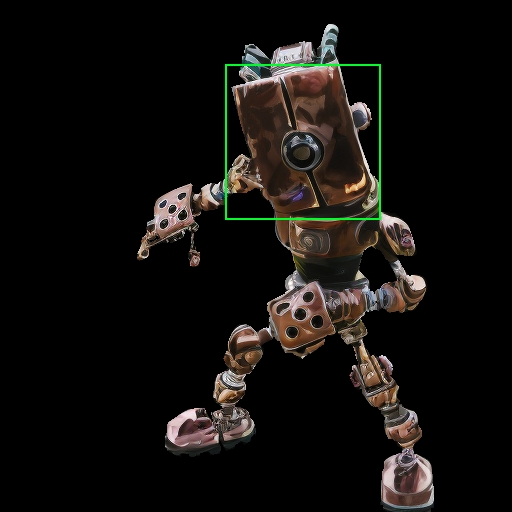}
 &\includegraphics[width=\ablationvisualSuppFigWidth]{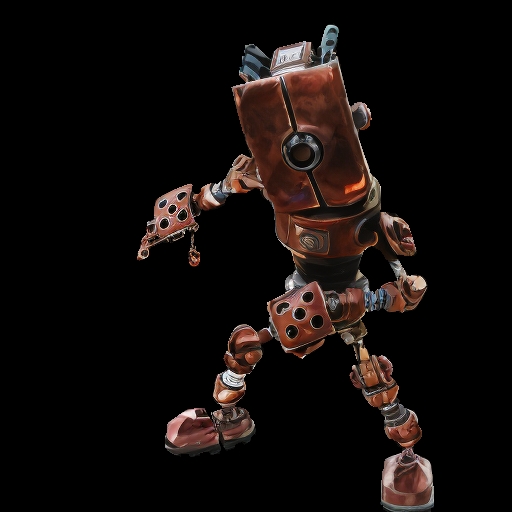}\\
  Provisional & Reference & w/o Mask & w/ Mask \\
 \end{tabular} 
 \caption{Not passing the mask as an extra input channel will result in more occurences of incorrect specular highlights.}
   \label{fig:mask_ablation}
\end{figure*}

\begin{figure*}[h]
\centering
\renewcommand{\arraystretch}{0.25}
\addtolength{\tabcolsep}{-6.0pt}
\begin{tabular}{ ccccc }
 \includegraphics[width=\ablationvisualSuppFigWidth]{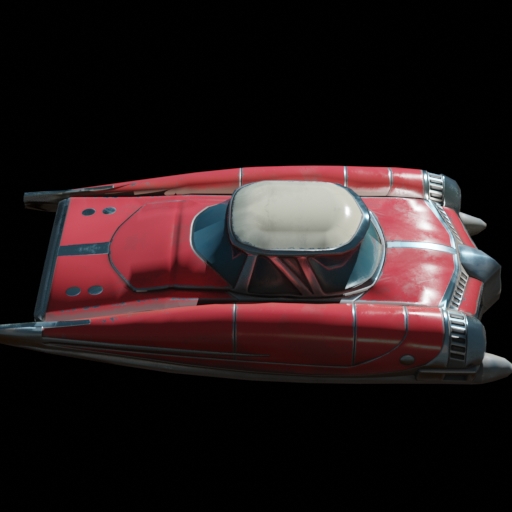}
 &\includegraphics[width=\ablationvisualSuppFigWidth]{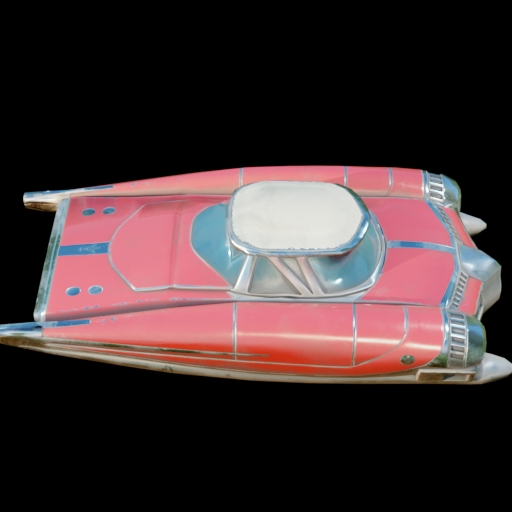}
 &\includegraphics[width=\ablationvisualSuppFigWidth]{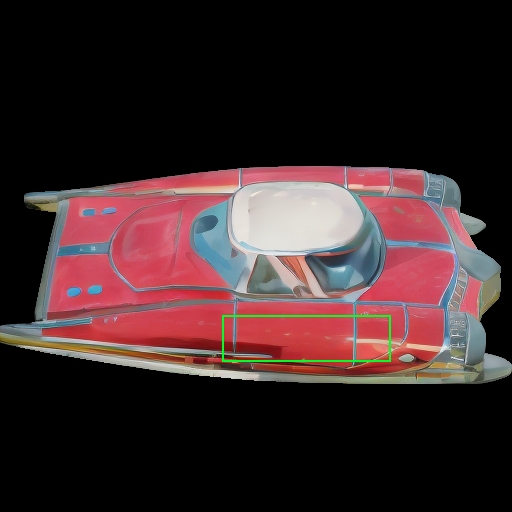}
 &\includegraphics[width=\ablationvisualSuppFigWidth]{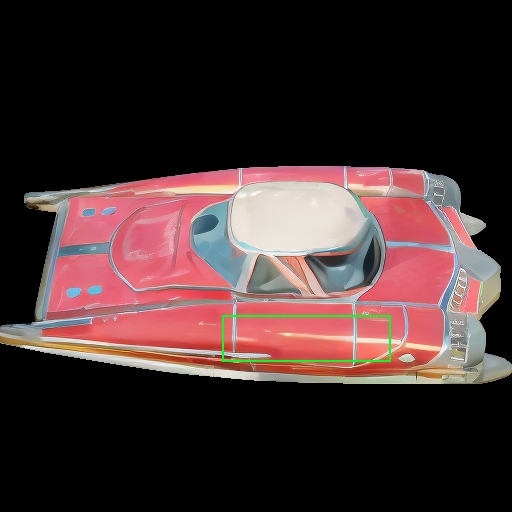}
 &\includegraphics[width=\ablationvisualSuppFigWidth]{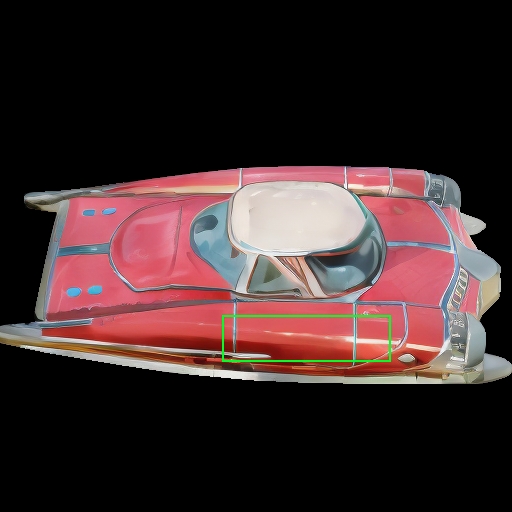}\\
 Provisional & Reference & 3 Radiance Hints & 4 Radiance Hints (Ours) & 5 Radiance Hints\\
 \end{tabular} 
 \caption{Ablation comparison of using a different number of radiance
   hints.  With only \emph{3 radiance hints}, DiLightNet misses some
   specular highlights, while too many hints (\emph{5 radiance hints})
   can also adversely affect results due to the inaccuracies in the
   depth estimates used to generate the specular radiance hints. In
   our implementaion we opt for using \emph{4 radiance hints} which
   produces visually more plausible results.}
  \label{fig:ablation_hint_visual}
\end{figure*}

%!TEX root = ../../supplementary.tex

\newcommand{\hintexpFigWidth}{0.166\textwidth}
\begin{figure*}
\centering
\renewcommand{\arraystretch}{0.25}
\addtolength{\tabcolsep}{-6.0pt}
 \begin{tabular}{ cccccc }
 \includegraphics[width=\hintexpFigWidth]{src/figures/effect_of_seed/pov_33.jpg}
 &
 \includegraphics[width=\hintexpFigWidth]{src/figures/effect_of_seed/33_2.jpg}
 &
 \includegraphics[width=\hintexpFigWidth]{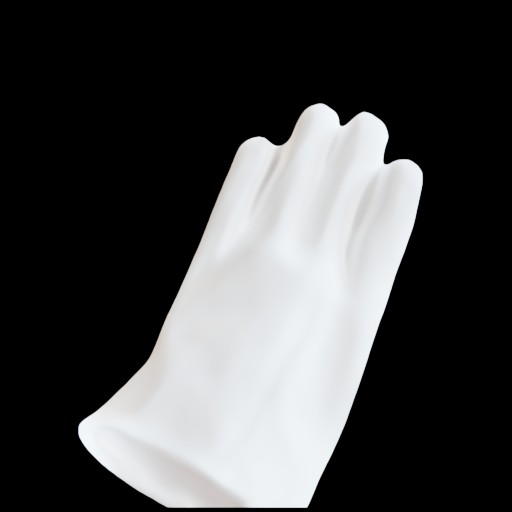}
 &
 \includegraphics[width=\hintexpFigWidth]{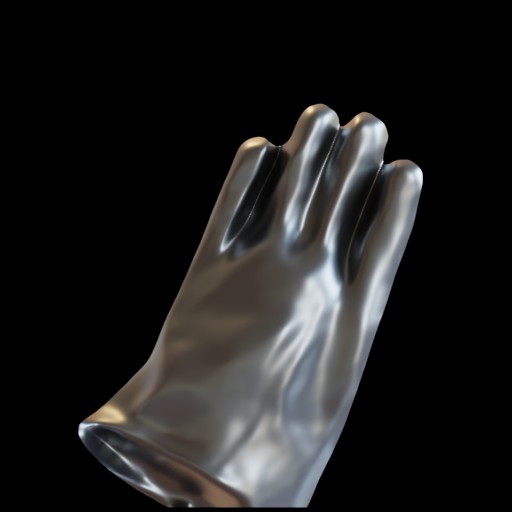}
 &
 \includegraphics[width=\hintexpFigWidth]{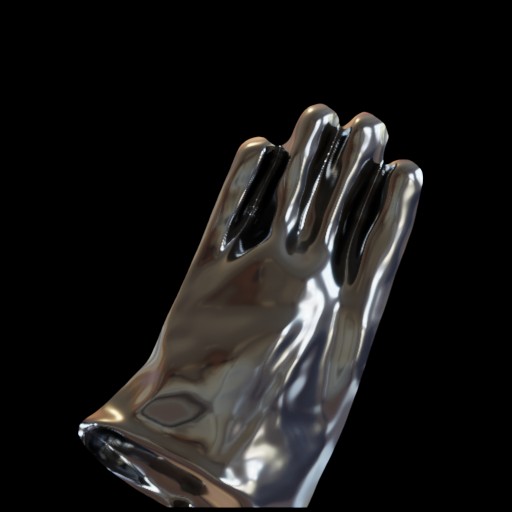}
 &
 \includegraphics[width=\hintexpFigWidth]{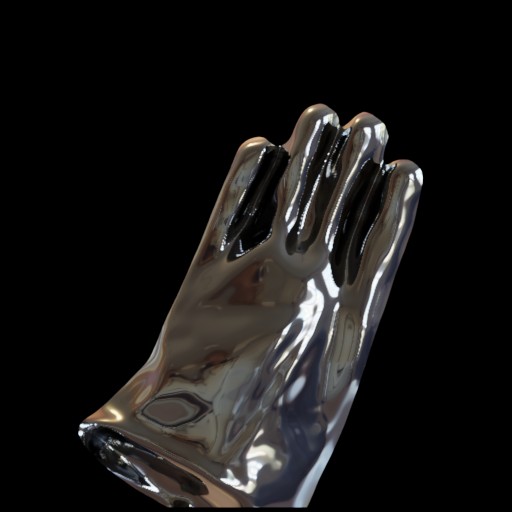}\\
 Provisional image & Our result & Diffuse hint& Roughness $0.34$ hint & Roughness $0.13$ hint & Roughness $0.05$ hint\\
 \end{tabular} 
 \caption{Example visualizations of the radiance hints for a
   \emph{``leather glove''}. Note that DiLightNet leverages the
   learned space of images embedded in the diffusion model to generate
   rich shading details from the smoothed shading information encoded
   in the radiance hints.}
  \label{fig:hints}
\end{figure*}

\newcommand{\moreResFullFigWidth}{0.25\textwidth}
%!TEX root = ../../supplementary.tex

\begin{figure*}
\renewcommand{\arraystretch}{0.8}
\addtolength{\tabcolsep}{-5.0pt}
 \begin{tabular}{ cccc }
  \includegraphics[width=\moreResFullFigWidth]{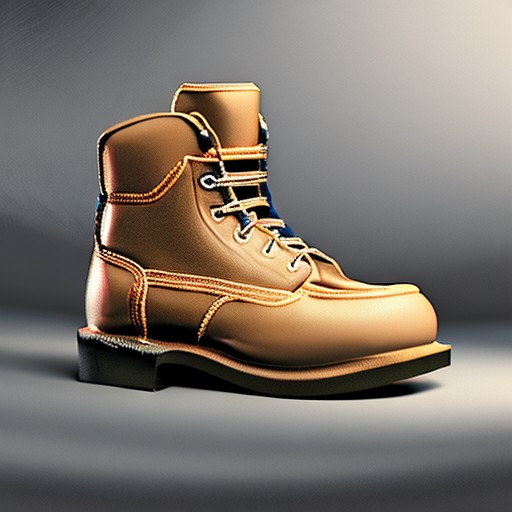}
 &\includegraphics[width=\moreResFullFigWidth]{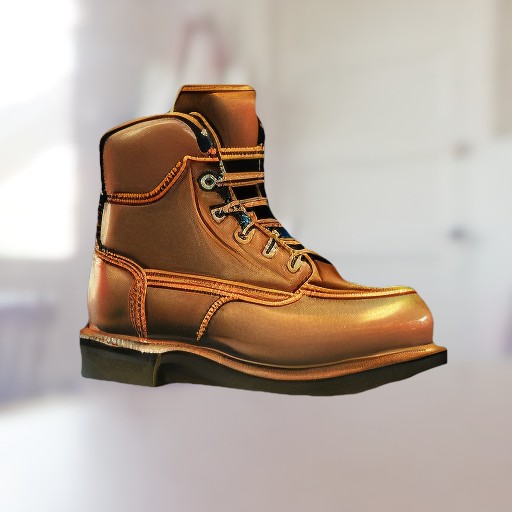}
 &\includegraphics[width=\moreResFullFigWidth]{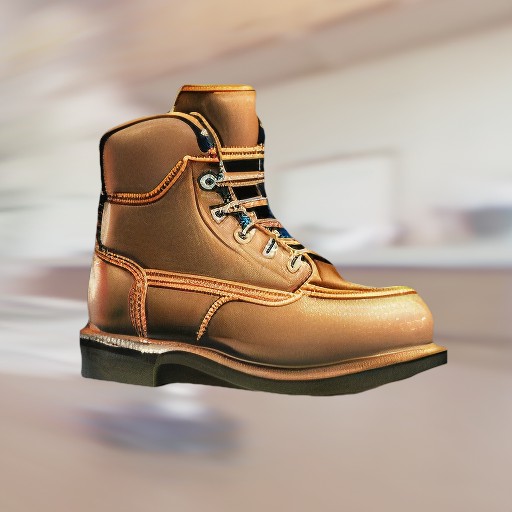}
 &\includegraphics[width=\moreResFullFigWidth]{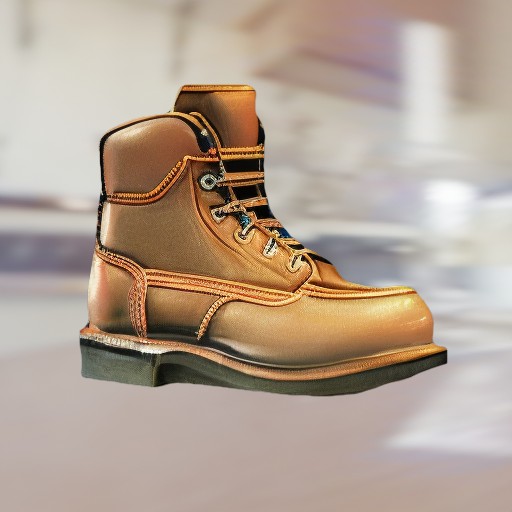}\\
  \multicolumn{4} {c} {
    Prompt: \emph{``caterpillar work boot''}.
    }\\
 \end{tabular} 
 \caption{Text-to-image generated results with lighting control. The
   first column shows the provisional image as a reference, whereas
   the last three columns are generated under different user-specified
   environment lighting conditions. }
  \label{fig:text2img_supp05}
\end{figure*}

%!TEX root = ../../supplementary.tex

\begin{figure*}
\renewcommand{\arraystretch}{0.8}
\addtolength{\tabcolsep}{-5.0pt}
 \begin{tabular}{ cccc }
  \includegraphics[width=\moreResFullFigWidth]{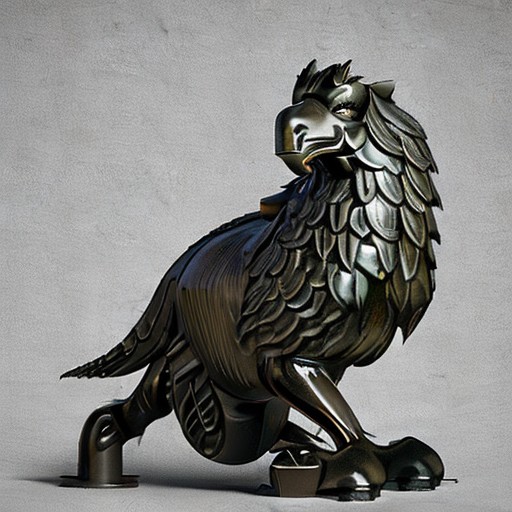}
 &\includegraphics[width=\moreResFullFigWidth]{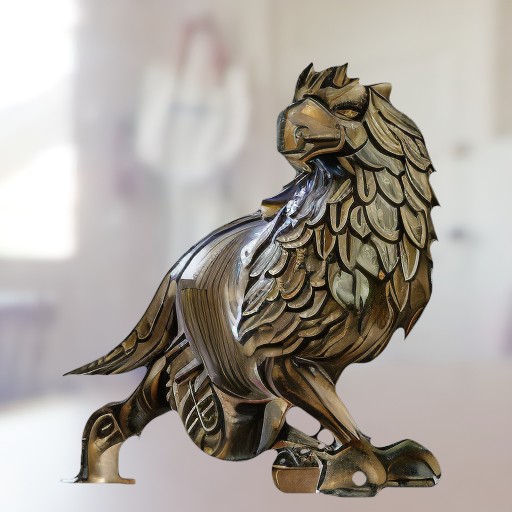}
 &\includegraphics[width=\moreResFullFigWidth]{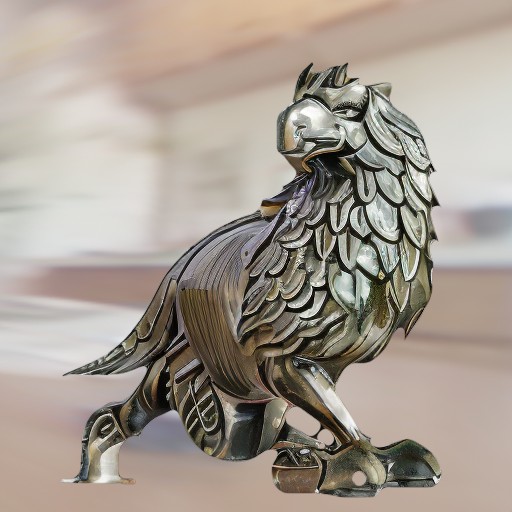}
 &\includegraphics[width=\moreResFullFigWidth]{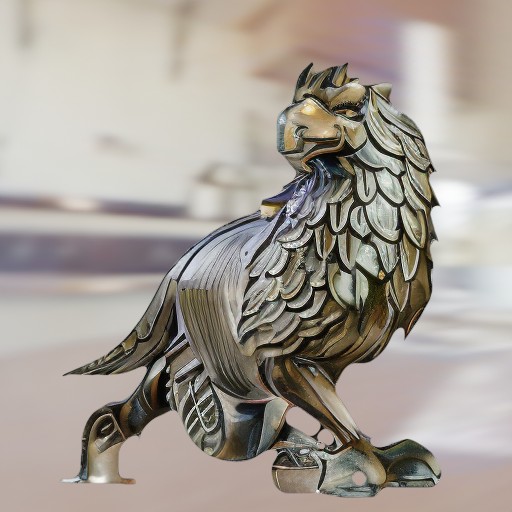}\\
  \multicolumn{4} {c} {
    Prompt: \emph{``stone griffin''}.
    }\\
  \includegraphics[width=\moreResFullFigWidth]{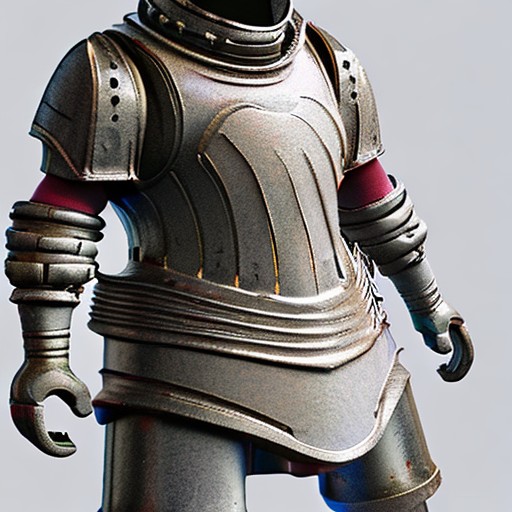}
 &\includegraphics[width=\moreResFullFigWidth]{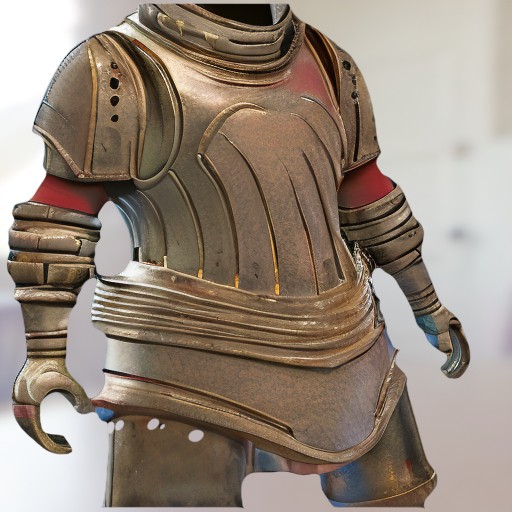}
 &\includegraphics[width=\moreResFullFigWidth]{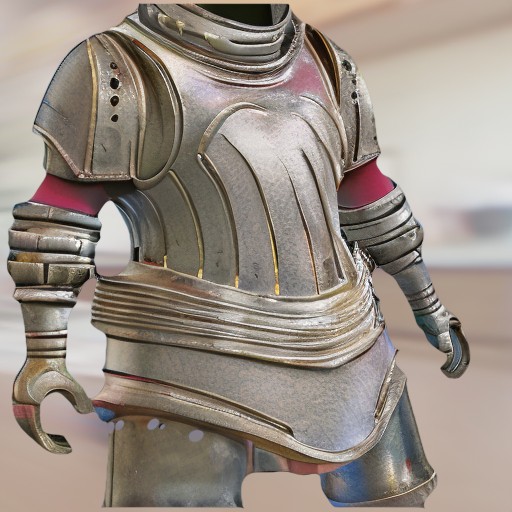}
 &\includegraphics[width=\moreResFullFigWidth]{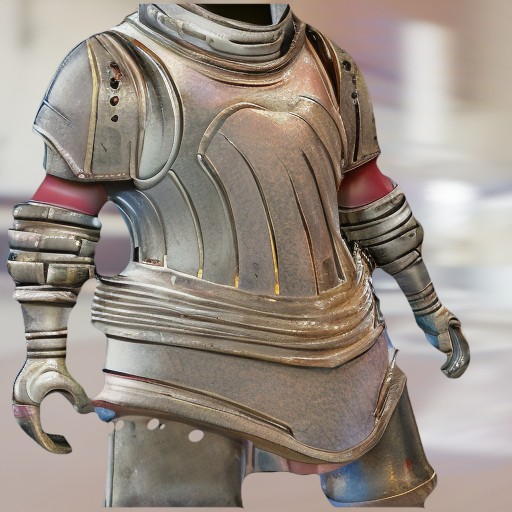}\\
  \multicolumn{4} {c} {
    Prompt: \emph{``full plate armor''}.
    }\\
  \includegraphics[width=\moreResFullFigWidth]{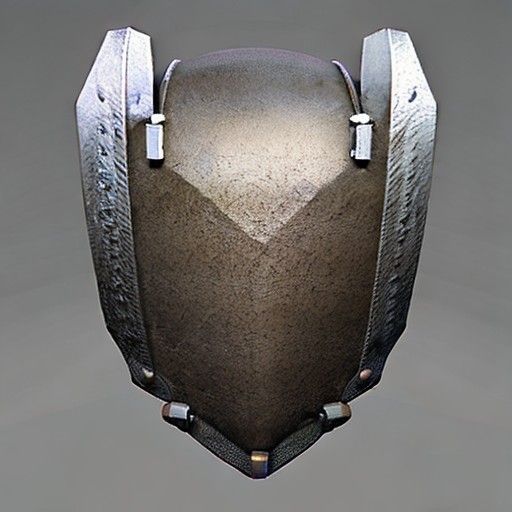}
 &\includegraphics[width=\moreResFullFigWidth]{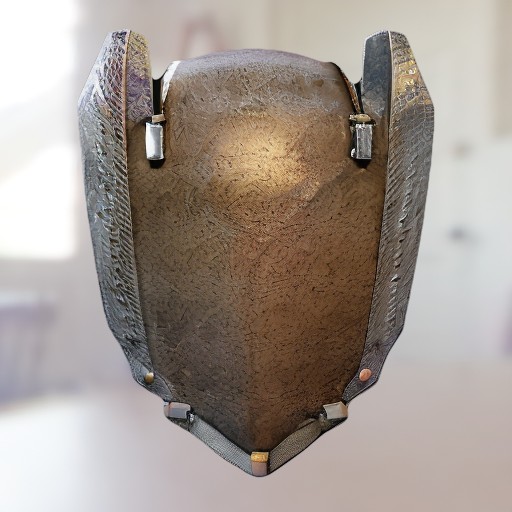}
 &\includegraphics[width=\moreResFullFigWidth]{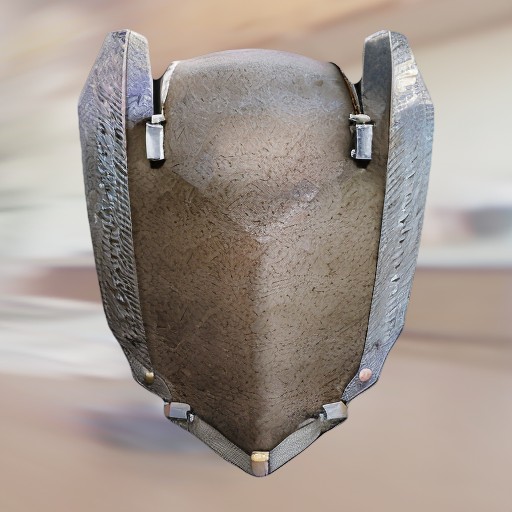}
 &\includegraphics[width=\moreResFullFigWidth]{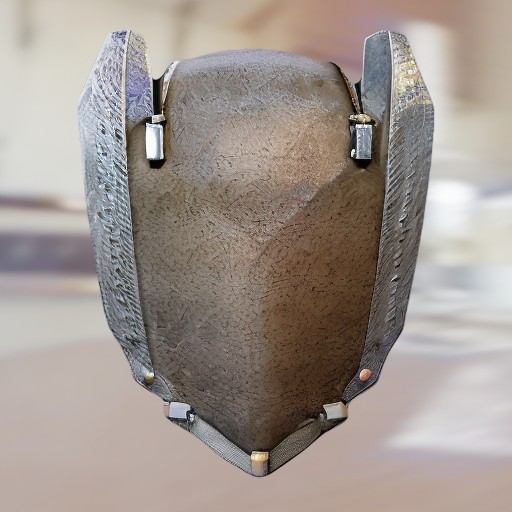}\\
  \multicolumn{4} {c} {
    Prompt: \emph{``full plate armor''}.
    }\\
  \includegraphics[width=\moreResFullFigWidth]{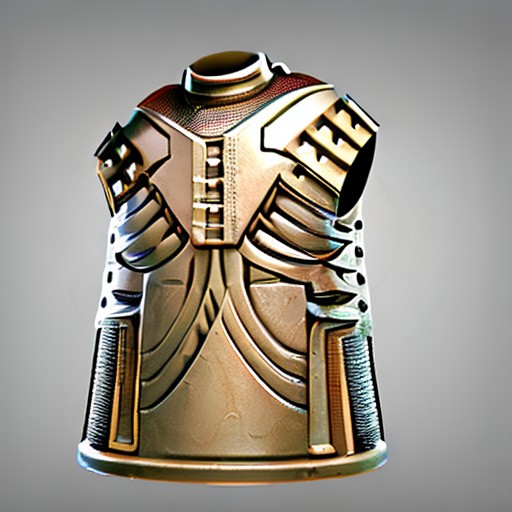}
 &\includegraphics[width=\moreResFullFigWidth]{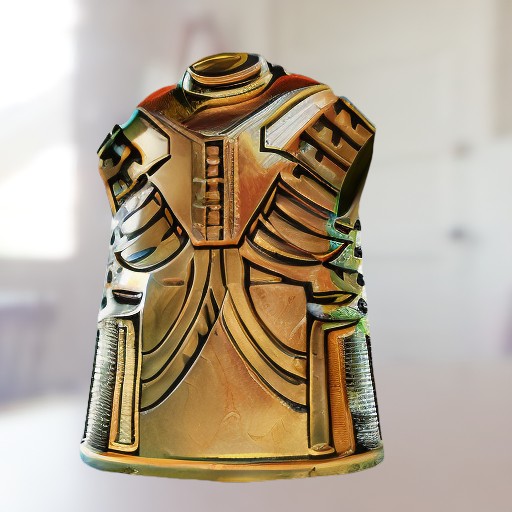}
 &\includegraphics[width=\moreResFullFigWidth]{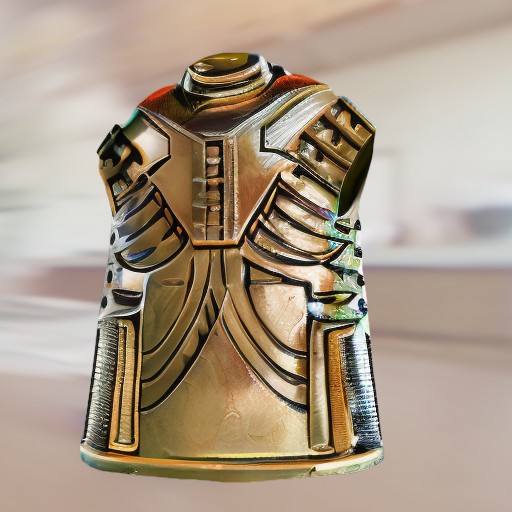}
 &\includegraphics[width=\moreResFullFigWidth]{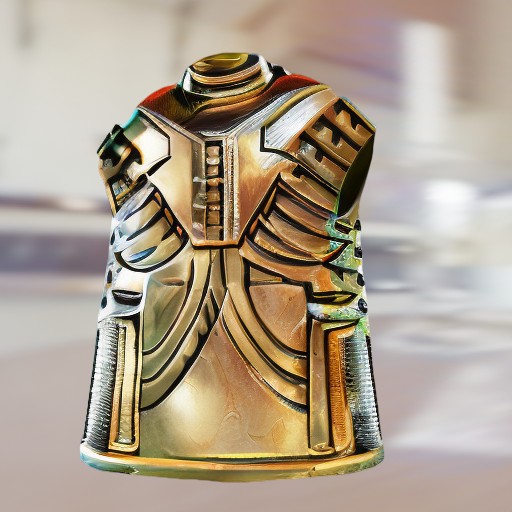}\\
  \multicolumn{4} {c} {
    Prompt: \emph{``full plate armor''}.
    }\\
 \end{tabular} 
 \caption{Text-to-image generated results with lighting control. The
   first column shows the provisional image as a reference, whereas
   the last three columns are generated under different user-specified
   environment lighting conditions. }
  \label{fig:text2img_supp01}
\end{figure*}

%!TEX root = ../../supplementary.tex

\begin{figure*}
\renewcommand{\arraystretch}{0.8}
\addtolength{\tabcolsep}{-5.0pt}
 \begin{tabular}{ cccc }
  \includegraphics[width=\moreResFullFigWidth]{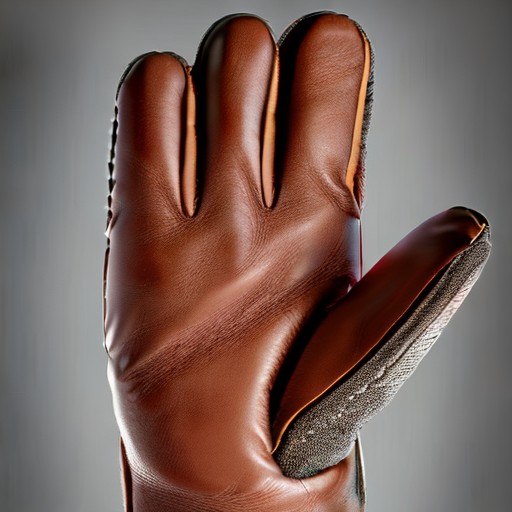}
 &\includegraphics[width=\moreResFullFigWidth]{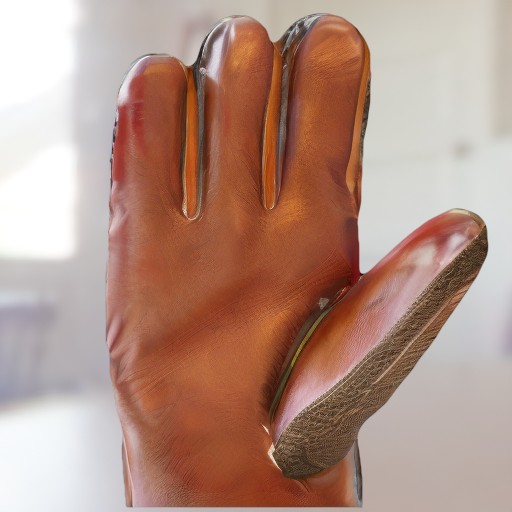}
 &\includegraphics[width=\moreResFullFigWidth]{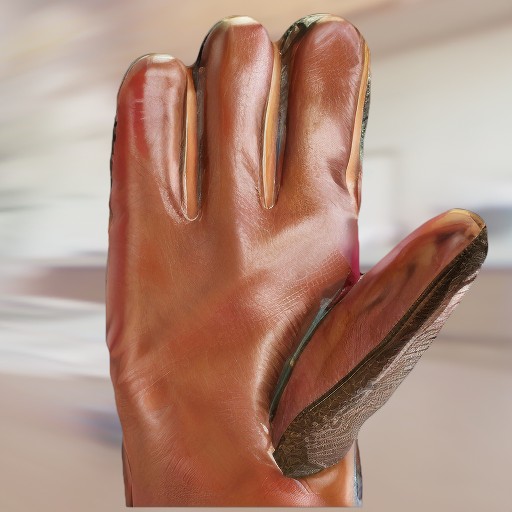}
 &\includegraphics[width=\moreResFullFigWidth]{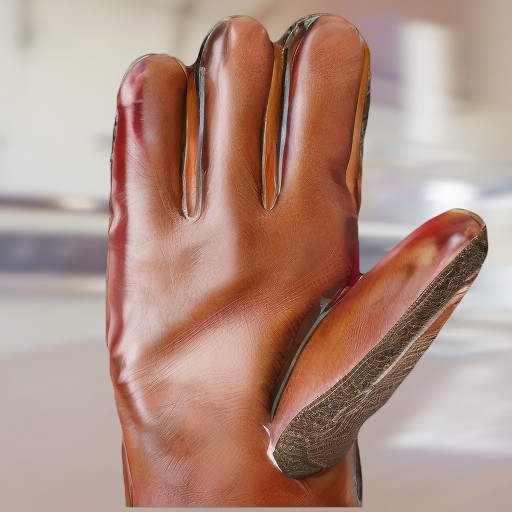}\\
  \multicolumn{4} {c} {
    Prompt: \emph{``leather glove''}.
    }\\
  \includegraphics[width=\moreResFullFigWidth]{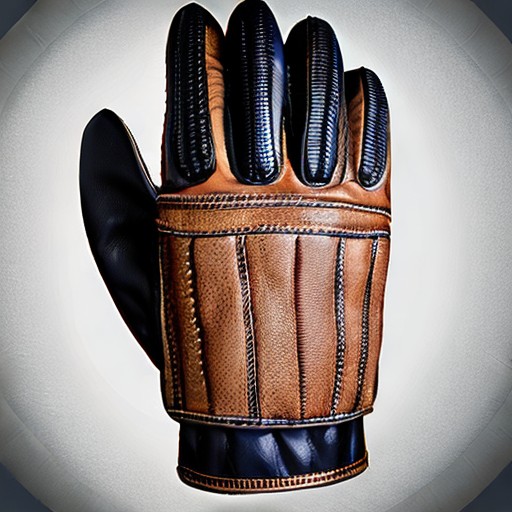}
 &\includegraphics[width=\moreResFullFigWidth]{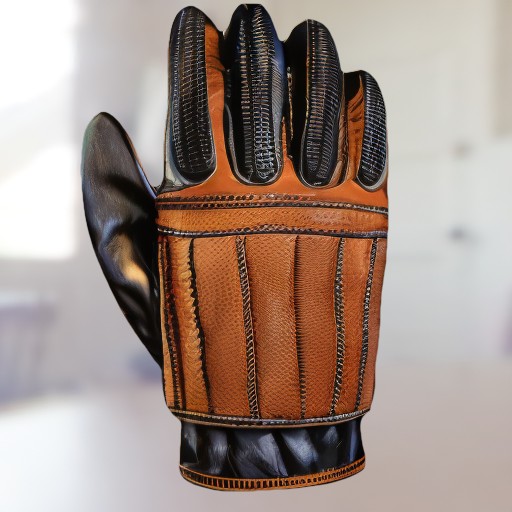}
 &\includegraphics[width=\moreResFullFigWidth]{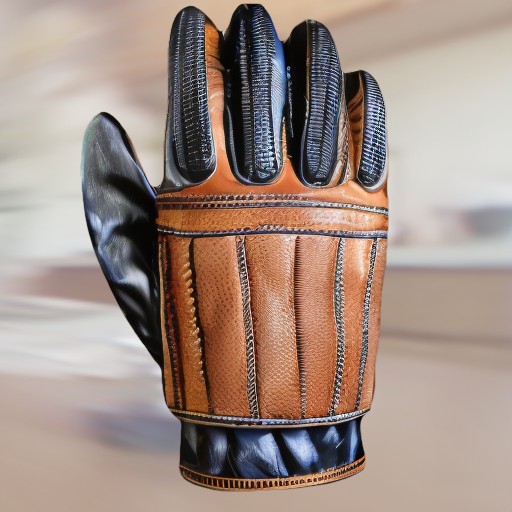}
 &\includegraphics[width=\moreResFullFigWidth]{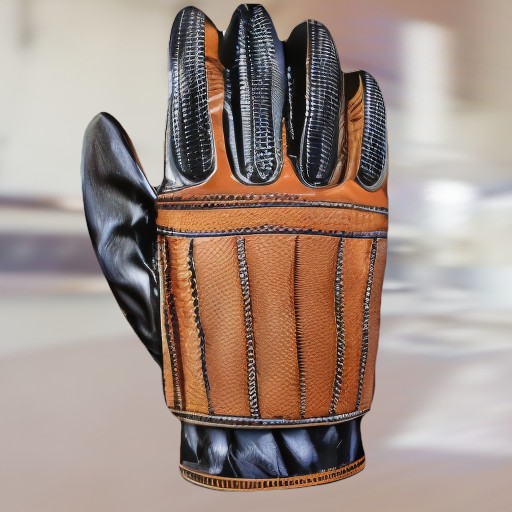}\\
  \multicolumn{4} {c} {
    Prompt: \emph{``leather glove''}.
    }\\
  \includegraphics[width=\moreResFullFigWidth]{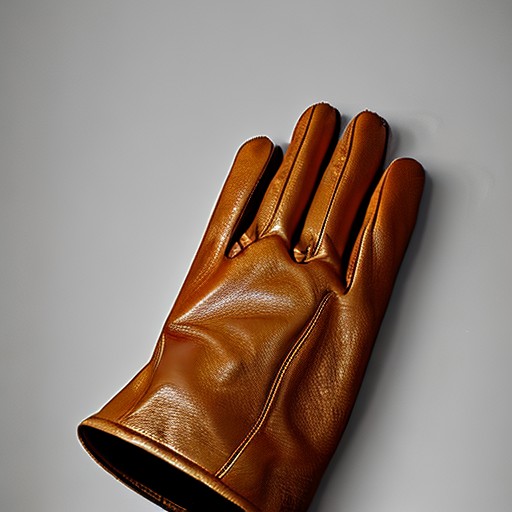}
 &\includegraphics[width=\moreResFullFigWidth]{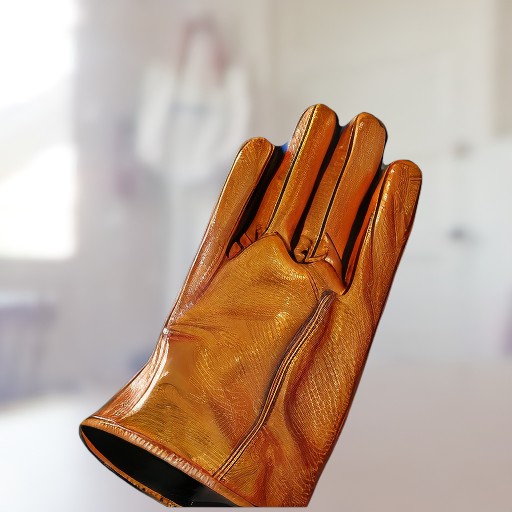}
 &\includegraphics[width=\moreResFullFigWidth]{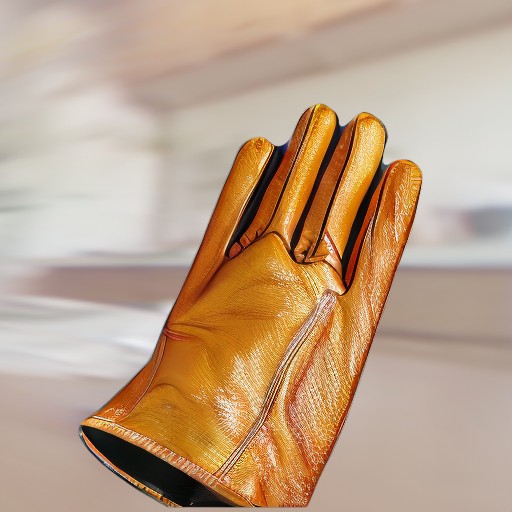}
 &\includegraphics[width=\moreResFullFigWidth]{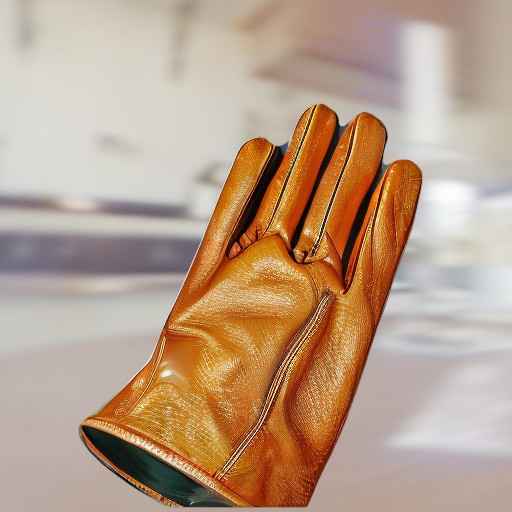}\\
  \multicolumn{4} {c} {
    Prompt: \emph{``leather glove''}.
    }\\
  \includegraphics[width=\moreResFullFigWidth]{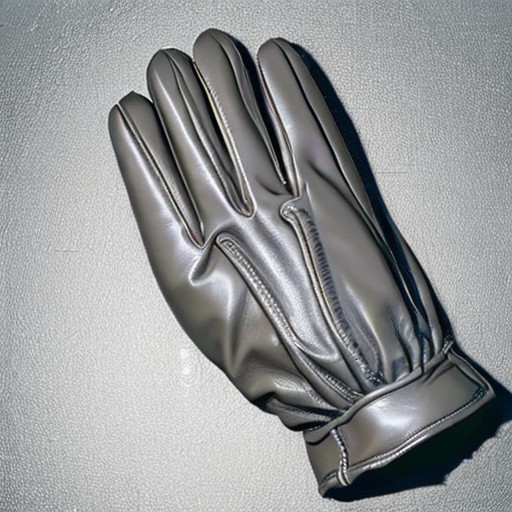}
 &\includegraphics[width=\moreResFullFigWidth]{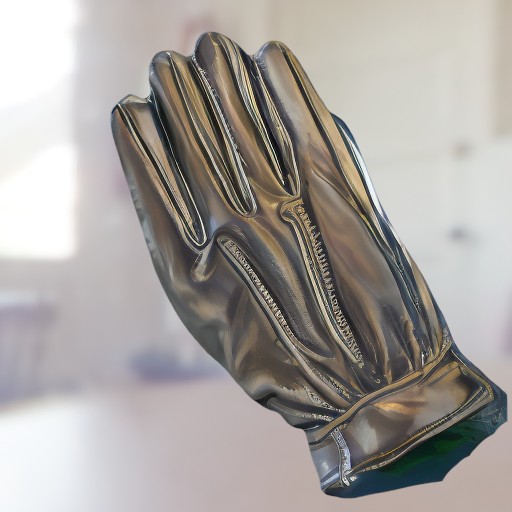}
 &\includegraphics[width=\moreResFullFigWidth]{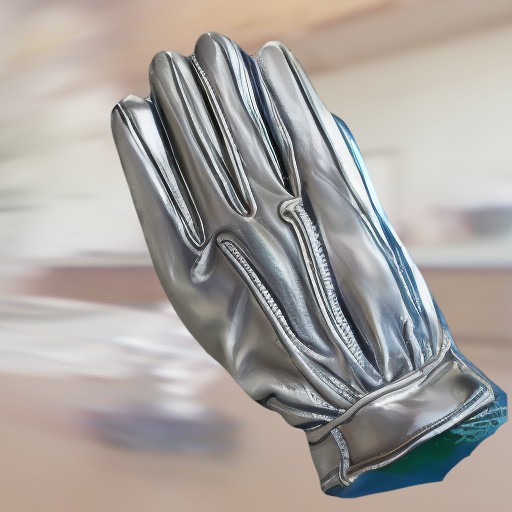}
 &\includegraphics[width=\moreResFullFigWidth]{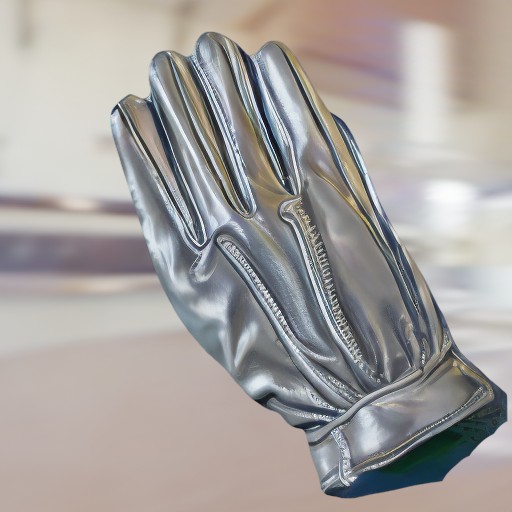}\\
  \multicolumn{4} {c} {
    Prompt: \emph{``leather glove''}.
    }\\
 \end{tabular} 
 \caption{Text-to-image generated results with lighting control. The
   first column shows the provisional image as a reference, whereas
   the last three columns are generated under different user-specified
   environment lighting conditions. }
  \label{fig:text2img_supp04}
\end{figure*}

%!TEX root = ../../supplementary.tex

\begin{figure*}
\renewcommand{\arraystretch}{0.8}
\addtolength{\tabcolsep}{-5.0pt}
 \begin{tabular}{ cccc }
\includegraphics[width=\moreResFullFigWidth]{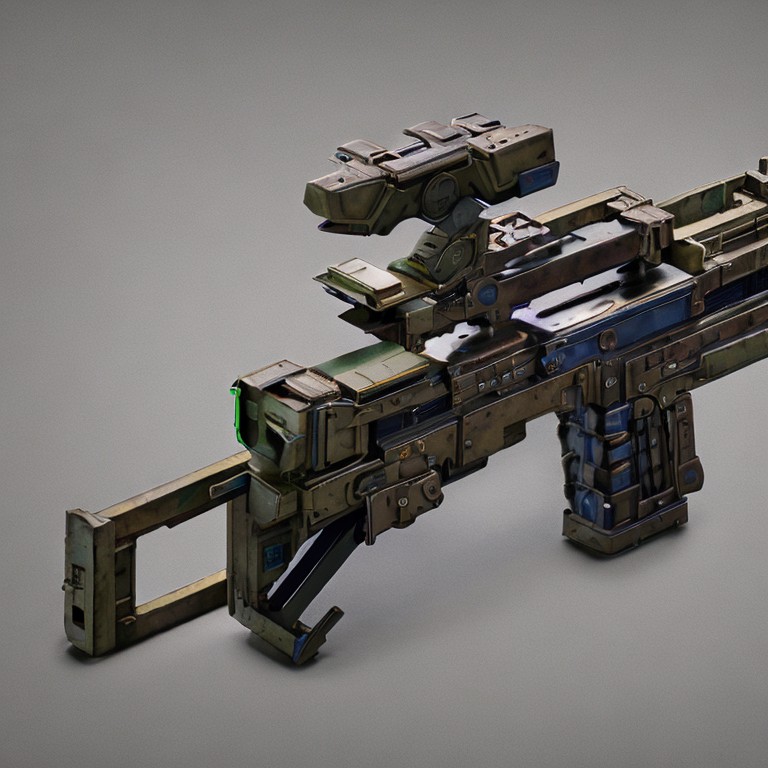}
 &\includegraphics[width=\moreResFullFigWidth]{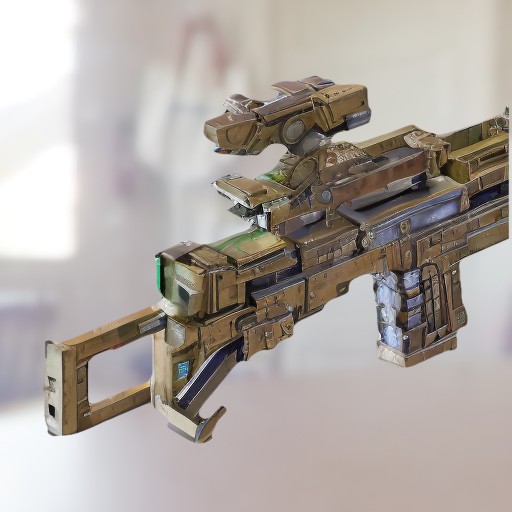}
 &\includegraphics[width=\moreResFullFigWidth]{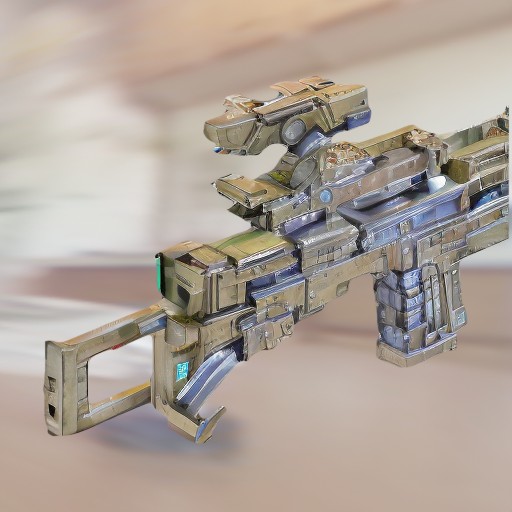}
 &\includegraphics[width=\moreResFullFigWidth]{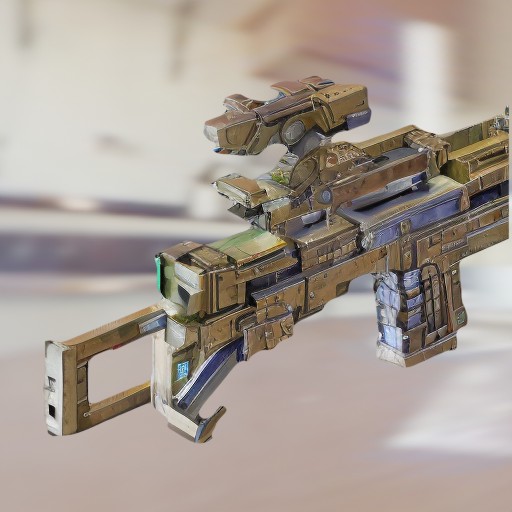}\\
  \multicolumn{4} {c} {
    Prompt: \emph{``starcraft 2 marine machine gun''}.
    }\\
  \includegraphics[width=\moreResFullFigWidth]{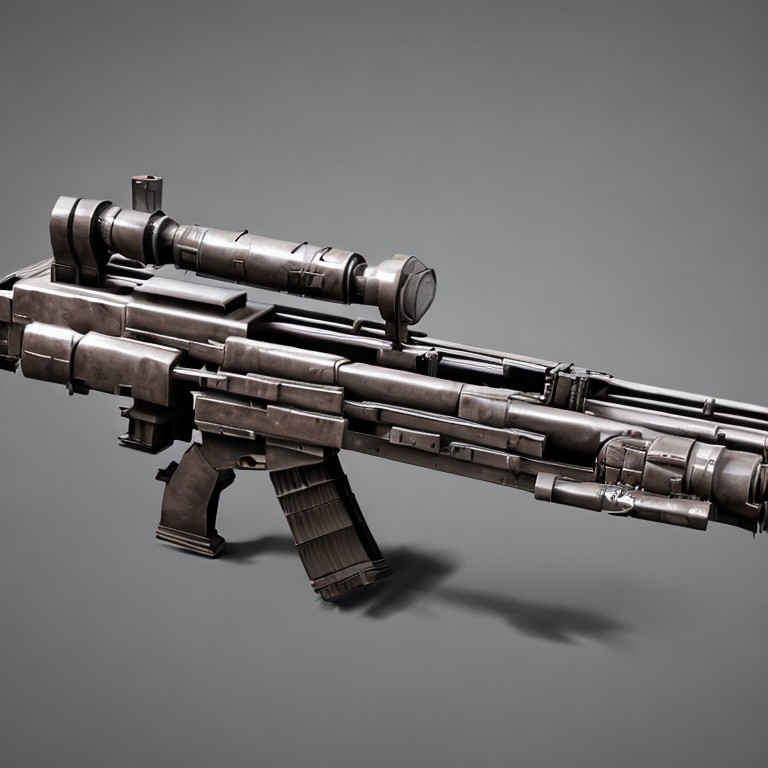}
 &\includegraphics[width=\moreResFullFigWidth]{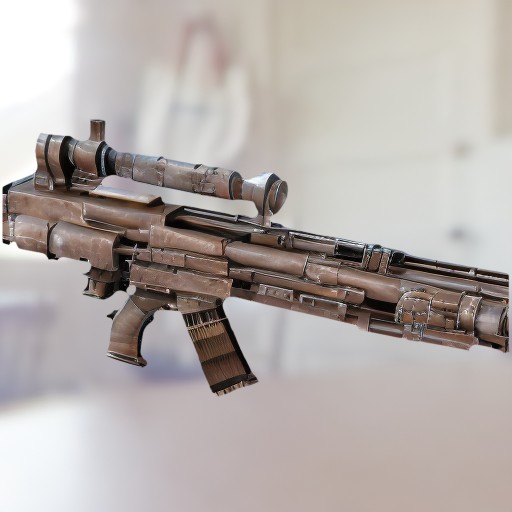}
 &\includegraphics[width=\moreResFullFigWidth]{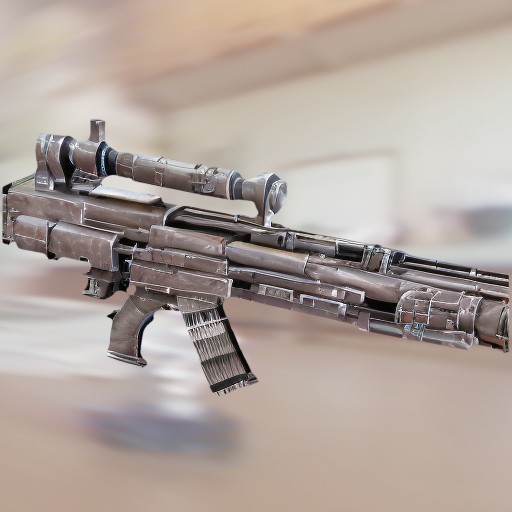}
 &\includegraphics[width=\moreResFullFigWidth]{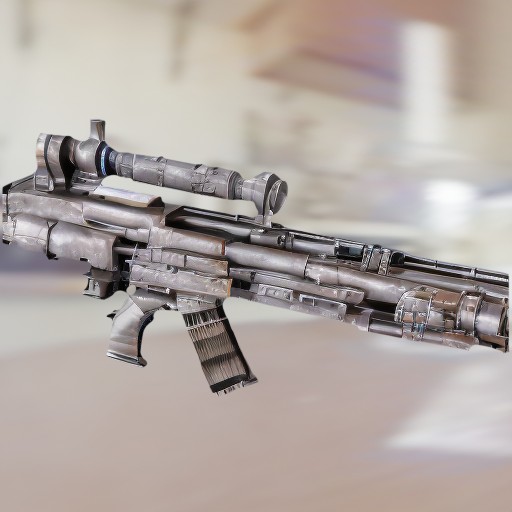}\\
  \multicolumn{4} {c} {
    Prompt: \emph{``starcraft 2 marine machine gun''}.
    }\\
  \includegraphics[width=\moreResFullFigWidth]{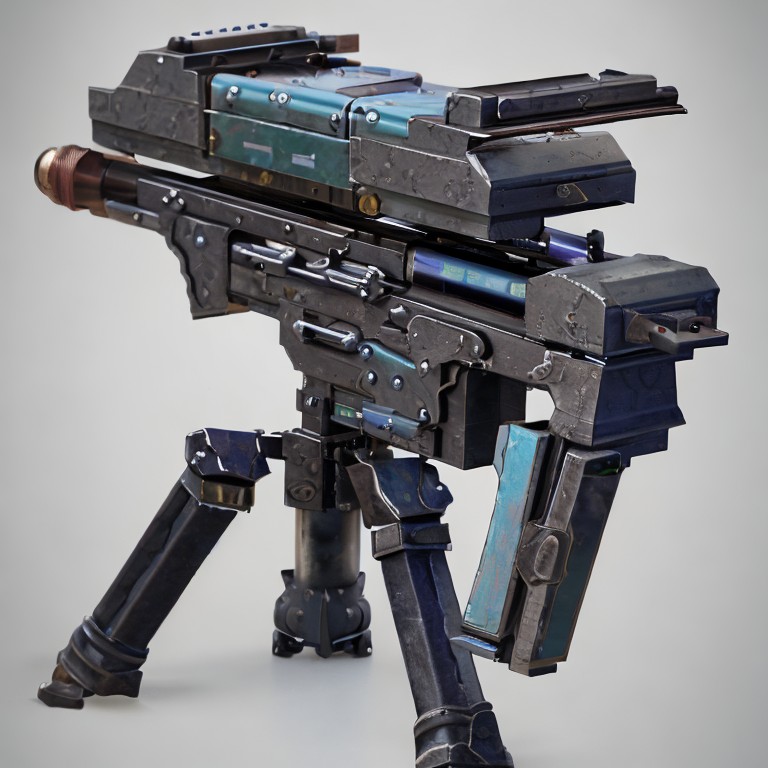}
 &\includegraphics[width=\moreResFullFigWidth]{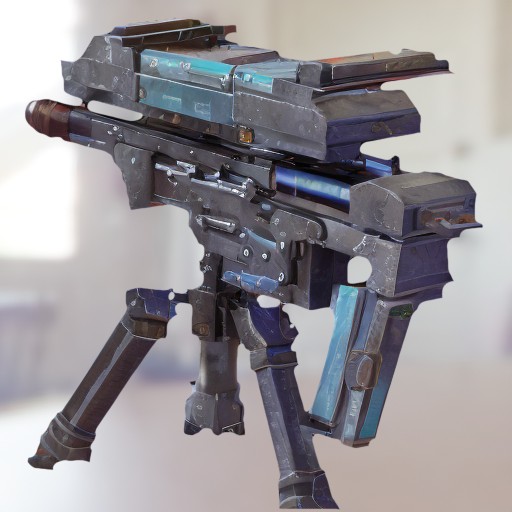}
 &\includegraphics[width=\moreResFullFigWidth]{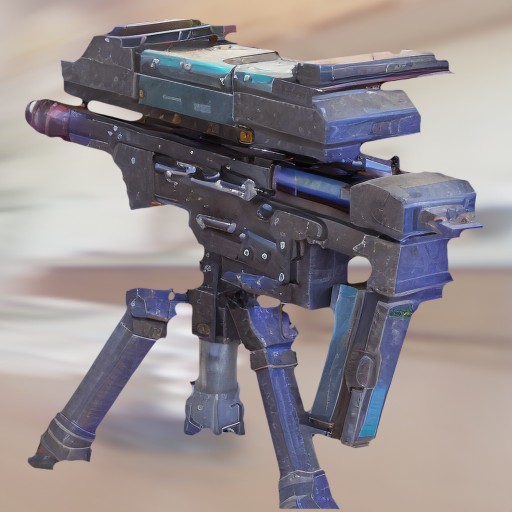}
 &\includegraphics[width=\moreResFullFigWidth]{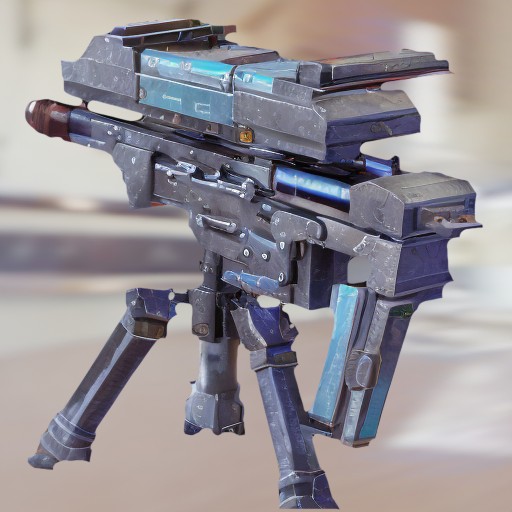}\\
  \multicolumn{4} {c} {
    Prompt: \emph{``starcraft 2 marine machine gun''}.
    }\\
  \includegraphics[width=\moreResFullFigWidth]{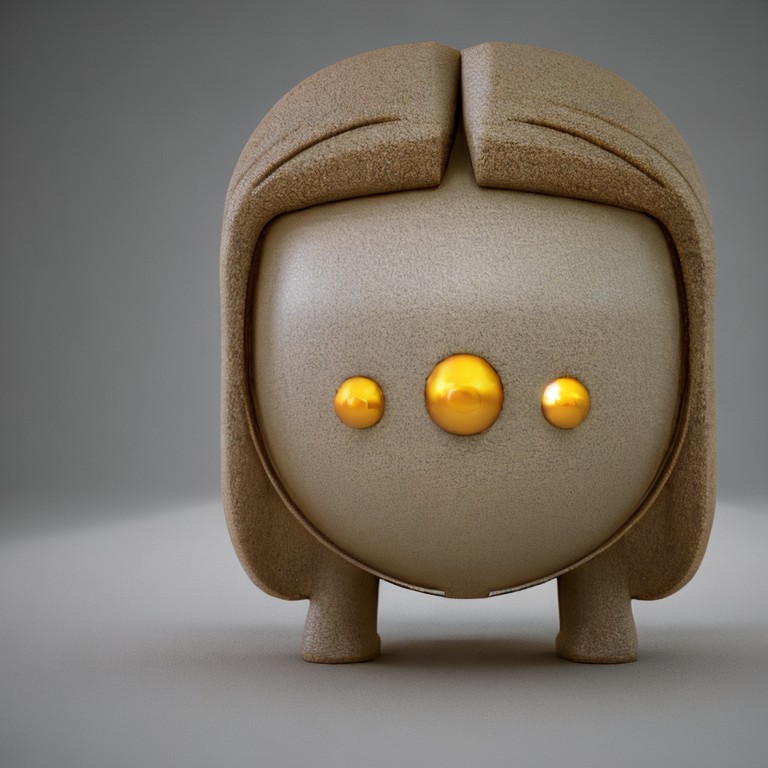}
  &\includegraphics[width=\moreResFullFigWidth]{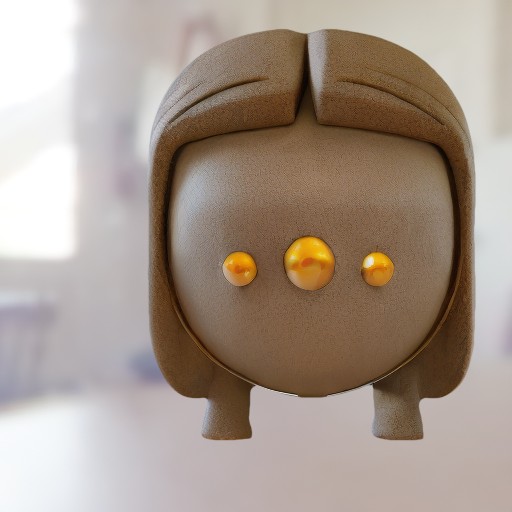}
  &\includegraphics[width=\moreResFullFigWidth]{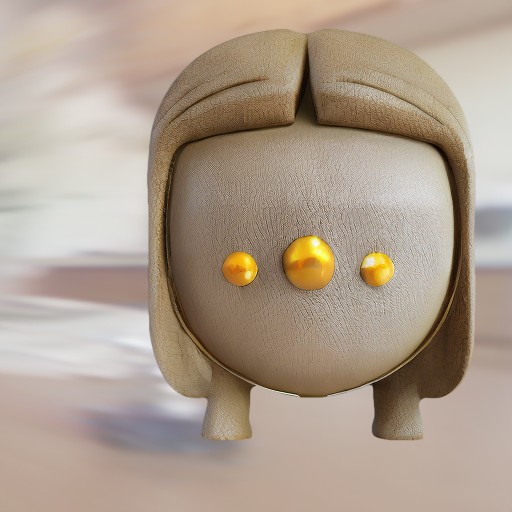}
  &\includegraphics[width=\moreResFullFigWidth]{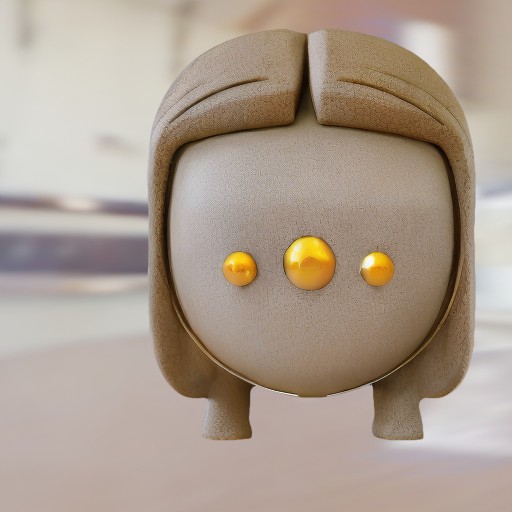}\\
    \multicolumn{4} {c} {
      Prompt: \emph{``3d animation character minimal art toy''}.
      }\\
 \end{tabular} 
 \caption{Text-to-image generated results with lighting control. The
   first column shows the provisional image as a reference, whereas
   the last three columns are generated under different user-specified
   environment lighting conditions. }
  \label{fig:text2img_supp02}
\end{figure*}

%!TEX root = ../../supplementary.tex

\begin{figure*}
\renewcommand{\arraystretch}{0.8}
\addtolength{\tabcolsep}{-5.0pt}
 \begin{tabular}{ cccc }
 \includegraphics[width=\moreResFullFigWidth]{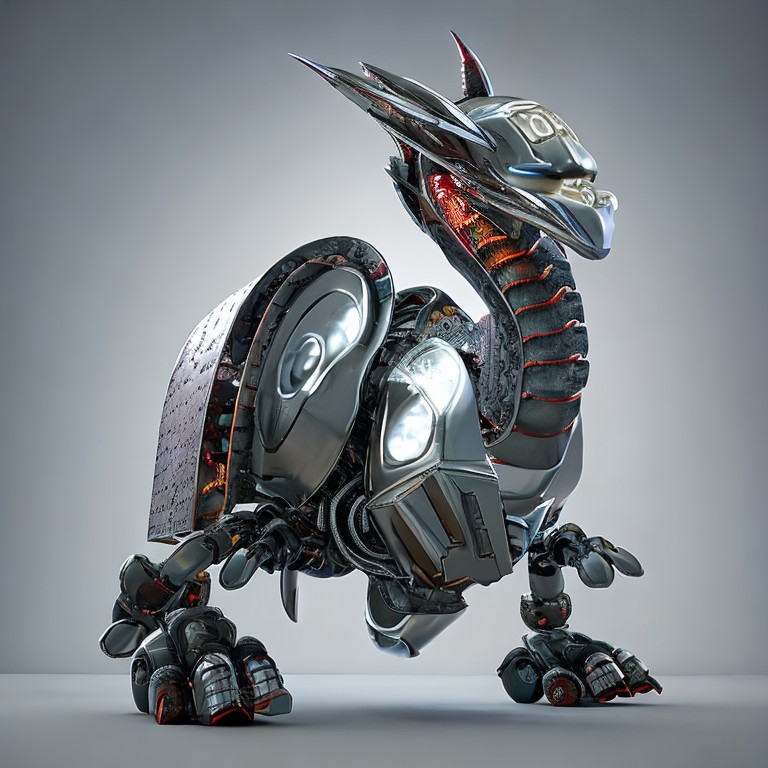}
 &\includegraphics[width=\moreResFullFigWidth]{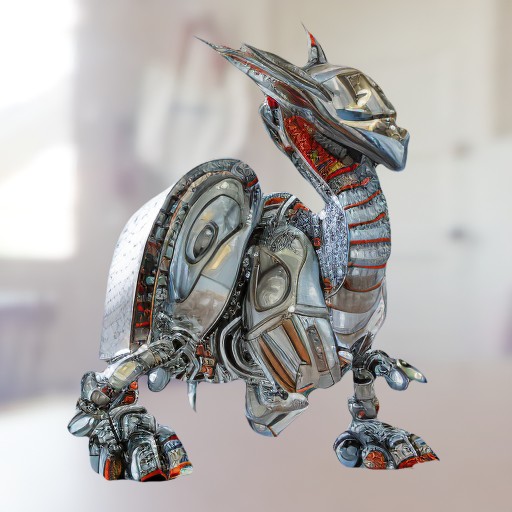}
 &\includegraphics[width=\moreResFullFigWidth]{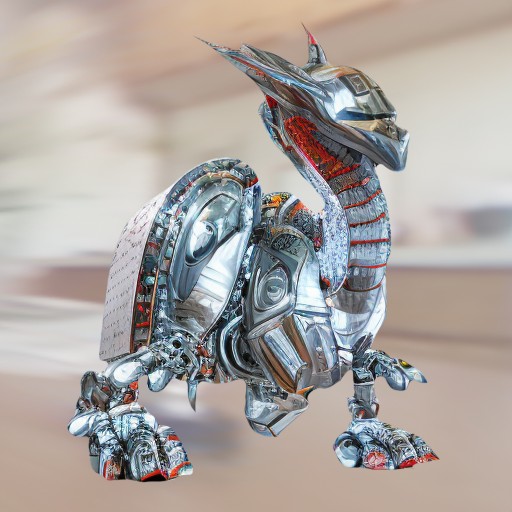}
 &\includegraphics[width=\moreResFullFigWidth]{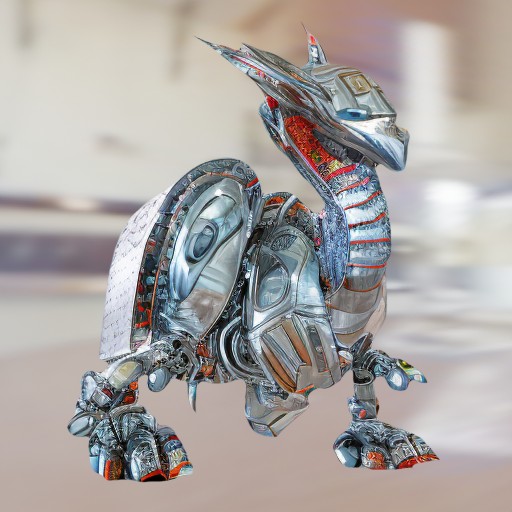}\\
  \multicolumn{4} {c} {
    Prompt: \emph{``machine dragon robot in platinum''}.
    }\\
  \includegraphics[width=\moreResFullFigWidth]{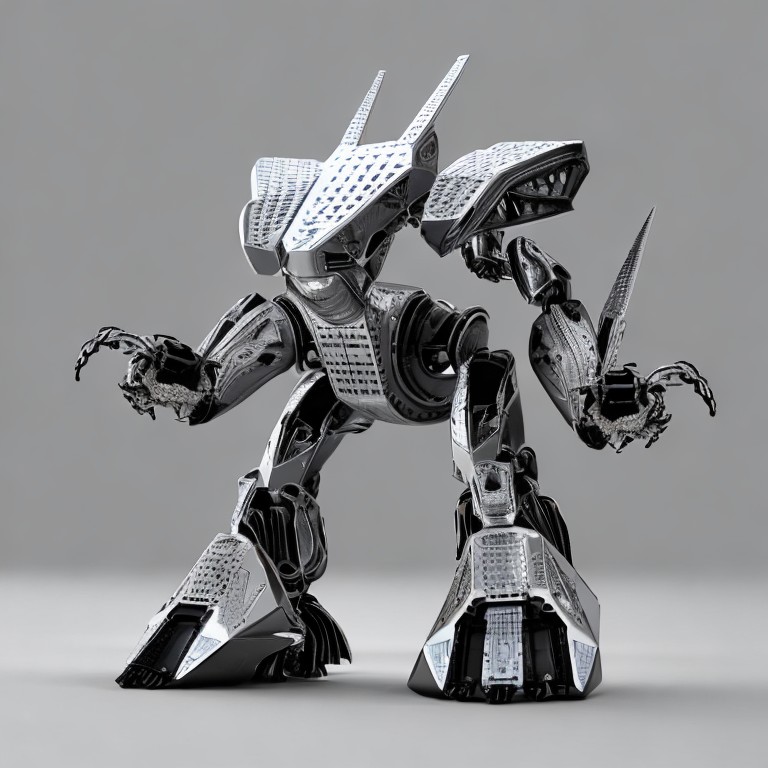}
 &\includegraphics[width=\moreResFullFigWidth]{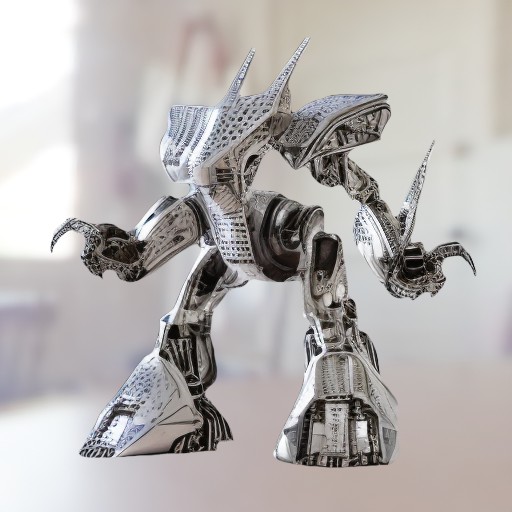}
 &\includegraphics[width=\moreResFullFigWidth]{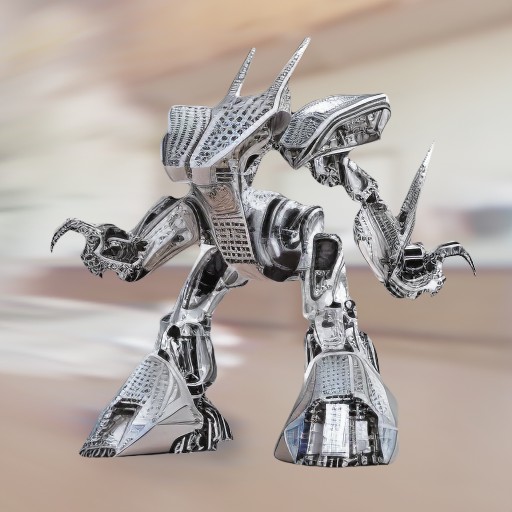}
 &\includegraphics[width=\moreResFullFigWidth]{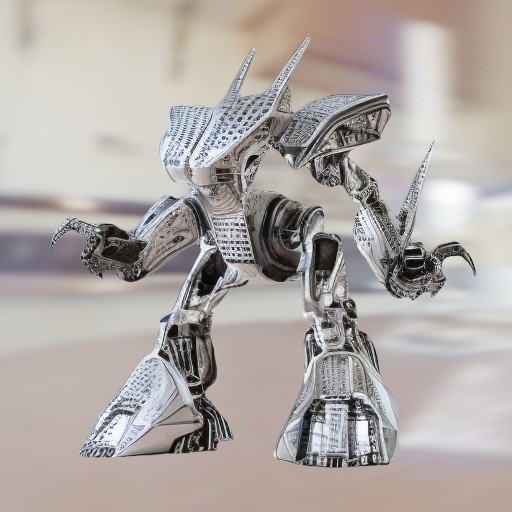}\\
  \multicolumn{4} {c} {
    Prompt: \emph{``machine dragon robot in platinum''}.
    }\\
  \includegraphics[width=\moreResFullFigWidth]{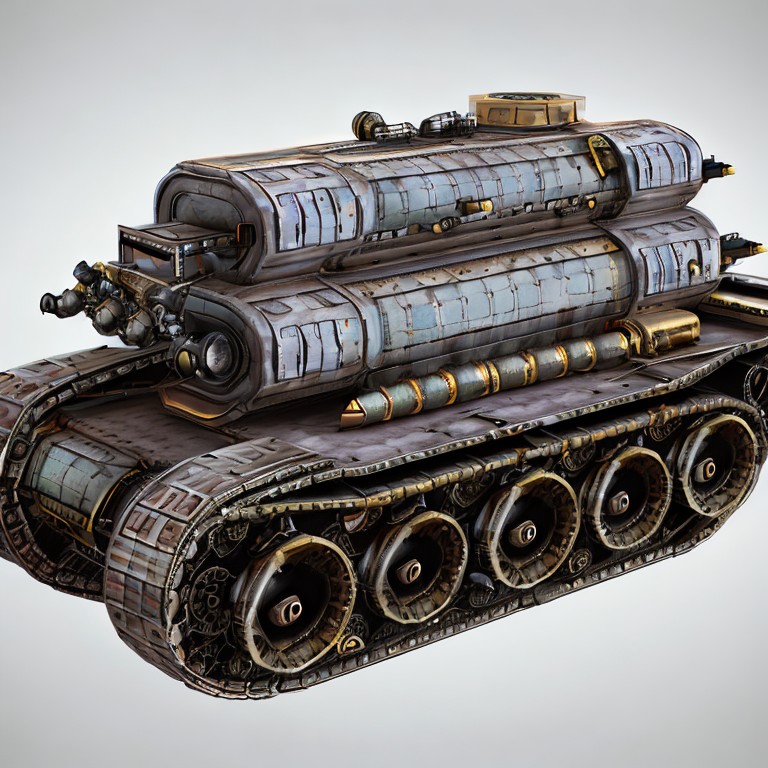}
  &\includegraphics[width=\moreResFullFigWidth]{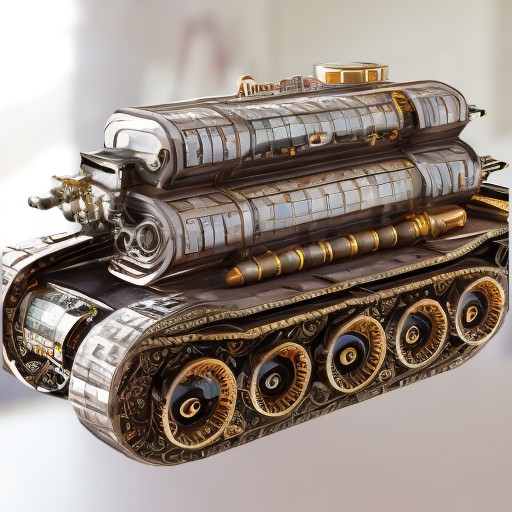}
  &\includegraphics[width=\moreResFullFigWidth]{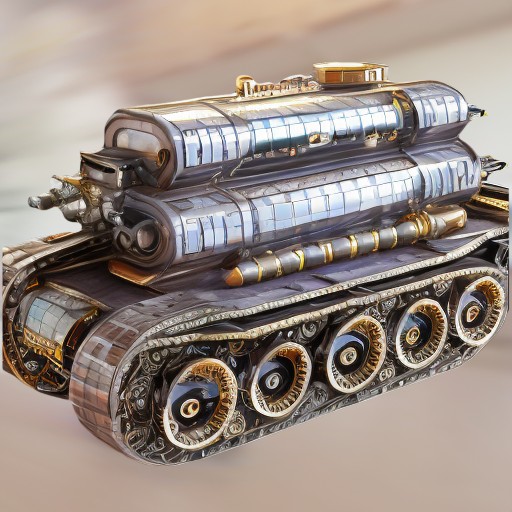}
  &\includegraphics[width=\moreResFullFigWidth]{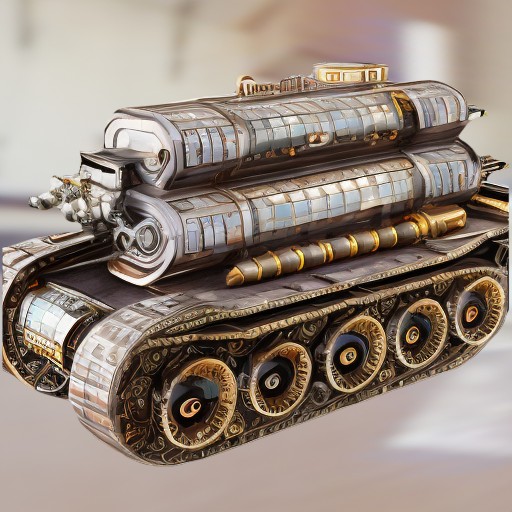}\\
    \multicolumn{4} {c} {
      Prompt: \emph{``steampunk space tank with delicate details''}.
    }\\
  \includegraphics[width=\moreResFullFigWidth]{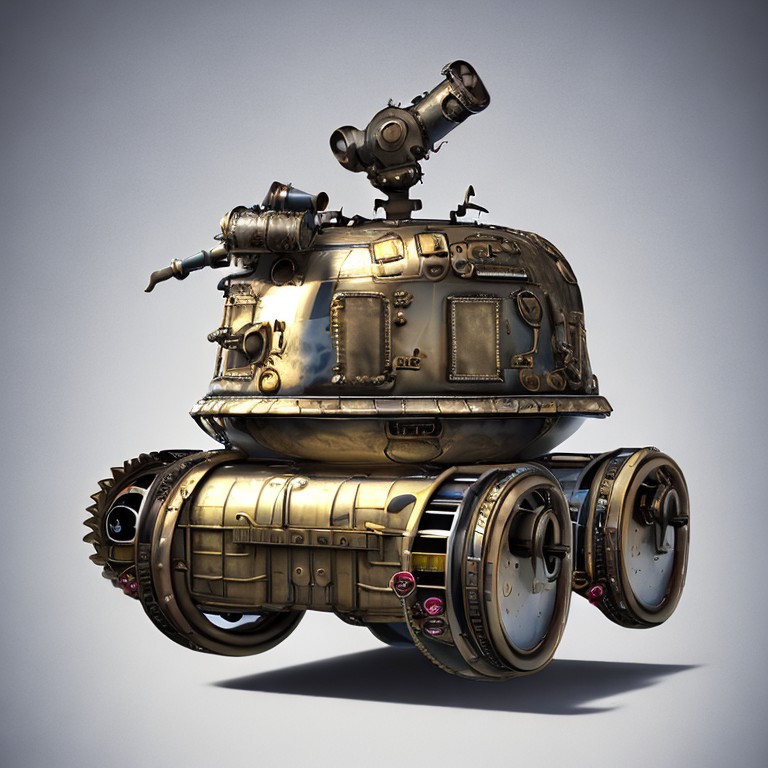}
  &\includegraphics[width=\moreResFullFigWidth]{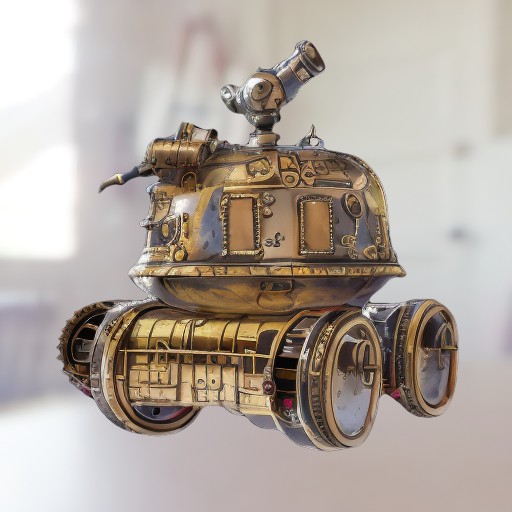}
  &\includegraphics[width=\moreResFullFigWidth]{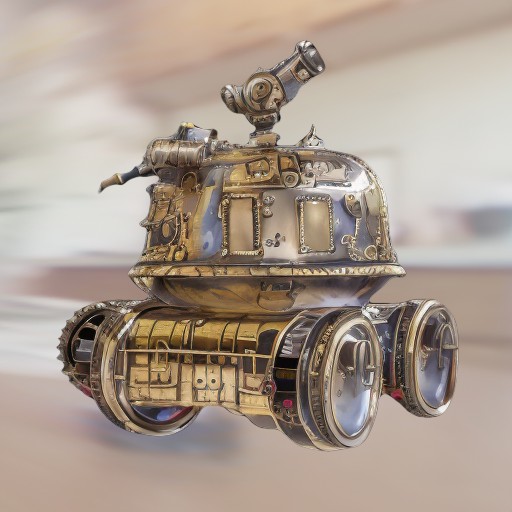}
  &\includegraphics[width=\moreResFullFigWidth]{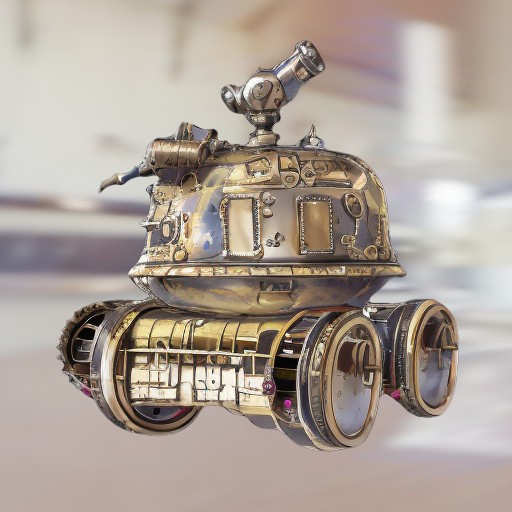}\\
  \multicolumn{4} {c} {
    Prompt: \emph{``steampunk space tank with delicate details''}.
    }\\
 \end{tabular} 
 \caption{Text-to-image generated results with lighting control. The
   first column shows the provisional image as a reference, whereas
   the last three columns are generated under different user-specified
   environment lighting conditions. }
  \label{fig:text2img_supp03}
\end{figure*}

%!TEX root = ../../supplementary.tex

\begin{figure*}
\renewcommand{\arraystretch}{0.8}
\addtolength{\tabcolsep}{-5.0pt}
 \begin{tabular}{ cccc }
 \includegraphics[width=\moreResFullFigWidth]{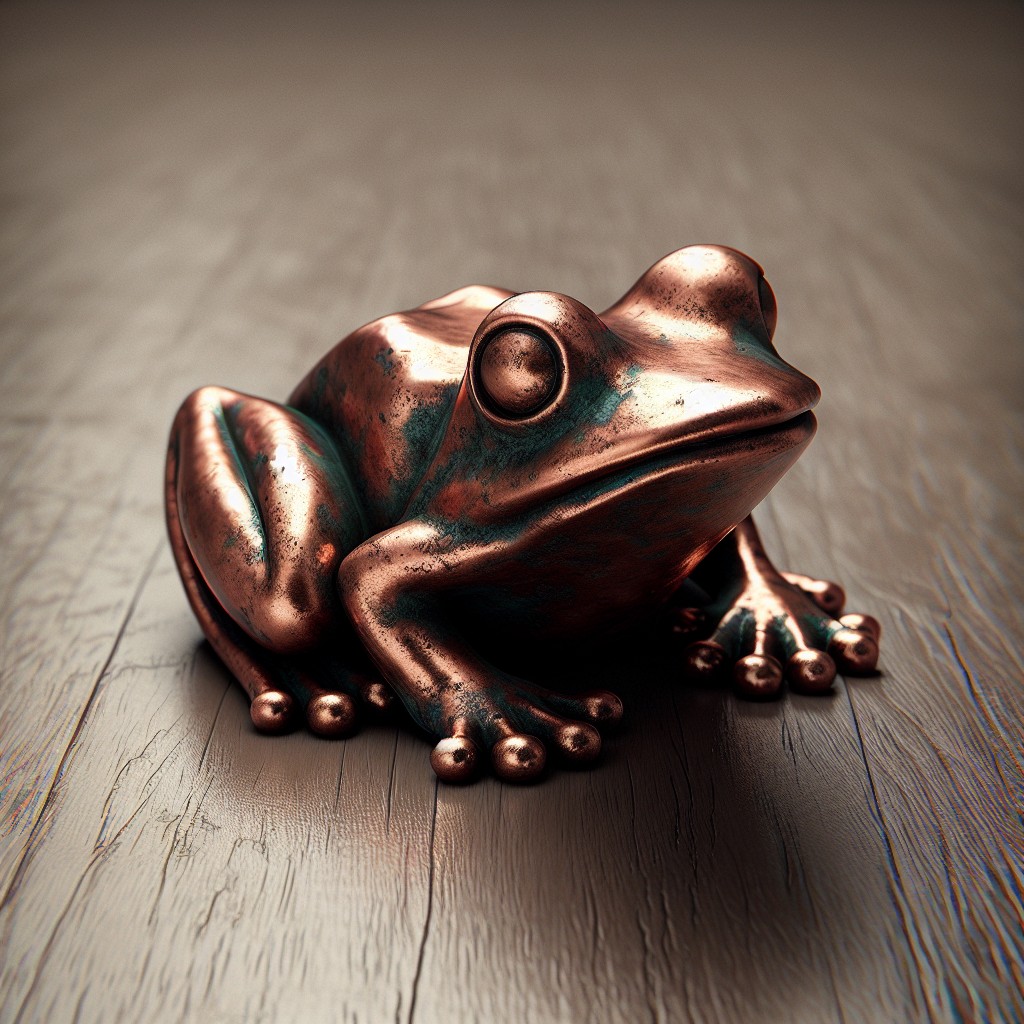}
 &\includegraphics[width=\moreResFullFigWidth]{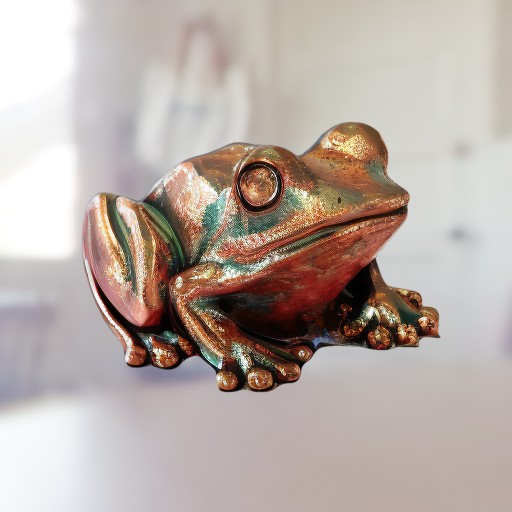}
 &\includegraphics[width=\moreResFullFigWidth]{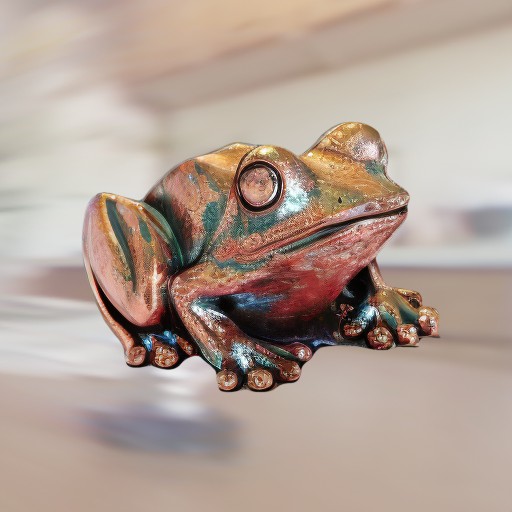}
 &\includegraphics[width=\moreResFullFigWidth]{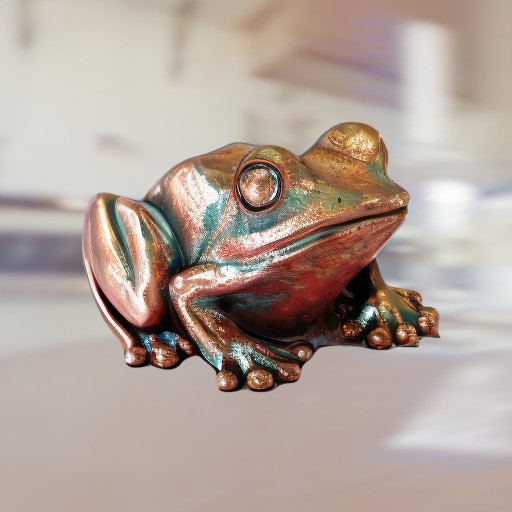}\\
  \multicolumn{4} {c} {
    Prompt: \emph{``Rusty copper toy frog with spatially varying materials some parts are shinning other parts are rough''}.
    }\\ 
  \includegraphics[width=\moreResFullFigWidth]{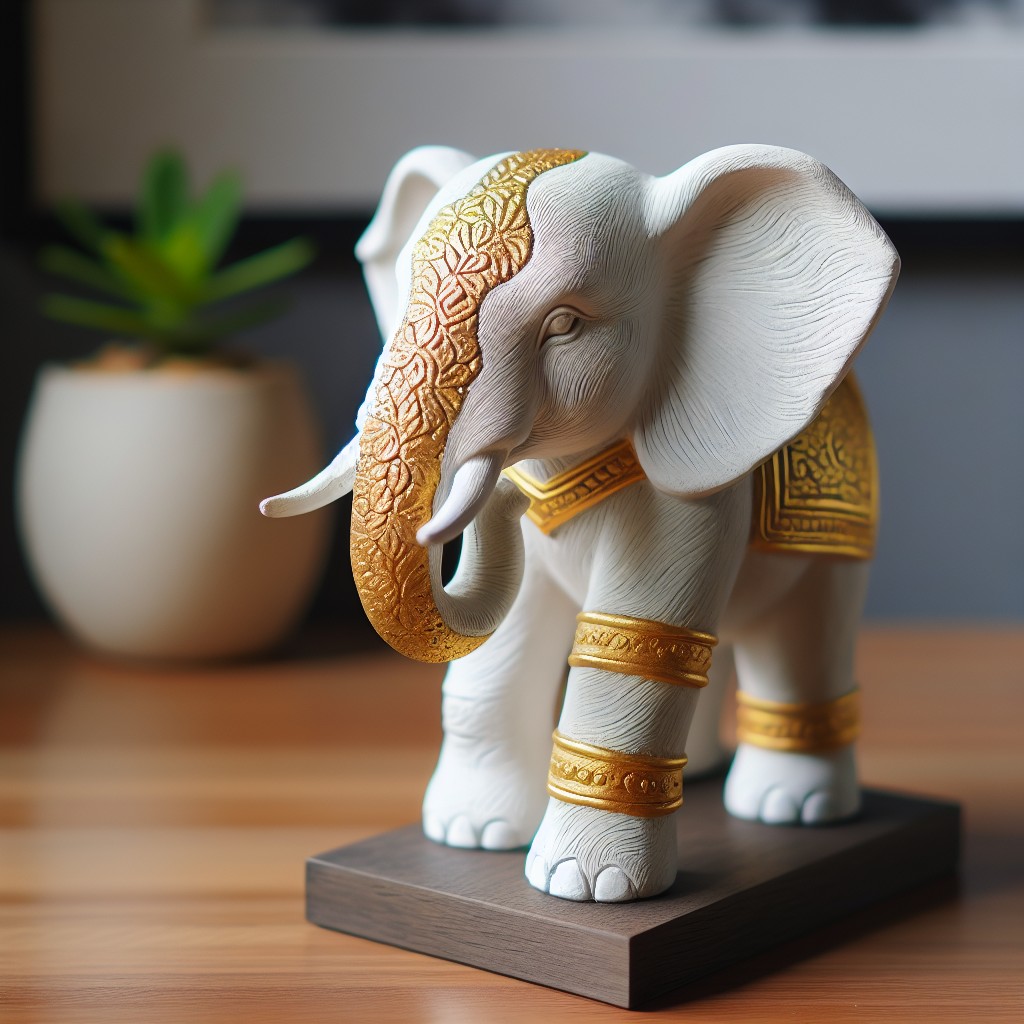}
 &\includegraphics[width=\moreResFullFigWidth]{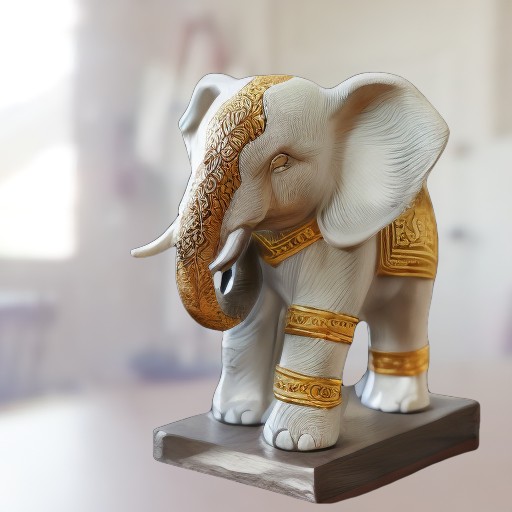}
 &\includegraphics[width=\moreResFullFigWidth]{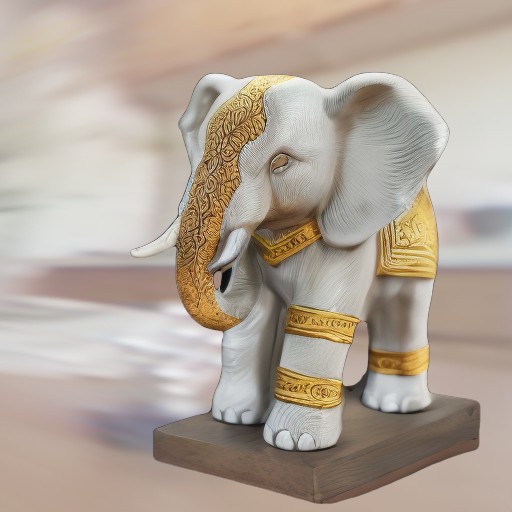}
 &\includegraphics[width=\moreResFullFigWidth]{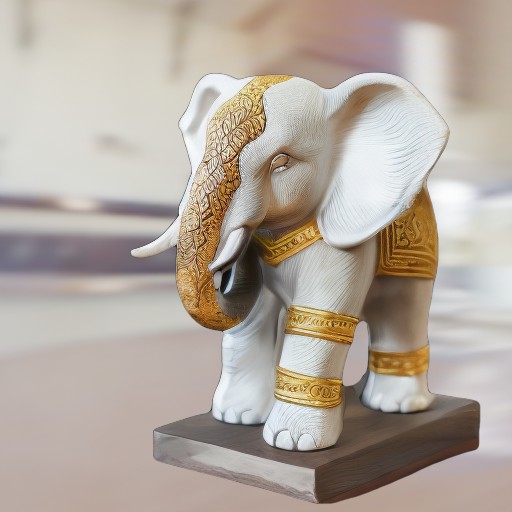}\\
  \multicolumn{4} {c} {
    Prompt: \emph{``An elephant sculpted from plaster and the elephant nose is decorated with the golden texture''}.
    }\\
  \includegraphics[width=\moreResFullFigWidth]{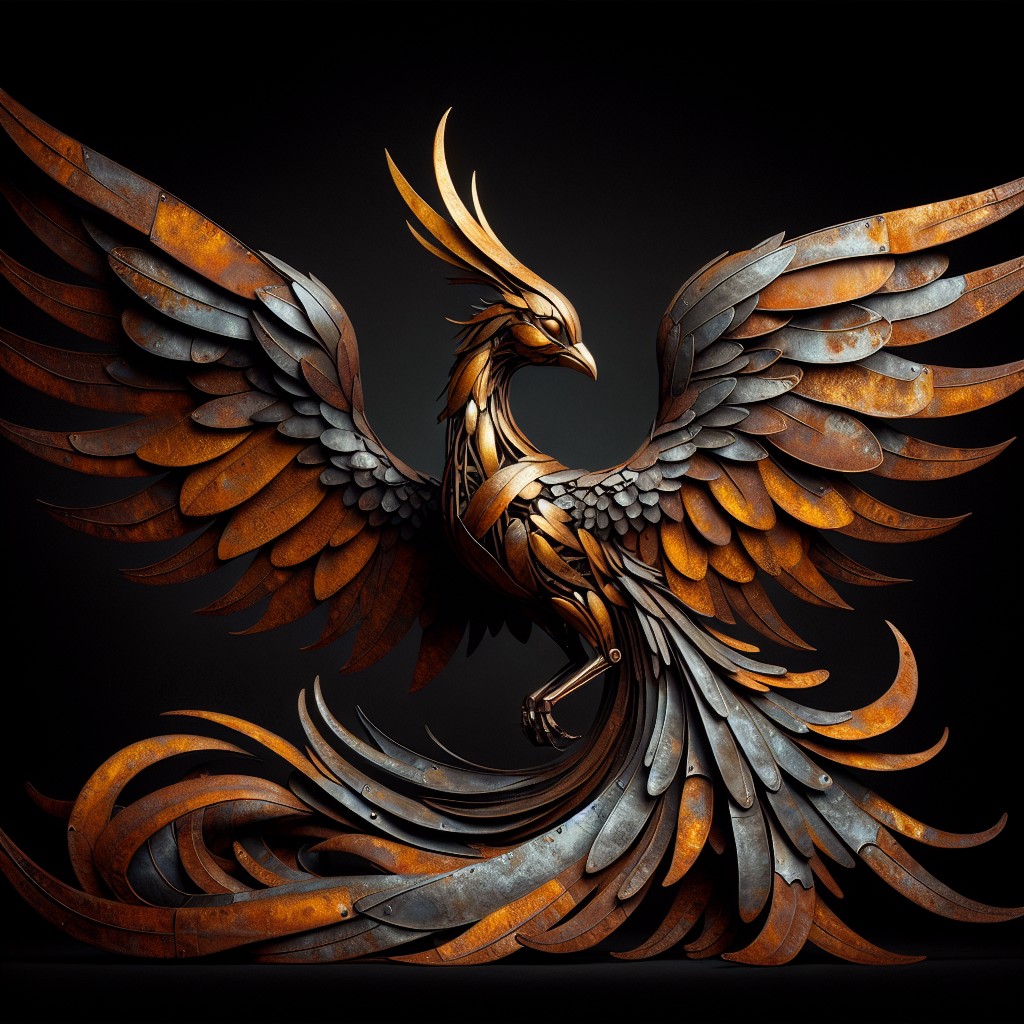}
 &\includegraphics[width=\moreResFullFigWidth]{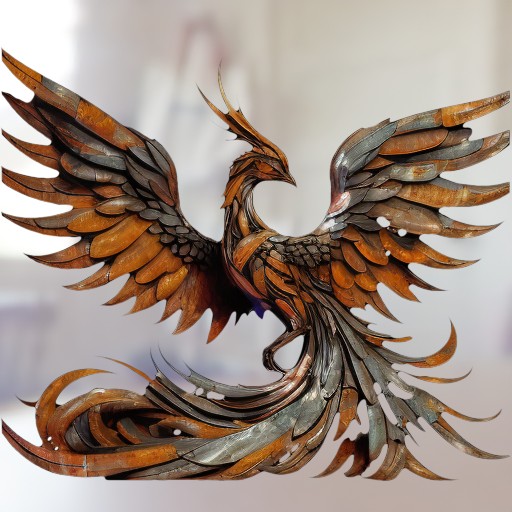}
 &\includegraphics[width=\moreResFullFigWidth]{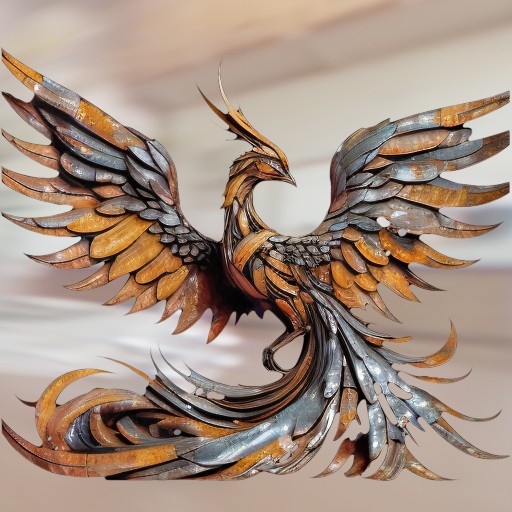}
 &\includegraphics[width=\moreResFullFigWidth]{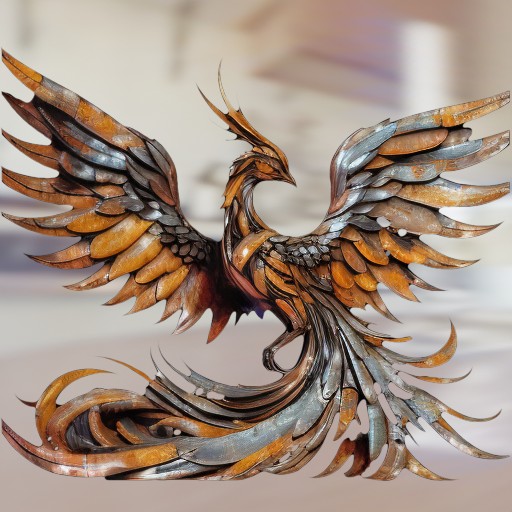}\\
  \multicolumn{4} {c} {
    Prompt: \emph{``Rusty sculpture of a phoenix with its head more polished yet the wings are more rusty''}.
    }\\    
  \includegraphics[width=\moreResFullFigWidth]{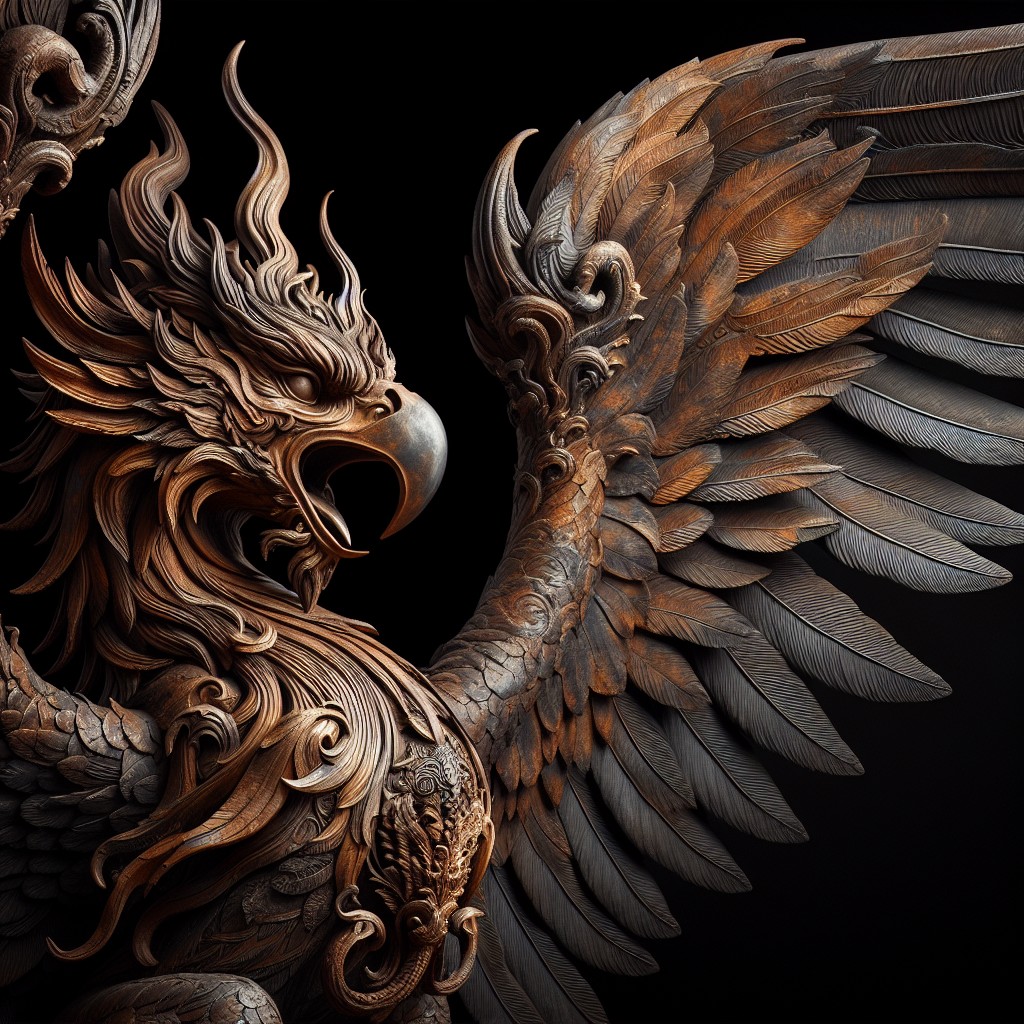}
 &\includegraphics[width=\moreResFullFigWidth]{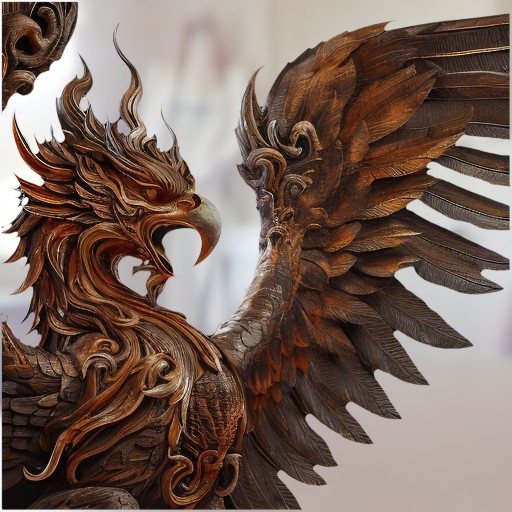}
 &\includegraphics[width=\moreResFullFigWidth]{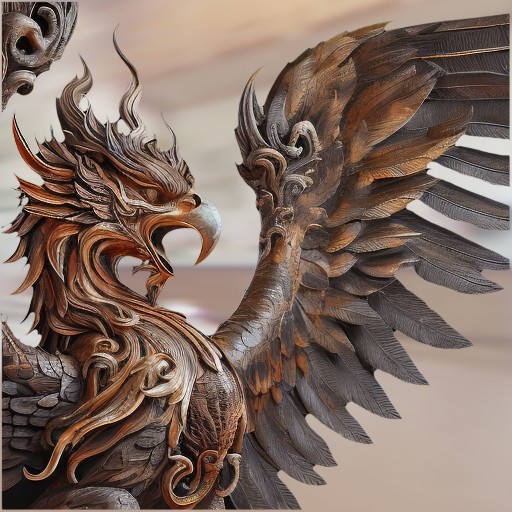}
 &\includegraphics[width=\moreResFullFigWidth]{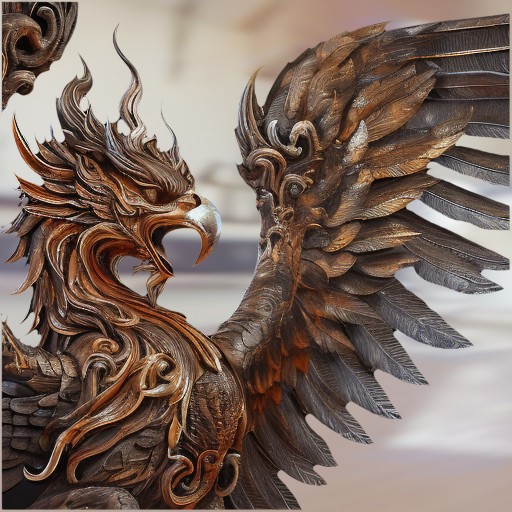}\\
  \multicolumn{4} {c} {
    Prompt: \emph{``Rusty sculpture of a phoenix with its head more polished yet the wings are more rusty''}.
    }\\   
 \end{tabular} 
 \caption{Text-to-image generated results with lighting control. The
   first column shows the provisional image as a reference, whereas
   the last three columns are generated under different user-specified
   environment lighting conditions. The provisional images 
   are generated with \emph{DALL-E3} instead of \emph{stable
     diffusion v2.1} to better handle the more complex prompt.}
  \label{fig:text2img_DE01}
\end{figure*}

%!TEX root = ../../supplementary.tex

\begin{figure*}
\renewcommand{\arraystretch}{0.8}
\addtolength{\tabcolsep}{-5.0pt}
 \begin{tabular}{ cccc }
 \includegraphics[width=\moreResFullFigWidth]{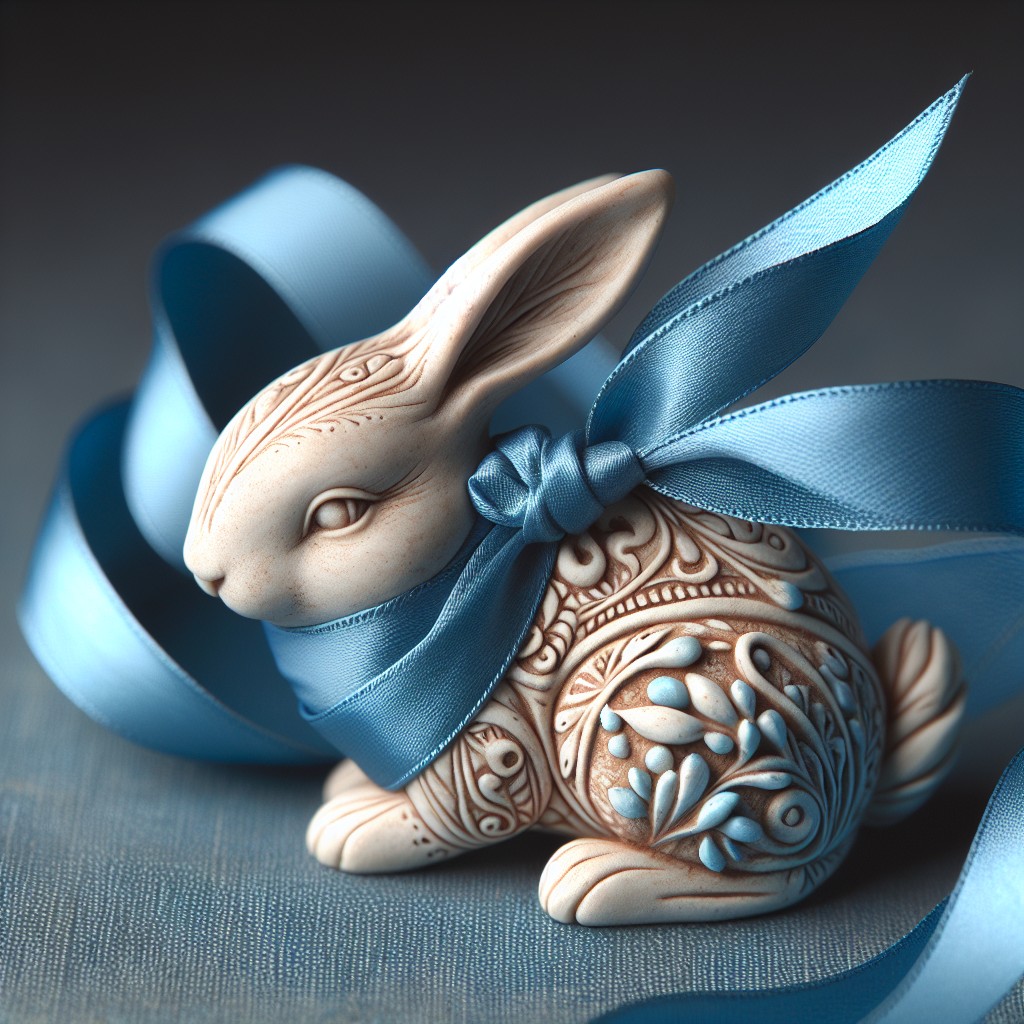}
 &\includegraphics[width=\moreResFullFigWidth]{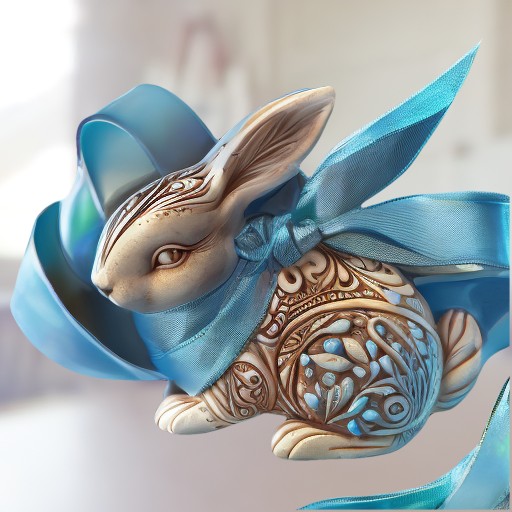}
 &\includegraphics[width=\moreResFullFigWidth]{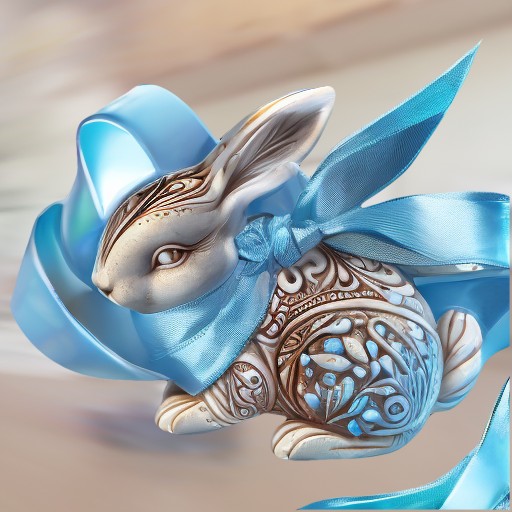}
 &\includegraphics[width=\moreResFullFigWidth]{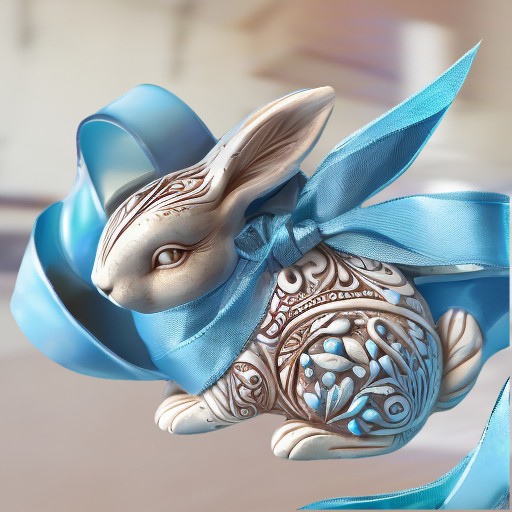}\\
  \multicolumn{4} {c} {
    Prompt: \emph{``A decorated plaster rabbit toy plate with blue fine silk ribbon around it''}.
    }\\ 
  \includegraphics[width=\moreResFullFigWidth]{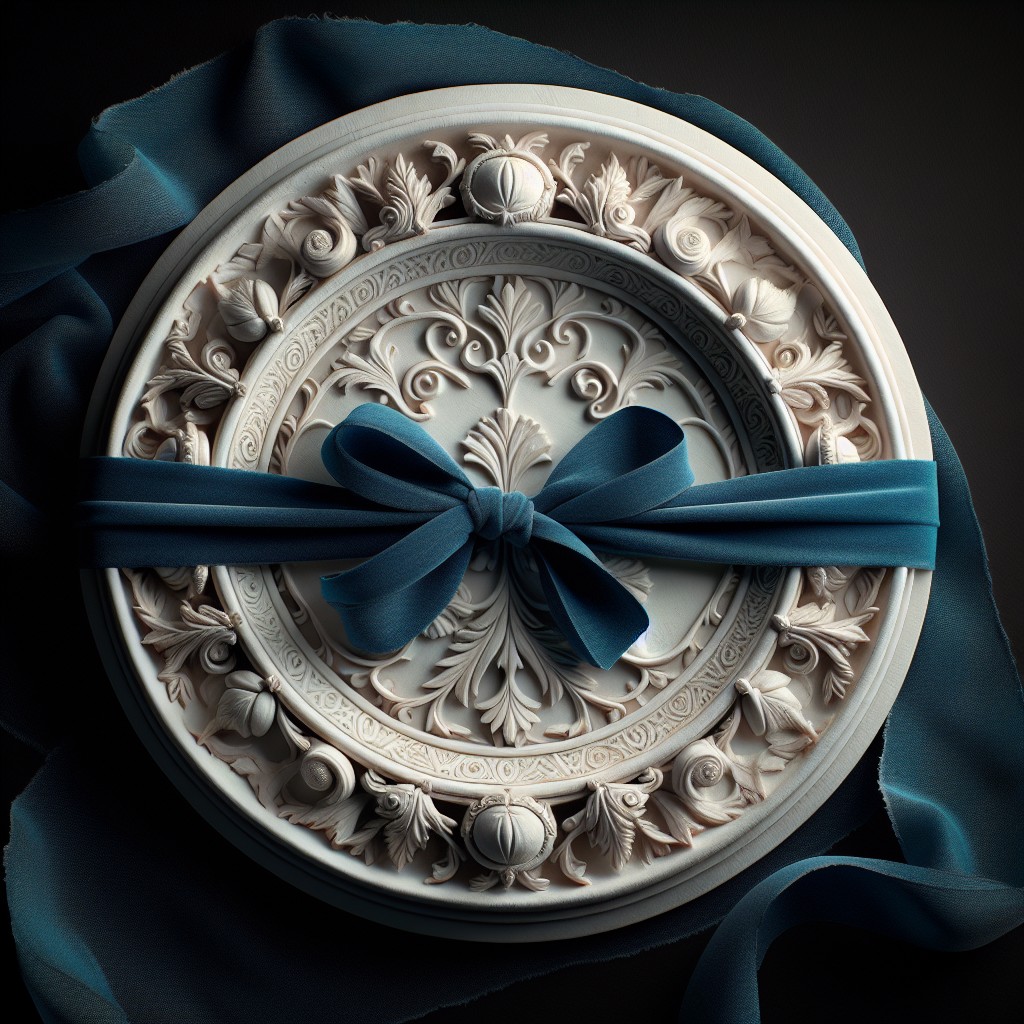}
 &\includegraphics[width=\moreResFullFigWidth]{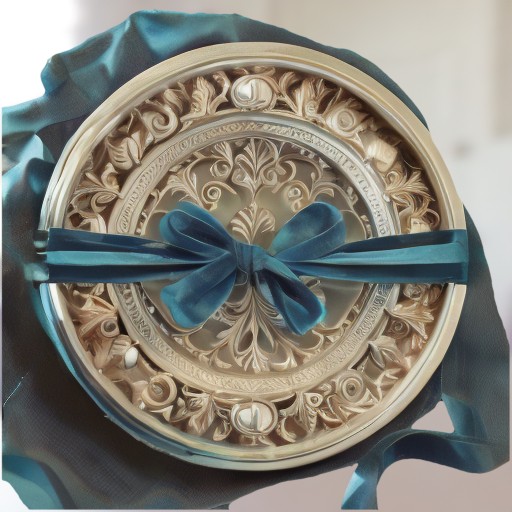}
 &\includegraphics[width=\moreResFullFigWidth]{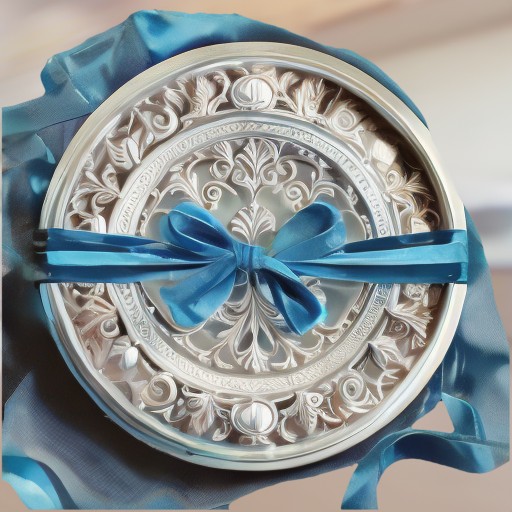}
 &\includegraphics[width=\moreResFullFigWidth]{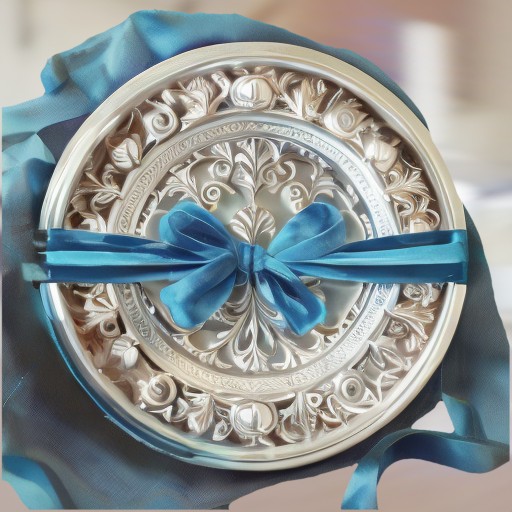}\\
  \multicolumn{4} {c} {
    Prompt: \emph{``A decorated plaster round plate with blue fine silk ribbon around it''}.
    }
 \end{tabular} 
 \caption{Text-to-image generated results with lighting control. The
   first column shows the provisional image as a reference, whereas
   the last three columns are generated under different user-specified
   environment lighting conditions. The provisional images 
   are generated with \emph{DALL-E3} instead of \emph{stable
     diffusion v2.1} to better handle the more complex prompt.}
  \label{fig:text2img_DE02}
\end{figure*}

%!TEX root = ../../supplementary.tex

\end{document}